\documentclass{article} \usepackage{iclr2020_conference,times}

\usepackage{hyperref}
\usepackage{url}

\usepackage[ruled]{algorithm2e}
\usepackage{amsmath}
\usepackage{graphicx}
\usepackage{stackengine}
\usepackage{xcolor} \usepackage{placeins}
\usepackage{amssymb}

\SetKwInOut{Input}{Input}
\SetKwInOut{Output}{Output}

\renewcommand{\vec}[1]{\mathbf{#1}}

\def\delequal{\mathrel{\ensurestackMath{\stackon[1pt]{=}{\scriptscriptstyle\Delta}}}}

\newcommand{\EnvDist}{{p_\mathcal{E}}}
\newcommand{\Proposal}{{\mu}}
\newcommand{\Uniform}{{U}}
\newcommand{\Best}{{*}}
\newcommand{\Batch}{{b}}
\DeclareMathOperator*{\argmax}{arg\ max}

\newcommand{\PolicyParams}{\theta}
\newcommand{\ProposalParams}{\PolicyParams^{\Proposal}}
\newcommand{\QParams}{\PolicyParams^{Q}}

\newcommand{\EpsilonProbability}{\epsilon}
\newcommand{\NumProposalActions}{N}
\newcommand{\NumUniformActions}{M}

\newcommand{\EpisodeLength}{T}

\newcommand{\State}{s}
\newcommand{\Action}{\vec{a}}
\newcommand{\UniformAction}{\Action^\Uniform}
\newcommand{\ProposalAction}{\Action^\Proposal}
\newcommand{\ActionBest}{\Action^\Best}
\newcommand{\ActionBestBatch}{\Action^{\Best, \Batch}}
\newcommand{\ActionSpace}{\mathcal{A}}
\newcommand{\StateTrajectory}{\State_{1:\EpisodeLength}}
\newcommand{\ActionTrajectory}{\Action_{1:\EpisodeLength}}
\newcommand{\ActionBestTrajectory}{\ActionBest_{1:\EpisodeLength}}
\newcommand{\RewardTrajectory}{\Reward_{1:\EpisodeLength}}

\newcommand{\Reward}{r}

\newcommand{\ReplayBuffer}{\mathcal{R}}
\newcommand{\ExperienceBatch}{\mathcal{B}}
\newcommand{\BatchIndex}{{b}}
\newcommand{\BatchSize}{{B}}

\title{Q-Learning in enormous action spaces via amortized approximate maximization}

\author{Tom Van de Wiele\thanks{Work done while at DeepMind. Address correspondence to \texttt{\{dwf,vmnih\}@google.com}},
	David Warde-Farley,
	Andriy Mnih \&
	Volodymyr Mnih \\
	DeepMind \\
	London, United Kingdom \\
	\texttt{tvdwiele@gmail.com}, \texttt{\{dwf,amnih,vmnih\}@google.com} \\
}

\iclrfinalcopy \begin{document}

\maketitle

\begin{abstract}
Applying Q-learning to high-dimensional or continuous action spaces can be difficult due to the required maximization over the set of possible actions.
Motivated by techniques from amortized inference, we replace the expensive maximization over all actions with a maximization over a small subset of possible actions sampled from a learned proposal distribution.
The resulting approach, which we dub Amortized Q-learning (AQL), is able to handle discrete, continuous, or hybrid action spaces while maintaining the benefits of Q-learning.
Our experiments on continuous control tasks with up to 21 dimensional actions show that AQL outperforms D3PG~\citep{barth2018distributed} and QT-Opt~\citep{kalashnikov2018qt}. Experiments on structured discrete action spaces demonstrate that AQL can efficiently learn good policies in spaces with thousands of discrete actions.
 \end{abstract}

\section{Introduction}
In the recent resurgence of interest in combining reinforcement learning with neural network function approximators, Q-learning \citep{watkins1992q} and its many neural variants \citep{mnih2015human,bellemare2017distributional} have remained competitive with the state-of-the-art methods~\citep{horgan2018distributed,kapturowski2019recurrent}.
The simplicity of Q-learning makes it relatively straightforward to implement, even in combination with neural networks \citep{mnih2015human}.
Because Q-learning is an off-policy reinforcement learning algorithm, it is trivial to combine with techniques like experience replay~\citep{lin1993reinforcement} for improved data efficiency or to implement in a distributed training setting where experience is generated using slightly stale network parameters, without requiring importance sampling-based off-policy corrections used by stochastic actor-critic methods~\citep{espeholt2018impala}.
Q-learning can also be easily and robustly combined with exotic exploration strategies~\citep{ostrovski2017count,mnih2016asynchronous,horgan2018distributed} as well as data augmentation through post-hoc modification~\citep{kaelbling1993goals,hindsight2017}.

One limitation of Q-learning is the requirement to maximize over the set of possible actions, limiting its applicability in environments with continuous or high-cardinality discrete action spaces.
In the case of Q-learning with neural network function approximation, the common approach of computing $Q$ values for all actions in a single forward pass becomes infeasible, requiring instead an architecture which accepts as inputs both the state and the action, producing a scalar $Q$ estimate as output.
Other types of methods, such stochastic actor-critic or policy gradient approaches can naturally handle discrete, continuous, or even hybrid action spaces~\citep{williams1992simple,schulman2015trust,mnih2016asynchronous} because they do not perform a maximization over actions and only require the ability to efficiently sample actions from an appropriately parameterized stochastic policy.
Furthermore, the structure of an action space often suggests a stochastic policy parametrization which admits efficient sampling even when the number of distinct actions is enormous.
For example, a $K$-level discretization of a continuous action space with $D$ degrees of freedom (throughout this work, we will refer to this as a $D$-\emph{dimensional} action space) will have $K^D$ distinct actions, but a policy can be designed to represent these actions as a product of $D$ discrete distributions each with arity $K$, allowing for sampling in $\mathcal{O}(KD)$ rather than $\mathcal{O}(K^D)$ time.
The same principle cannot be directly applied to Q-learning, where identification of the maximal $Q$ value would require an exhaustive enumeration of all $K^D$ actions.
While a number of approaches for dealing with the intractable maximization over actions in Q-learning have been proposed in order to make it applicable to richer action spaces, these approaches are usually specific to a particular form of action space~\citep{hafner2009nfqca,pazis2009binary,yoshida2015binary}.

Here, we show that instead of performing an exact maximization over the set of actions at each time step, it can be preferable to \emph{learn} to search for the best action, thus amortizing the action selection cost over the course of training~\citep{hafner2009nfqca,lillicrap2016continuous}.
We treat the search for the best action as another learning problem and replace the exact maximization over all actions with a maximization over a set of actions sampled from a learned proposal distribution.
We show that an effective proposal can be learned by training a neural network to predict the best known action found by a stochastic search procedure.
This approach, which we dub Amortized Q-learning (AQL), is able to naturally handle high-dimensional discrete, continuous, or even hybrid action spaces (consisting of both discrete and continuous degrees of freedom) because, like stochastic actor-critic methods, AQL only requires the ability to sample actions from an appropriately parameterized proposal distribution.

We evaluate the effectiveness of Amortized Q-learning on both continuous actions spaces and large, but structured, discrete action spaces.
On continuous control tasks from the DeepMind Control Suite, Amortized Q-learning outperforms D3PG and QT-Opt, two strong continuous control methods.
On foraging tasks from the DeepMind Lab 3D first-person environment, Amortized Q-learning is able to learn effective policies using an action space with over 3500 actions.
AQL thus bridges the gap between Q-learning and stochastic actor-critic methods in their flexibility towards the action space, removing the need for Q-learning methods tailored to specific action spaces.
 
\section{Background}
\label{sec:background}

We consider discrete time reinforcement learning problems with large action spaces.
At time step $t$ the agent observes a state $s_t$ and produces an action $\Action_t$.
The agent then receives a reward $r_t$ and transitions to the next state $\State_{t+1}$ according to the environment transition distribution defined as $\EnvDist \delequal p(r_t, s_{t+1}|s_t, \Action_t)$.
The aim of the agent is to maximize the expected future discounted return $R_t = \sum_{t'=t}^\infty \gamma^{t'-t} r_{t'}$ where $\gamma \in [0,1]$ is a discount factor.
While in the standard Markov Decision Process formulation, the actions $\Action_t$ come from a finite set of possible actions $\mathcal{A}$, we consider a more general case.
Specifically, we assume that the action space is combinatorially structured, i.e.\ defined as a Cartesian product of sub-action spaces.
Formally we assume $\mathcal{A}=\mathcal{A}_1 \times \ldots \mathcal{A}_D$ where each $\mathcal{A}_i$ is either a finite set or $\mathcal{A}_i=[a,b] \subset \mathbb{R}$.
We therefore consider $\Action_t$ to be a $D$-dimensional vector and each component of $\Action_t$ is referred to as a sub-action.

Given a policy $\pi$ that maps states to distributions over actions, the action-value function for policy $\pi$ is defined as $Q_\pi(s, \Action) = \mathbb{E}_{\pi,\EnvDist}\left[R_t|s_t=s,\Action_t=\Action\right]$.
The optimal action-value function defined as $Q^*(s,\Action) = \max_{\pi} Q_\pi(s, \Action)$ gives the expected return for taking action $\Action$ in state $\State$ and acting optimally thereafter.
If $Q^*$ is known, an optimal deterministic policy $\pi^*$ can be obtained by acting greedily with respect to $Q^*$, i.e.\ taking $\pi^*(s) = \argmax_\Action Q^*(s,\Action)$.
An important property of $Q^*$ is that it can be decomposed recursively according to the Bellman equation
\begin{equation}
\label{eqn:bellman}
	Q^*(s,\Action) = \mathbb{E}_{\EnvDist }\left[r_t + \gamma \max_{\Action'} Q^*(s_{t+1}, \Action') |s_t=s, \Action_t=\Action \right]
\end{equation}
In value-based reinforcement learning methods the aim is to learn the optimal value function $Q^*$ by starting with a parametric estimate $Q(s, \Action; \theta)$ and iteratively improving the parameters $\theta$ based on experience sampled from the environment.

The Q-learning algorithm \citep{watkins1992q} is one such value-based method, which uses an iterative update based on the recursive relationship in the Bellman equation to learn $Q(s, \Action; \theta)$.
Specifically, Q-learning can be formalized as optimizing the loss
\begin{equation}
\label{eqn:qlearning}
	L(\theta) = \mathbb{E}_{\pi,\EnvDist} \left[ \left(r_t + \gamma \max_{\Action} Q(s_{t+1}, \Action; \overline{\theta}) - Q(s_t, \Action_t; \theta) \right)^2 \right],
\end{equation}
where $\overline{\theta}=\theta$ but $\overline{\theta}$ is treated as a constant for the purposes of optimization, i.e.\ no gradients flow through it.
 
\section{Related Work}
Deterministic policy gradient algorithms~\citep{hafner2009nfqca,silver2014deterministic,lillicrap2016continuous} learn a deterministic policy that maps states to continuous actions by following the gradient of an action-value function critic with respect to the action $\Action$.
\citet{silver2014deterministic} justified this approach by proving the deterministic policy gradient theorem, which shows how the gradient of an action-value function $Q_\pi$ with respect to the action $\Action$ at state $\State$ can be used to improve the policy $\pi$ at state $\State$.
As others have noted~\citep{haarnoja2017reinforcement}, these methods can be interpreted as Q-learning, with the deterministic policy serving as an approximate maximizer over the action space, partially explaining why such methods work well with off-policy data.
Taking this view makes deterministic policy gradient methods similar to our proposed method.
However, AQL learns an approximate maximizer using explicit search and supervised learning rather than using the gradient of the critic.
This lack of reliance on gradients with respect to actions renders AQL agnostic to the type of the action space, be it discrete, continuous or a combination of the two, while deterministic policy gradients are inapplicable to discrete action spaces.

\citet{gu2016continuous} address the intractability of $Q$-maximization in the continuous setting by restricting the form of $Q$ to be a state-dependent quadratic function of the continuous actions, rendering the maximization trivial.
While considerably simpler than DDPG, the practical consequences of this representational restriction in the continuous case are unclear, and the technique is inapplicable to discrete or hybrid action spaces.

\citet{metz2017discrete} apply Q-learning to multi-dimensional continuous action spaces by discretizing each continuous sub-action.
Their Sequential DQN approach avoids maximizing over a number of actions that is exponential in the number of action sub-actions $D$ by considering an extended MDP in which the agent chooses its sub-actions sequentially.
Sequential DQN learns $Q$ functions for the original MDP as well as the $D$ extended MDPs, using the latter to perform Q-learning backups for the former across environment transitions.
\citet{tavakoli2017} model Q-values for all sub-actions using a shared state value and a sub-action-specific advantage parametrization. Their best working parametrization subtracts the mean advantage for each action advantage dimension. The target Q-value is shared for all action dimensions and computed using the observed rewards and a bootstrapped value equal to the mean argmax action of the target network.
Notably, \citet{metz2017discrete} conditions the lower-level $Q$ function on sub-actions already taken as part of the current macro-action; we experiment with a similar strategy for our learned proposal distribution.

Another alternative to performing an exact maximization over actions is to replace this maximization with a fixed stochastic search procedure.
\citet{kalashnikov2018qt} and \citet{quillen2018grasping} used the cross-entropy method to perform an approximate maximization over actions in order to apply Q-learning to robotic grasping tasks.
This approach has been shown to work very well for tasks with up to 6-dimensional action spaces, but it is unclear if it can scale to higher-dimensional action spaces.
Notably, concurrent work\footnote{Versions of both~\cite{lim2019actor} and the present manuscript appeared concurrently at the NeurIPS 2018 Deep Reinforcement Learning Workshop.} of \citet{lim2019actor} develops an \emph{Actor-Expert} framework wherein an approximate $Q$ function is trained on continuous actions and a stochastic policy is updated with a state-conditional variant of the cross-entropy method.
Our work demonstrates that a simpler approach, rendered efficient by parallel computation of $Q$ values, can scale to even larger continuous action spaces, as well as high-dimensional discrete and hybrid action spaces.

On the subject of replacing policies with proposal distributions, concurrent work of \citet{hunt2018} learns proposal distributions in the stochastic case. Their proposal distribution consists of a mixture of truncated normal distributions that is learned by minimizing the forward KL divergence with the Boltzmann policy. This method is focused on sampling in continuous action spaces during transfer, and does not deal with discrete or hybrid action spaces.

Finally, we note also the concurrent work of \citet{wiehe2018sampled} applied the idea of learning an approximate maximizer over actions using search and supervised learning to continuous control.
Our work shows that this idea is more generally applicable and our experiments validate it on more challenging tasks.
\nopagebreak
 
\section{Amortized Q-Learning}
\label{sec:methods}
\SetKwBlock{With}{with}{}
\SetKwBlock{Otherwise}{otherwise}{}
\SetCommentSty{textrm}
\begin{algorithm}[t]
\DontPrintSemicolon
\Input{
	Proposal network parameters $\ProposalParams$,
	$Q$ network parameters $\QParams$,
	number of actions to draw from the proposal $\NumProposalActions$,
	number of actions to draw uniformly $\NumUniformActions$,
	unroll length $T$,
	exploration probability $\epsilon$

}
\Repeat{termination}{
	\For{$t \leftarrow 1\ldots\EpisodeLength$}{
		Observe state $\State_t$ \;
		$A_\Uniform := \{\UniformAction_j\}_{j=1}^M,\ 
		\UniformAction_j
		\sim
		\mathrm{Uniform}(\ActionSpace_1 \times \ldots \times \ActionSpace_D)$ \;
		$A_\Proposal := \{\ProposalAction_i\}_{i=1}^N,\ 
		\ProposalAction_i
		\sim
		\Proposal(\Action | \State_t;\ProposalParams)$ \;
		$\ActionBest_t := \argmax_{\Action \in A_\Uniform \cup A_\Proposal} Q(\State_t, \Action)$ \;
		\With(probability $\EpsilonProbability$,){
			$\Action_t := \UniformAction_1$ \tcp{Select $\Action_t$ uniformly at random} \;
		}
		\Otherwise{
			$\Action_t := \ActionBest_t$ \;
		}
		$\Reward_t, \State_{t+1} \sim p_E(\Reward_t, \State_{t+1} | \State_t, \Action_t)$
		\tcp{Take action $\Action_t$, receive reward $\Reward_t$} \;
	}
	Update $\QParams$ using $\StateTrajectory$, $\ActionTrajectory$, $\RewardTrajectory$ by descending the gradient of \eqref{eqn:aql-q-loss}. \;
	Update $\ProposalParams$ using $\StateTrajectory$ and $\ActionBestTrajectory$ by descending the gradient of \eqref{eq:proposal-loss}.
}
\caption{\textsc{AQL} \label{algo}}
\end{algorithm}
 Amortized inference \citep{dayan1995helmholtz} has revolutionized variational inference by eliminating the need for a costly maximization with respect to the parameters of the approximate variational posterior, making training latent variable models much more efficient.
Key to this approach is a form of \emph{amortization} in which an expensive procedure such as iterative optimization over the variational parameters is replaced with the more efficient forward pass in a parametric model, such as a neural network.
By training the model by optimizing the same objective as the original procedure, we obtain an efficient approximation to it that can be applied to new instances of the problem.
We propose to leverage the same principle in order to amortize the maximization of $Q(\State, \Action)$ with respect to the vector of sub-actions $\Action$.
Specifically, in addition to a neural network representing the $Q$ function (which takes the state $s$ and vector of sub-actions $\Action$ as inputs) we train an additional neural network to predict, from the state $s$, the sufficient statistics of a proposal distribution $\Proposal$ over possible actions.
We then replace exact maximization of $Q(\State, \Action)$ over actions $\Action$, which is required both for acting greedily and performing Q-learning updates, with maximization over a set of samples from the proposal $\Proposal$.

As long as the proposal distribution assigns non-zero probability to all actions, this method will approach Q-learning with exact maximization for discrete action spaces as the number of samples from the proposal increases\footnote{To see why this is true, let $p_{best}$ be the probability assigned to the action with the highest $Q$-value by the proposal.
Then after drawing $n$ samples from the proposal, the probability that the best action has been sampled at least once is $1-\left(1-p_{best}\right)^n$ which tends to 1 as $n$ tends to infinity.}.
Since the computational cost of the approach will grow linearly with the number of samples taken from the proposal, for practical reasons we are interested in learning a proposal from which we need to draw the fewest possible samples.

In this paper, we consider a simple approach to learning a proposal distribution which, at a high level, works as follows.
To update the proposal for state $\State$ we first perform an approximate stochastic search for the action with the highest $Q$-value at state $\State$.
We then update the proposal to predict the best action found by the search procedure, i.e.\ the one with the highest $Q$-value, using supervised learning to make the action found by the search procedure more probable under the proposal distribution.
Because the proposal is meant to track approximately maximal actions for a given state as prescribed by a parametric $Q$ function which is itself continually changing, we add a regularization term to the proposal objective in order to prevent the proposal from becoming deterministic and potentially becoming stuck in a bad local optimum.

More formally, we update the proposal distribution $\Proposal$ for a state $\State$, by drawing a set of $\NumUniformActions$ samples from the uniform distribution over all actions and a set of $\NumProposalActions$ samples from the current proposal for state $\State$.
Denoting the action with the highest $Q$-value from the union of these two sets as $\ActionBest(\State)$, the regularized loss for training the proposal distribution is given by
\begin{align}
\mathcal{L}(\ProposalParams; \State) = -\log \Proposal(\ActionBest(\State) | \State; \ProposalParams) - \lambda H(\Proposal(\Action | \State; \ProposalParams)).
	\label{eq:proposal-loss}
\end{align}
The first term in the objective is the negative log-likelihood for the proposal with $\ActionBest$ as the target.
Minimizing it makes the sample with the highest $Q$ value more likely under the proposal.
The second term is the negative entropy of the proposal distribution, which encourages
uncertainty in $\Proposal$ throughout training and prevents it from collapsing to a deterministic distribution.

\begin{figure}[t]
\centering
\includegraphics[width=2.5in]{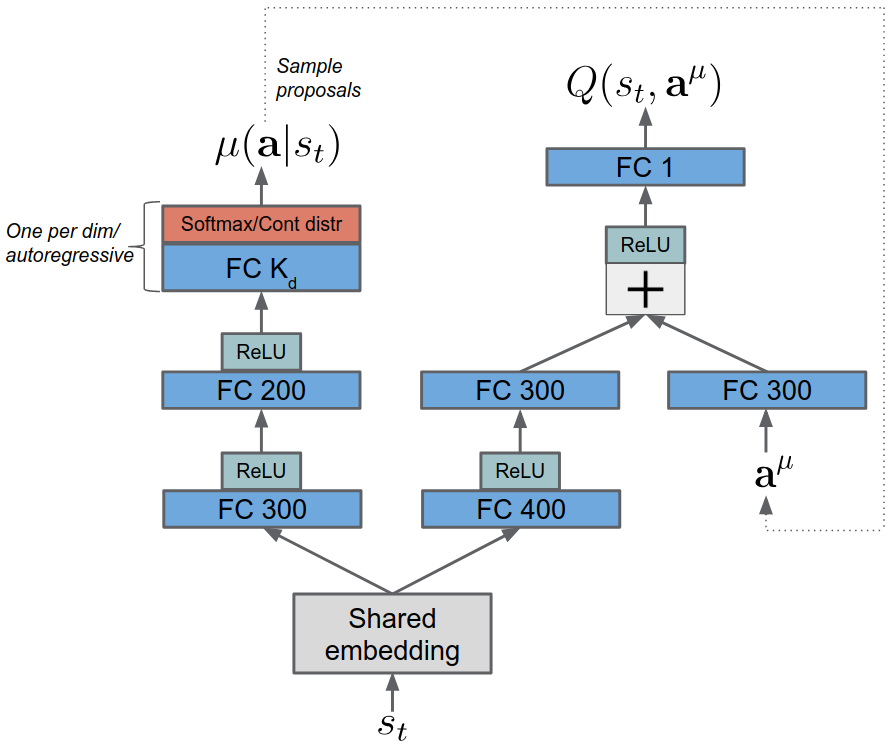}
\includegraphics[width=0.4in]{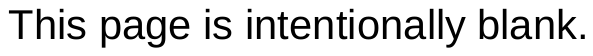}
\includegraphics[width=2.5in]{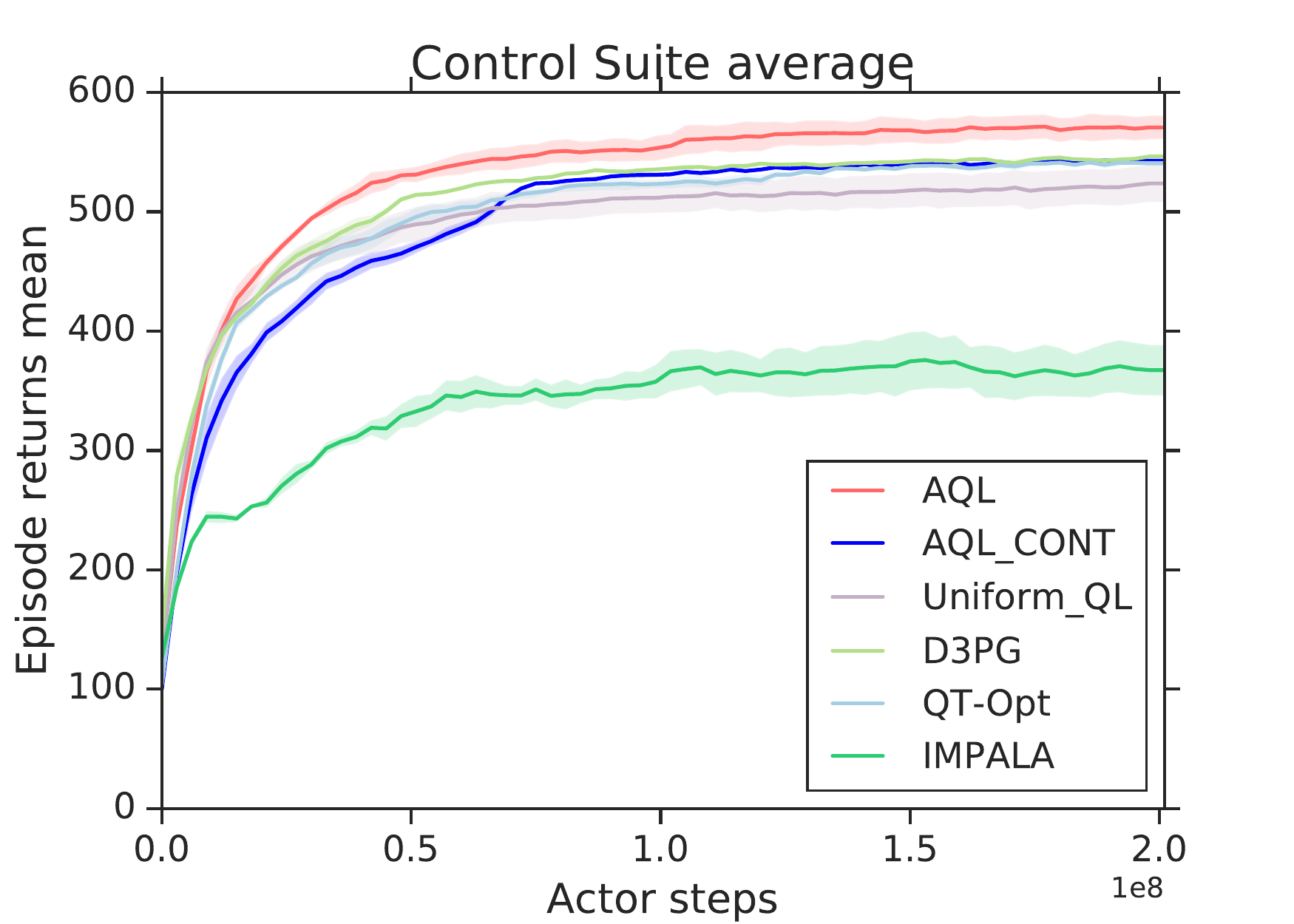}
\caption{\textbf{Left:} The AQL architecture used in the experiments.
	The shared state embedding network is the identity function for the Control Suite experiments (i.e.\ the proposal network and $Q$ network each operate directly on the observations and share no parameters).
	For DeepMind Lab experiments, where the input consists of pixel observations, the shared state embedding consists of 3 layers of ResNet blocks followed by a fully connected layer with 256 units and a recurrent LSTM core with 256 cells as in \citet{espeholt2018impala}.
	The left head represents the proposal distribution.
There is a proposal output layer of dimension $K_d$ for each sub-action $d$.
$K_d$ represents the number of choices for dimension $d$ for discrete or discretized continuous actions and the number of distribution parameters for continuous actions.
    Samples from the proposal distribution are embedded and concatenated with an embedding of the state and used to compute $Q(\State_t, \Action^{\Proposal})$. \textbf{Right:} Learning curves of the mean return across all tasks in the Control Suite.
The error bars represent the standard error of the mean episode return.}
\label{fig:architecture}
\end{figure}

Finally, we can define the AQL loss for updating the action-value function as
\begin{align}
	& \mathcal{L}(\QParams) = \mathbb{E}_{\pi,\EnvDist} \left[ \left(r_t + \gamma Q(s_{t+1}, \ActionBest(s_{t+1})) - Q(s_t, \Action_t; \QParams) \right)^2 \right], \label{eqn:aql-q-loss}
\end{align}
which is identical to the Q-learning loss but with maximization over all actions replaced by the stochastic search procedure based on the proposal distribution.
Pseudocode for the complete Amortized Q-learning algorithm is shown in Algorithm~\ref{algo}.
While the algorithm considers the action-value and proposal parameters separate, we consider the possibility of sharing some of the parameters between the two following standard practice.
The parameter sharing is made clear in the network architecture diagram in Figure~\ref{fig:architecture}, which details the architecture used for all experiments.

The AQL algorithm can be applied to continuous, discrete, and hybrid discrete/continuous action spaces simply by changing the form of the proposal distribution $\Proposal$.
We use autoregressive proposals of the form
\begin{align*}
\Proposal(\Action | \State; \ProposalParams) = \textstyle \prod_{d=1}^D \Proposal_d(a_d|\State, \Action_{<d}; \ProposalParams),
\end{align*}
in order to incorporate the dependencies among sub-actions.
For discrete sub-actions we parameterized the proposal $\mu_d$ using a softmax.
We experimented with discretized and continuous proposal distributions for continuous sub-actions.
Proposal distributions for continuous sub-actions are either parameterized using a softmax over a set of discrete choices representing uniformly-spaced values from the continuous sub-action's range or using a Gaussian distribution with fixed variance.
A detailed description of the architecture is included in the Appendix.

\section{Experiments}
\label{sec:experiments}
We performed experiments on two families of tasks: the set of DeepMind Control Suite~\citep{tassa2018deepmind} tasks with constrained actions (16 distinct domains, with a total of 39 tasks spread between them) and two tasks in the DeepMind Lab~\citep{beattie2016deepmind} 3D first-person environment.
Descriptions of the architectures and hyperparameter settings are available in the Appendix.

\subsection{Control Suite}
\label{sec:csexp}
The DeepMind Control Suite is a collection of continuous control tasks built on the MuJoCo physics engine~\citep{todorov2012mujoco}.
The Control Suite is considered a good benchmark as it spans a wide range of tasks of varying action complexity.
In the simplest tasks, the agent controls only a single actuator while the \texttt{humanoid} tasks require selection of at least 21 sub-actions at every time step. As is common, we employed the underlying state variables rather than visual representations of the state as observations, as the focus of our inquiry is the complexity of the action space rather than the observation space.
We explore a domain where the observations are pixel renderings in Section~\ref{sec:dmlab}.

We compared AQL to several baseline continuous control methods.
First, we considered an ablation which replaces the learned proposal distribution with a fixed uniform proposal distribution.
We dub this method \textbf{Uniform Q-learning}.
We also considered two strong continuous control methods: D3PG~\citep{barth2018distributed}, a distributed version of DDPG~\citep{lillicrap2016continuous}, and QT-Opt~\citep{kalashnikov2018qt} which corresponds to Q-learning where action maximization is performed using the cross-entropy method (CEM)~\citep{kalashnikov2018qt,rubinstein2004}.
CEM is a derivative-free iterative optimization algorithm that samples $K$ values at each iteration, fits a Gaussian distribution to the best $L<K$ of these samples, and then samples the next batch of $K$ from that distribution.
The iterative process is initialized with uniform samples from the action space.
In our implementation, we used the same number of initial samples as the number of proposal actions $K = \NumProposalActions = 100$ and $L=10$, performed three iterations and fit independent Gaussian distributions for each sub-action to the $L$ actions with the highest corresponding $Q$-values. Following previous work~\citep{kalashnikov2018qt,hafner2018Planet}, $L$ is chosen to be an order of magnitude smaller than $K$.
The final baseline method consisted of the IMPALA agent~\citep{espeholt2018impala}, an importance weighted advantage actor-critic method that is inspired by A3C~\citep{mnih2016asynchronous}.

Our AQL implementation used an autoregressive proposal distribution which models sub-actions sequentially using the ordering in which they appear in the Control Suite task specifications.
We considered both discretized categorical and Gaussian action distributions, the latter with fixed variance $\sigma^2 = 0.25$. The logits (in the discretized case) or distribution means (continuous case) of the proposal distribution for sub-action $i$ are computed by a linear transformation of a concatenation of features derived from the state and one-hot encodings of all previously sampled sub-actions $1, 2, \ldots, i-1$. Note that entire composite actions sampled from the proposal distribution remain independent and identically distributed.

For the discretized AQL method and the Uniform Q-learning method, actions were discretized for each degree of freedom to one of five sub-actions in $\{-1, -0.5, 0, 0.5, 1\}$.
This yields a total of $5^D$ distinct actions.
All methods except D3PG used $\NumProposalActions = 100$ proposal samples and $\NumUniformActions = 400$ uniform search samples as described in Algorithm~\ref{algo}.
We considered more uniform search samples since they are less expensive to compute, as they do not require the relatively expensive autoregressive sampling procedure.
The $Q$-value evaluations were implemented as a single batched operation for each actor step, which rendered evaluation highly efficient.
Following the literature, our D3PG experiments made use of an Ornstein-Uhlenbeck process~\citep{uhlenbeck1930} for exploration with $\theta=0.15$ and $\sigma = 0.2$.
We employed the Adam optimizer~\citep{adamoptimizer2014} to train all models.
In order fairly compare across methods, all architectural details that are not specific to the method under consideration were held fixed across conditions.
Further details are included in the Appendix.

Figure~\ref{fig:architecture} depicts the learning curves for 200 million actor steps averaged over all tasks (task specific learning curves are reported in the Appendix), which shows that discretized AQL performs best on average.
Returns at a given point in training were obtained by allowing the agent to act in the environment for 1000 time steps and summing the rewards.
Rewards range between 0 and 1 for all tasks in the Control Suite, which makes the average rewards comparable between tasks.
While not shown in Figure~\ref{fig:architecture} we note that all methods also substantially outperform the A3C results reported in~\citep{tassa2018deepmind} except for the IMPALA results which are only marginally better.

Closer inspection revealed that the difference between the methods in Figure~\ref{fig:architecture} was mostly explained by the performance on the medium- and high-dimensional tasks.
Exploration in high-dimensional action spaces is typically a harder problem, and the same appears to hold for searching for the action with the maximum $Q$-value.
Analysis of tasks with high-dimensional action spaces revealed that learning a discretized proposal distribution (AQL) or a deterministic policy (D3PG) clearly outperformed Uniform Q-learning and QT-Opt, both of which use fixed stochastic search procedures for action selection.
IMPALA performed worse than the other methods on most Control Suite tasks.

\begin{figure}[t]
\centering
\includegraphics[width=2.6in]{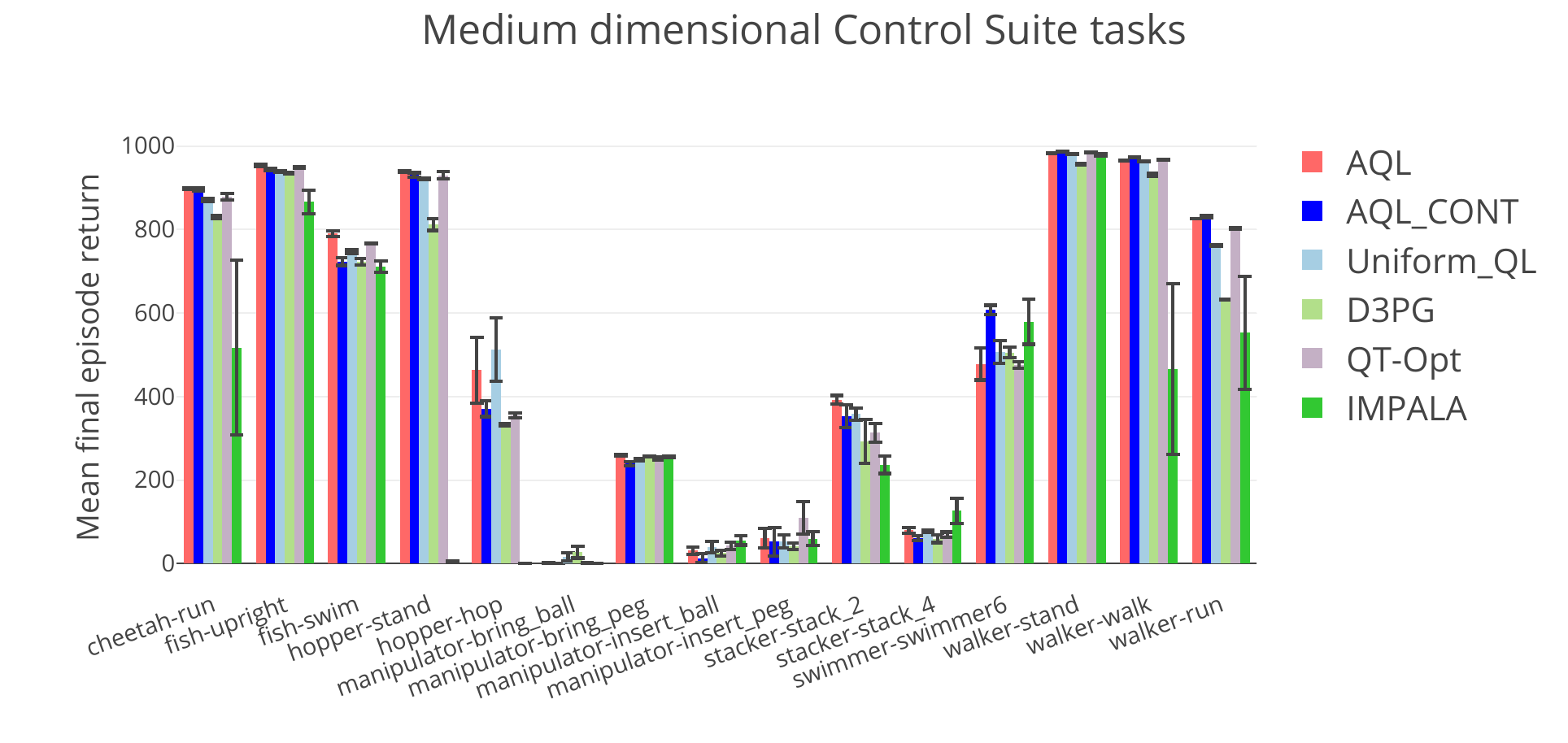}
\includegraphics[width=2.6in]{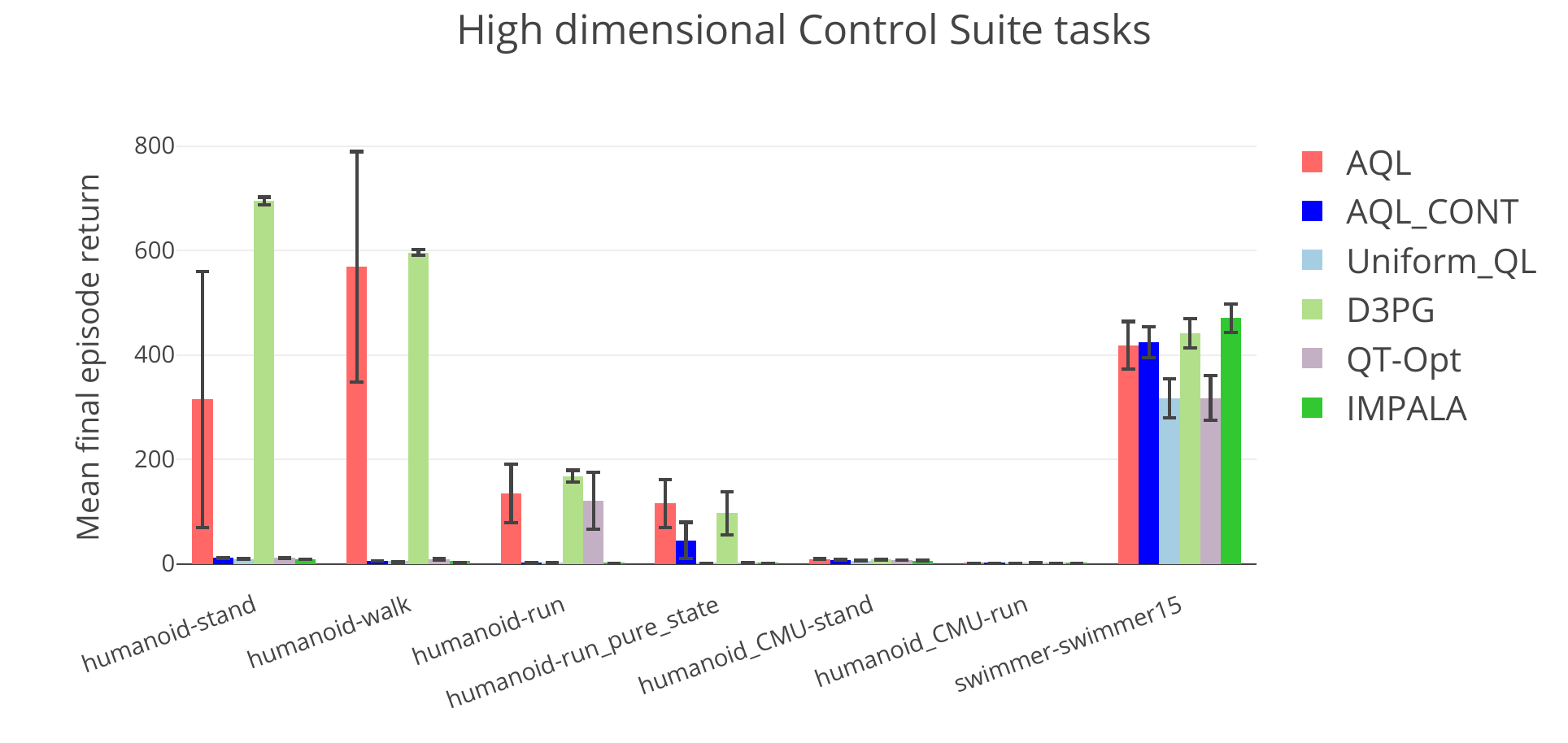}
\caption{Mean final performance, averaged over 3 seeds, for the medium- and high-dimensional Control Suite tasks.
The error bars represent the standard error of the mean episode return.}
\label{fig:control_suite_medium_high_dim_drop_indep}
\end{figure}

The continuous implementation of AQL performed similarly on the low- and medium-dimensional tasks as the discretized AQL variant but failed to learn on the high-dimensional tasks.
This may be related to the observation that exploration is often easier when the action space is discretized~\citep{metz2017discrete,peng2017learning}.
Figure~\ref{fig:control_suite_medium_high_dim_drop_indep} shows the average mean final performance for the medium- and high-dimensional tasks.
Results on the low-dimensional Control Suite tasks are available in the Appendix and show less drastic differences between the methods.
These experimental results support the hypothesis that learning a proposal distribution is beneficial for more complex or higher-dimensional action spaces.

Overall, these results demonstrate that AQL can scale up to problems with a relatively high-dimensional action space.
The 21-dimensional \texttt{humanoid} tasks are of particular interest: although our discretization scheme leads to a total of $5^{21}$ possible discrete actions on these tasks, AQL was able to learn policies competitive with D3PG\@.

Uniform Q-learning and QT-Opt, both of which correspond to Q-learning with a fixed procedure for maximizing over actions, largely failed to learn on the \texttt{humanoid tasks}, demonstrating the advantage of learning a state-dependent maximization procedure in high-dimensional action spaces.

\subsection{DeepMind Lab}
\label{sec:dmlab}
DeepMind Lab~\citep{beattie2016deepmind} is a platform for 3D first person reinforcement learning environments.
We evaluated AQL on DeepMind Lab because it offers a combination of a rich observation space (we consider $84\times84$ RGB observations) and a complex, structured action space.

The full action space consists of 7 discrete sub-actions which can be selected independently.
The first two sub-actions specify the angular velocity of the viewport rotation (left/right and up/down).
Each of these represents an integer between -512 and 512.
The next two sub-actions represent sets of three mutually exclusive movement-related options: strafe left/right/no-op and forward/backward/no-op.
The last three sub-actions are binary: fire/no-op, jump/no-op and crouch/no-op.
We considered two action sets: a curated set with 11 actions based on the minimal subset that allows good performance on all tasks, and a second, larger action set with 3528 actions consisting of the Cartesian product of 7 rotation actions for each of the two rotational axes and the 5 remaining ternary and binary sub-actions ($3528 = 7^2 \cdot 3^2 \cdot 2^3$).

We compared three methods on each of the two action sets: our distributed implementation of AQL with $\NumProposalActions = 100$ proposal samples and $\NumUniformActions = 500$ uniform search samples, an ablation that performs exact Q-learning (i.e.\ evaluates the Q-values for all possible discrete actions at every time step) and the IMPALA method.
As with our AQL implementation, the Q-learning ablation agent was distributed, replay-based, and used a recurrent Q-function trained with $Q(\lambda)$ making it a strong baseline similar to R2D2~\citep{kapturowski2019recurrent}.
Other details such as network architecture and optimizer choice were held fixed across conditions.
The AQL implementation for the large action set incorporated the 7-dimensional structure of the discrete action space and used the same style of autoregressive proposal as the Control Suite experiments.
Every action was repeated 4 times and each seed was run for 200 million environment steps (50 million actor steps given the repeated actions).
The precise network architecture is described in the Appendix.

\begin{figure*}[h]
\centering
\includegraphics[width=2.0in]{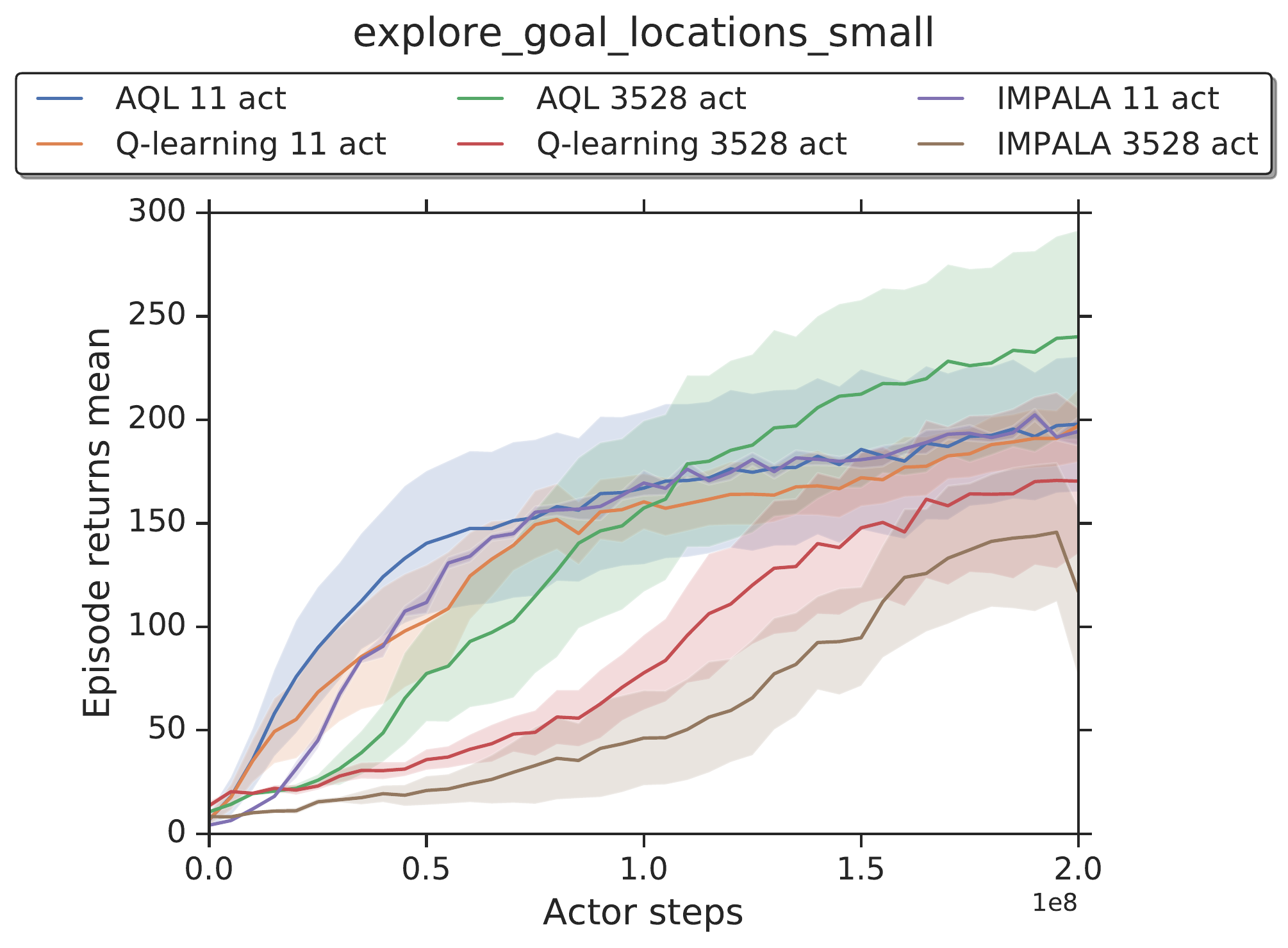}
\hspace*{2cm}\includegraphics[width=2.0in]{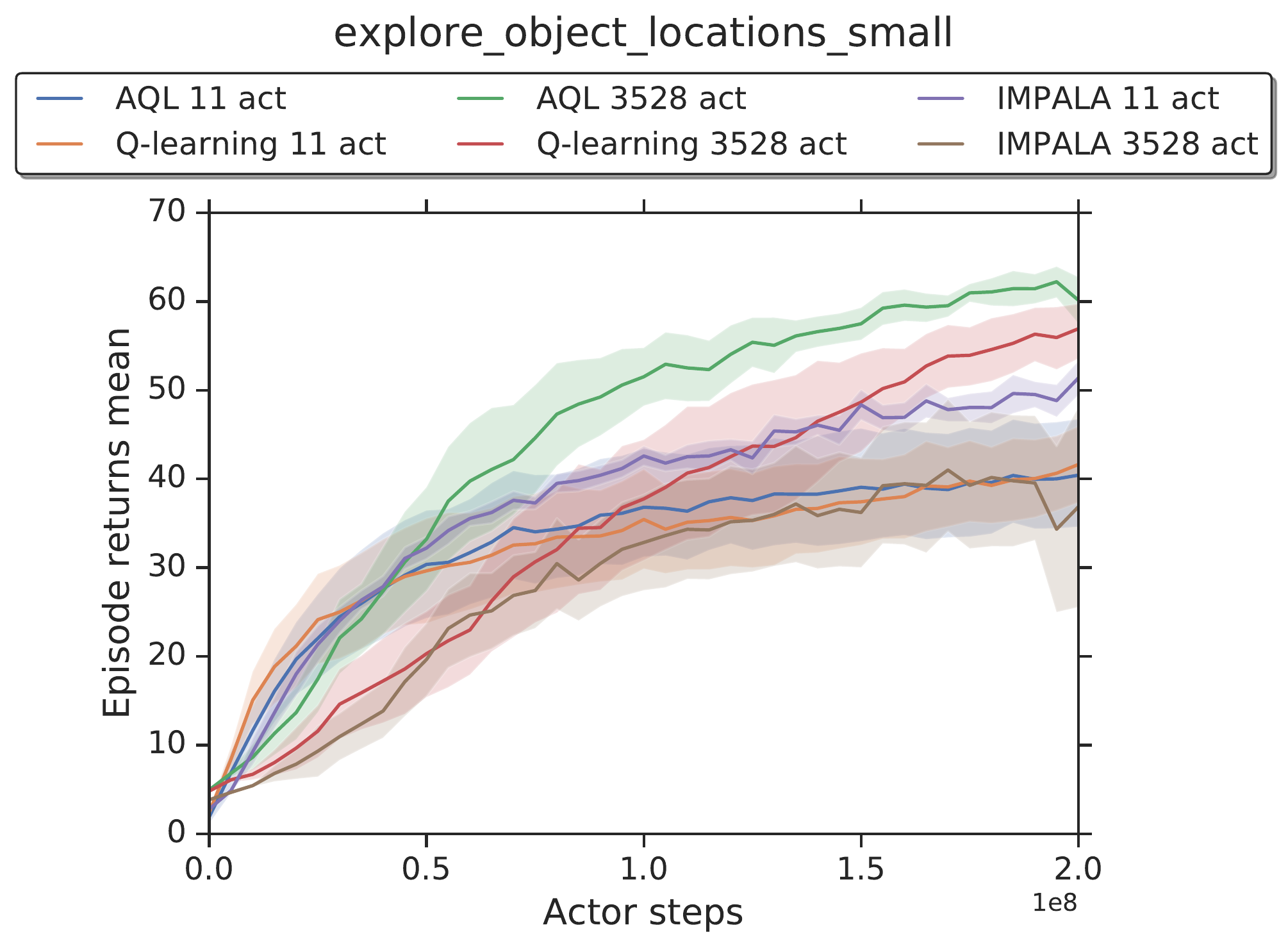}\\
\includegraphics[width=2.0in]{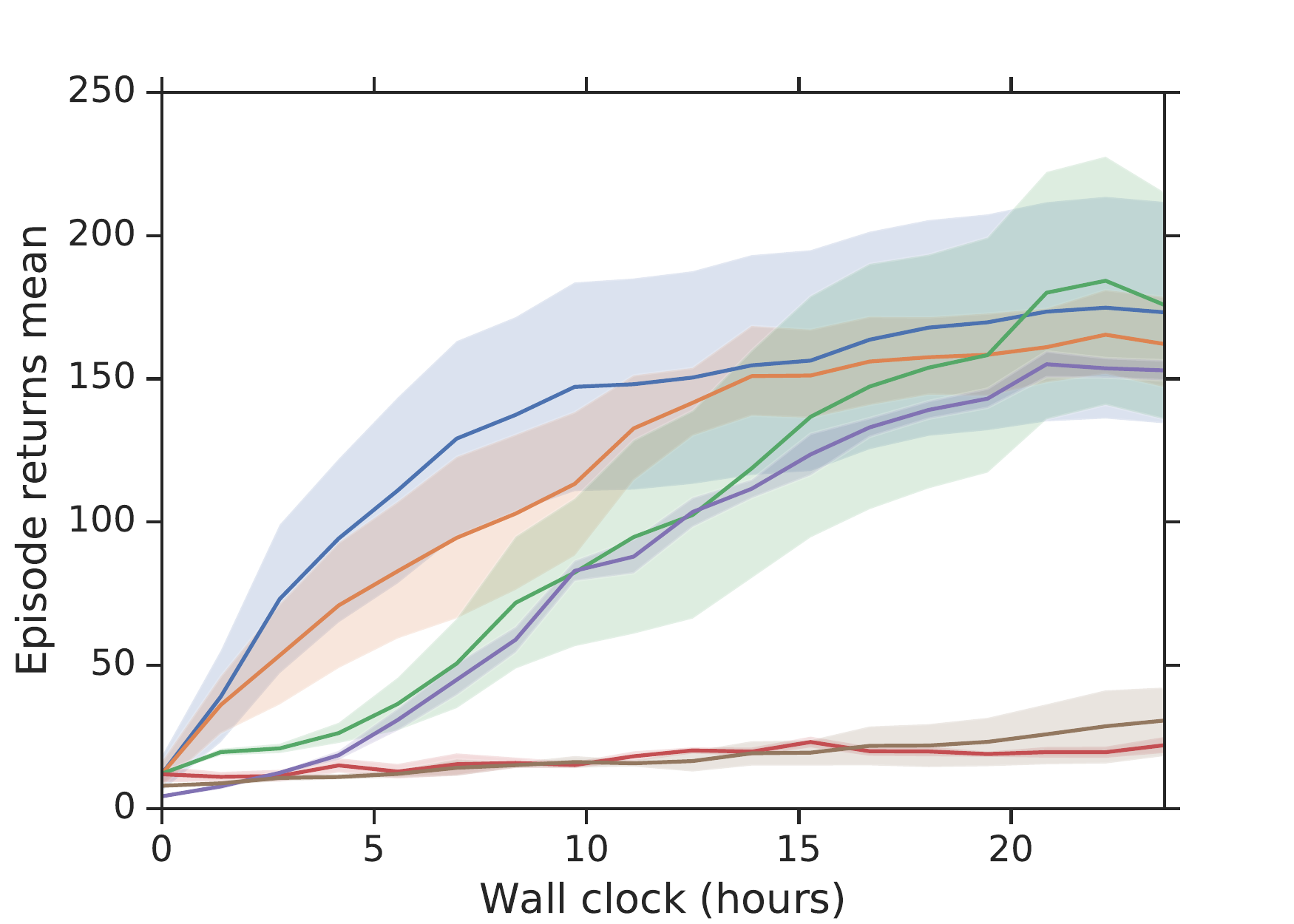}
\hspace*{2cm}\includegraphics[width=2.0in]{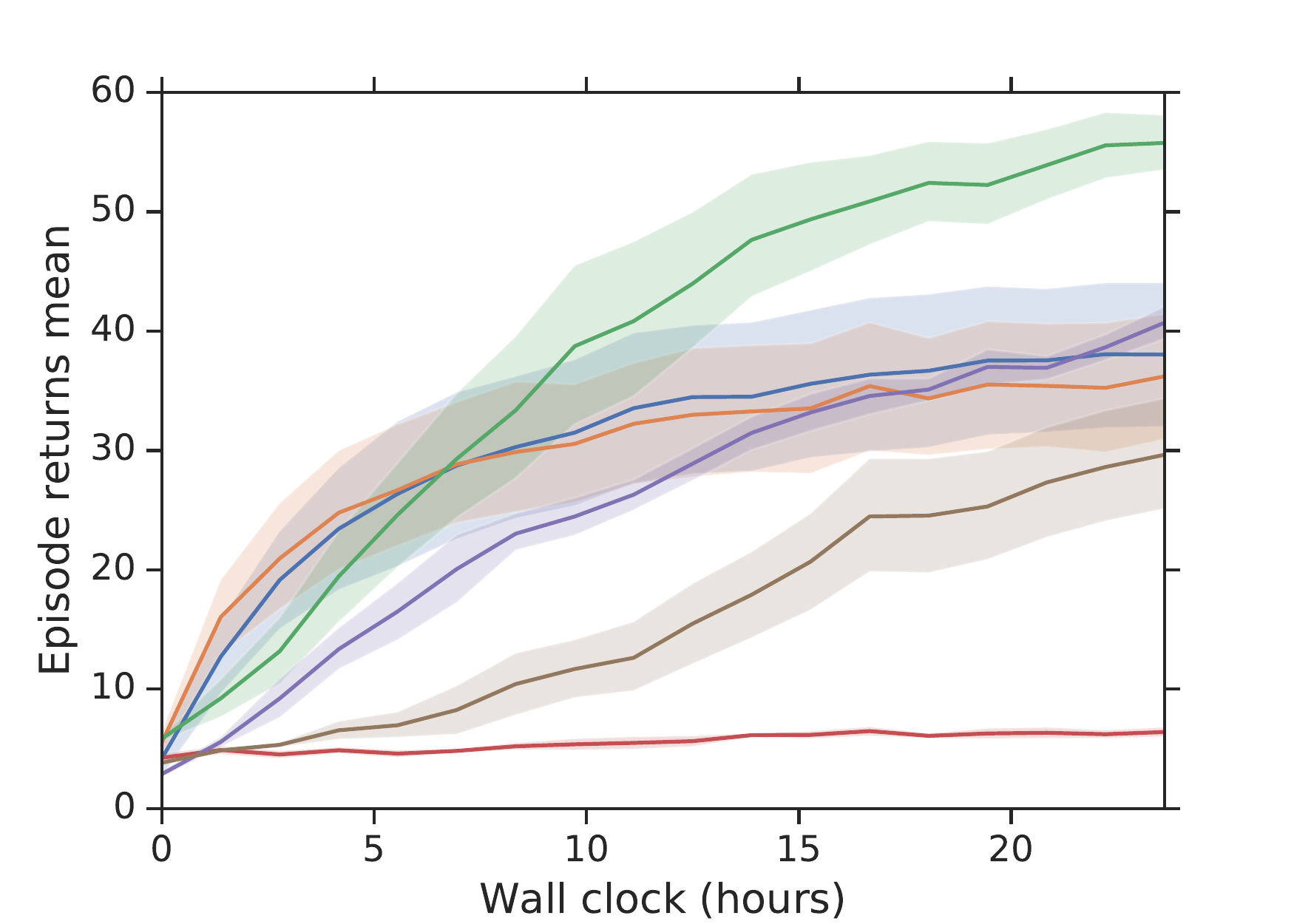}
\caption{Learning curves on two DeepMind Lab exploration tasks.
The error bars represent the standard error of the mean episode return over 5 seeds.
The top row shows the mean episode return versus the number of actor steps and the bottom row represent the mean episode return versus time for the first 24 hours of the experiment.}
\label{fig:explore_locations_small}
\end{figure*}

Figure~\ref{fig:explore_locations_small} shows the results for two exploration tasks from DeepMind Lab.
In both tasks the agent was rewarded for navigating to the object(s) in a maze. The \texttt{explore\_goal} task only contains a single object with goal location(s), level layout and theme randomized per episode.
As expected, there was no significant performance difference between AQL and exact Q-learning (i.e. computing Q-values for all 11 actions) when using the curated action set.
IMPALA also performs comparably with the curated action set.
When training with the large action set, exploration is harder, which explains why the initial performance was worse for all methods.
AQL achieved the best final score on both tasks, eventually outperforming IMPALA, exact Q-learning with the curated action set and exact Q-learning on the large action set.
The agent trained with AQL on the large action set was able to outperform agents trained with the curated action set by navigating more efficiently (e.g.\ by simultaneously strafing and moving forward).
While it may be surprising that AQL achieves better data efficiency than Q-learning on the large action set (where Q-learning evaluates all possible actions), we speculate that AQL benefits from the additional stochasticity in action selection due to approximate maximization, which potentially improves exploration.
It is also possible that by using approximate maximization for bootstrapping, AQL is less prone to over-estimation of Q-values, from which Q-learning with exact maximization is known to suffer~\citep{hasselt2010double}; we leave this investigation to future research.
Critically, considering all 3528 actions with exact Q-learning takes about 10 times longer in wall-clock time because the Q-function must be evaluated for each action.
IMPALA is also slowed down drastically when considering 3528 actions. These experiments show that AQL can indeed efficiently learn good policies in large discrete structured action spaces and, unsurprisingly, performs comparably to exact Q-learning in low cardinality action spaces.
 
\section{Discussion}
\label{sec:discussion}
We presented AQL, a simple approach to scaling up Q-learning to multi-dimensional action spaces where the sub-actions can be discrete, continuous or a combination of the two.
We showed that AQL is competitive with several strong continuous-control methods on low-dimensional control tasks, and is able to outperform them on medium- and high-dimensional action spaces.
Perhaps most notably, AQL closes the gap between Q-learning and stochastic actor-critic methods in their ability to handle high-dimensional and structured action spaces.
While actor-critic methods have been able to handle such action spaces simply by changing the form of the stochastic policy~\citep{schulman2015trust,mnih2016asynchronous}, different variants of Q-learning have been developed to handle each specific case, i.e. DDPG for continuous actions and DQN for low-dimensional discrete action spaces.
AQL allows one to handle different action spaces simply by varying the form of the proposal distribution while keeping the rest of the algorithm unchanged.
While AQL and stochastic actor-critics are equally general in terms of which action spaces they can handle (being limited only by the ability to efficiently sample from a distribution over actions), AQL may have some advantages over actor-critic methods.
Namely, AQL supports off-policy learning without the need for importance sampling-based correction terms used by actor-critic methods, making the method easier to implement in replay-based agents.

There are several promising avenues for future work.
AQL could be combined with the cross-entropy method (CEM) by modeling the parameters of the initial distribution of CEM with the proposal distribution.
Another interesting direction would be to use the proposal distribution for more intelligent exploration than epsilon-greedy exploration.
Currently AQL uses a simple way of learning the proposal based on supervised learning.
Better ways of optimizing the proposal parameters could improve the overall algorithm.
For example, one natural choice would be to train the proposal by maximizing the expected $Q$-value of a sample from it using REINFORCE~\citep{williams1992simple}.
Considering sequential proposal distributions is another interesting possibility.
By conditioning each sample from the proposal on all previous samples it may be possible to learn a proposal that achieves the same overall performance using many fewer samples.
  \subsubsection*{Acknowledgements}

We thank Marlos C. Machado, Jonathan Hunt and Tim Harley for invaluable feedback on drafts of the manuscript.
We additionally thank Jonathan Hunt, Benigno Uria, Siddhant Jayakumar and Catalin Ionescu for helpful discussions. Additional credit goes to Wojtek Czarnecki for creating the DeepMind Lab action set with 3528 actions.
 
\bibliography{paper}
\bibliographystyle{iclr2020_conference}

\appendix
\section*{Appendix}
\renewcommand{\thesubsection}{A\arabic{subsection}}
\subsection{Architecture Details}
\label{sec:arch_details}
We employ a distributed reinforcement learning architecture inspired by the IMPALA reinforcement learning architecture~\citep{espeholt2018impala}.
There is one centralized GPU learner batching parameter updates on experience collected by a large number of CPU-based parallel actors.
In all our experiments we use 100 parallel actors.
All experiments employ the architecture detailed in the main paper.

The state embedding network consists of the identity function for the Control Suite, so there is no parameter sharing between the proposal and the value networks.
The state embedding network for DeepMind Lab experiments is set to the convolutional network employed in \citet{espeholt2018impala}: 3 layers of ResNet blocks followed by a fully connected layer with 256 units and a recurrent LSTM core with 256 cells.
The weights of the network are optimized with two optimizers: one for the proposal distribution head and one for the shared state embedding network and $Q$-value head.
The parameters of the shared state embedding network are considered constant when optimizing the proposal loss.
The actors use a local FIFO replay buffer from which unrolls are sampled without prioritization.

\subsection{Experimental Details}
\label{sec:experiment_details}
The actors send the current trajectory to the learner queue at the end of an unroll along with two samples from the local replay buffer.
Each local replay buffer can store $10^5$ steps, representing a total effective replay size of $10^7$ since 100 parallel actors were used.
The unroll length is set to 30 for Control Suite experiments and to 100 for DeepMind Lab experiments.
We train both optimizers with Adam~\citep{adamoptimizer2014} with a learning rate of $2 \cdot 10^{-4}$, default TensorFlow hyperparameters and mini-batches of size  $32$.
We could use a target $Q$-network as described in the main text, but we do not since we did not observe significant gain in any of our hyperparameter iterations. 
The discount factor $\gamma$ is set to $0.99$. We use the $Q(\lambda)$ variant due to \citet{peng1994incremental} (also Chapter 7, \citet{sutton1998book}) to compute $Q$-targets with $\lambda=0.8$.
Following \citet{mnih2016asynchronous} and \citet{horgan2018distributed}, we use a different amount of $\epsilon$-greedy exploration for each actor, as this has been shown to improve exploration.
The first 10 actors use $\epsilon=0.5$ and the remaining actors use $\epsilon = 0.05$.

\SetKwBlock{With}{with}{}
\SetKwBlock{Otherwise}{otherwise}{}
\SetKwProg{Proc}{procedure}{}{}
\SetCommentSty{textrm}
\begin{algorithm}[h]
\DontPrintSemicolon
\Proc(
){\textsc{Actor}}{
\Input{Proposal network parameters $\ProposalParams$, \\
	$Q$ network parameters $\QParams$, \\
	number of actions to draw from the proposal $\NumProposalActions$, \\
	number of actions to draw uniformly $\NumUniformActions$, \\
	unroll length $T$, \\
	number of replay steps $R$, \\
	exploration probability $\epsilon$ \\
	}
\;
Initialize local replay buffer $\ReplayBuffer$. \;
\Repeat{termination}{
		\For{$t \leftarrow 1\ldots\EpisodeLength$}{
	        Observe state $\State_t$ \;
                $A_\Uniform := \{\UniformAction_j\}_{j=1}^M,\
                \UniformAction_j
                \sim
                \mathrm{Uniform}(\ActionSpace_1 \times \ldots \times \ActionSpace_D)$ \;
                $A_\Proposal := \{\ProposalAction_i\}_{i=1}^N,\
                \ProposalAction_i
                \sim
                \Proposal(\Action | \State_t;\ProposalParams)$ \;
                $\ActionBest_t := \argmax_{\Action \in A_\Uniform \cup A_\Proposal} Q(\State_t, \Action; \QParams)$ \;
                \With(probability $\EpsilonProbability$,){
                        $\Action_t := \UniformAction_1$ \tcp{Select $\Action_t$ uniformly at random} \;
                }
                \Otherwise{
                        $\Action_t := \ActionBest_t$ \;
                }
                $\Reward_t, \State_{t+1} \sim p_E(\Reward_t, \State_{t+1} | \State_t, \Action_t)$
                \tcp{Take action $\Action_t$, receive reward $\Reward_t$} \;

		}
		Send unroll $(\StateTrajectory, \ActionTrajectory, \ActionBestTrajectory, \RewardTrajectory)$ to the learner.\;
		Add unroll $(\StateTrajectory, \ActionTrajectory, \ActionBestTrajectory, \RewardTrajectory)$ to $\ReplayBuffer$. \;
		\For{$i \leftarrow 1 \ldots R$}{
			Sample unroll $(\StateTrajectory, \ActionTrajectory, \ActionBestTrajectory, \RewardTrajectory)$ from $\ReplayBuffer$. \;
			Send unroll $(\StateTrajectory, \ActionTrajectory, \ActionBestTrajectory, \RewardTrajectory)$ to the learner.\;

		}

		Poll the learner periodically for updated values of
		$\ProposalParams$, $\QParams$.
\;
		Reset the environment if the episode has terminated.
	}
}
\;
\Proc(
){\textsc{Learner}}{
	\Input{Batch size $\BatchSize$}
	\Repeat{termination}{
		Assemble batch of experience $\ExperienceBatch = \{
			(\StateTrajectory^\BatchIndex, \ActionTrajectory^\BatchIndex, {\ActionBestTrajectory}^\BatchIndex, \RewardTrajectory^\BatchIndex)\}_{\BatchIndex=1}^\BatchSize$ \;
Update $\QParams$ with a step of gradient descent on \\
		$\mathcal{L}(\QParams) = \sum_{b=1}^{B} \sum_{t=1}^{T-1} \left[ \left(r_t^b + \gamma Q(s^b_{t+1}, \ActionBestBatch_{t+1}; \overline{\QParams}) - Q(s_t^b, \Action_t^b; \QParams) \right)^2 \right] $ \;
		Update $\ProposalParams$ with a step of gradient descent on \\
$\mathcal{L}(\ProposalParams) = \sum_{b=1}^{B} \sum_{t=1}^{T} \left[-\log \Proposal(\ActionBestBatch_t(\State^b_t) | \State^b_t; \ProposalParams) - \lambda H(\Proposal(\Action^b_t | \State^b_t; \ProposalParams)) \right]$ \;
		Periodically set $\overline{\QParams} = \QParams$
	}
}
\caption{\textsc{AQL (Distributed)} \label{algo}}
\end{algorithm}
 
\subsection{Control Suite}
\label{sec:control_suite}
\begin{figure}[t]
\centering
\includegraphics[width=3.6in]{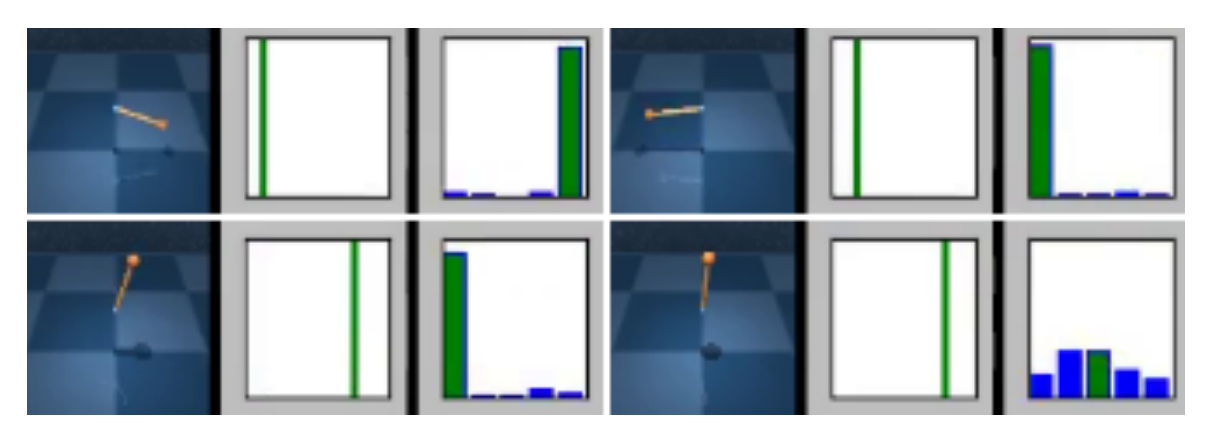}
\caption{Chronological visualization of the proposal distribution for the \texttt{pendulum-swingup} task. The top two images show the pendulum during the swingup phase and the bottom two images visualize the balancing phase. The leftmost part of each plot shows the behavior of the agent. The middle part plots a histogram of Q-values for 100 sampled actions from the proposal distribution with an x-axis that ranges from 0 to 100 (maximum possible value since $\gamma=0.99$). The right part shows the probabilities of the 5 discretized action options for the proposal distribution. Green bars in the proposal distribution belong to the argmax action and also the selected action since the agent follows the greedy policy. The proposal distribution is near-deterministic while swinging up and near-uniform when it gets close to the balancing equilibrium.}
\label{fig:pendulum_proposal_cropped}
\end{figure}

We have uploaded a video of the final performance of the discretized AQL agent for all Control Suite tasks at \url{https://youtu.be/WgTXjJhe6iQ}. The video shows the behavior of the greedy policy along with the proposal distribution and the Q-values of the sampled proposal actions. The videos are selected by picking the seed with the best performance after training.
Closer inspection of the trained proposal distributions of the AQL agent reveals that the sub-action proposal distributions have a general tendency to alternate between near-deterministic, low-entropy distributions at critical decision times and high-entropy distributions. The proposal distributions typically still have high entropy when the task is solved and there is a clear optimal action to maintain the equilibrium. Examples of this behavior can be found in the \texttt{finger}, \texttt{ball-in-cup}, \texttt{hopper}, \texttt{humanoid}, and \texttt{reacher} tasks. One notable example of the side effect of having high entropy in the proposal distribution can be seen in the \texttt{walker-stand} task. The high entropy of the policy results in alternating balancing and walking behavior.
Figure~\ref{fig:pendulum_proposal_cropped} visualizes the proposal distribution for the \texttt{pendulum-swingup} task. The proposal distribution has low entropy when swinging the pendulum up and low entropy during the balancing stage of the task.

\begin{figure}[h]
\centering
\includegraphics[width=3.5in]{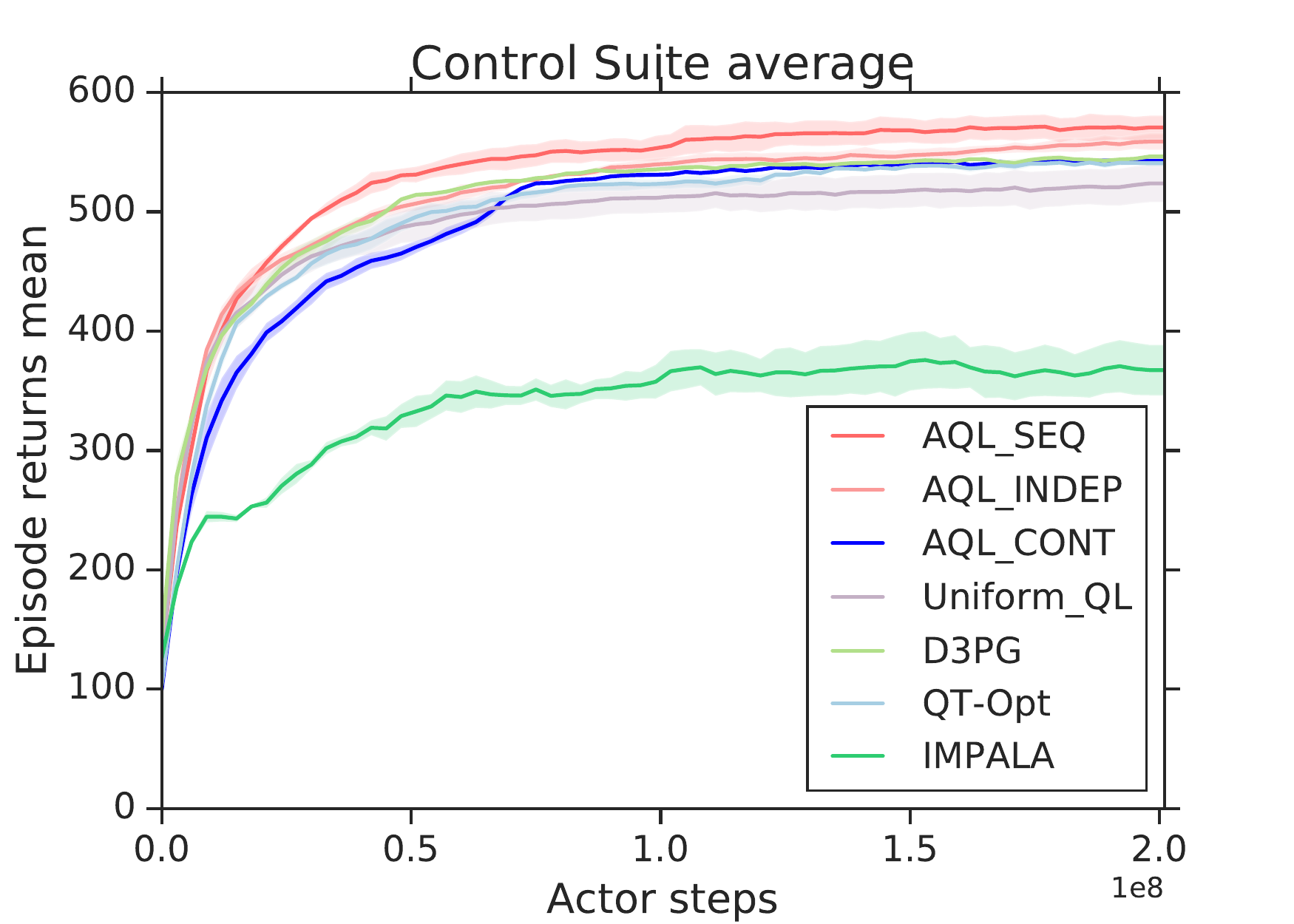}
\caption{Learning curves of the mean return across all tasks in the Control Suite.
The error bars represent the standard error of the mean episode return over 3 seeds.
The AQL variant where sub-actions are sampled independently performs slightly worse on average.}
\label{fig:control_suite_average}
\end{figure}

We also considered a simpler AQL method on the Control Suite.
The \textit{independent} AQL method models the different sub-actions as conditionally independent given the state as opposed to the variant from the main text where an order over sub-actions is assumed and each sub-action is conditioned on the state as well as preceding sub-actions. Figure~\ref{fig:control_suite_average} shows that the independent, discretized variant performed only slightly worse on average on the Control Suite. The AQL implementation with a continuous proposal distribution used an autoregressive policy.
We have also uploaded a video of the final performance of the independent proposal AQL agent for all Control Suite tasks on \url{https://youtu.be/9YIujaHjsQY}. The video shows the behavior along with the proposal distribution and the Q-values of the sampled proposal actions.

\begin{figure}[h]
\centering
\includegraphics[width=5.5in]{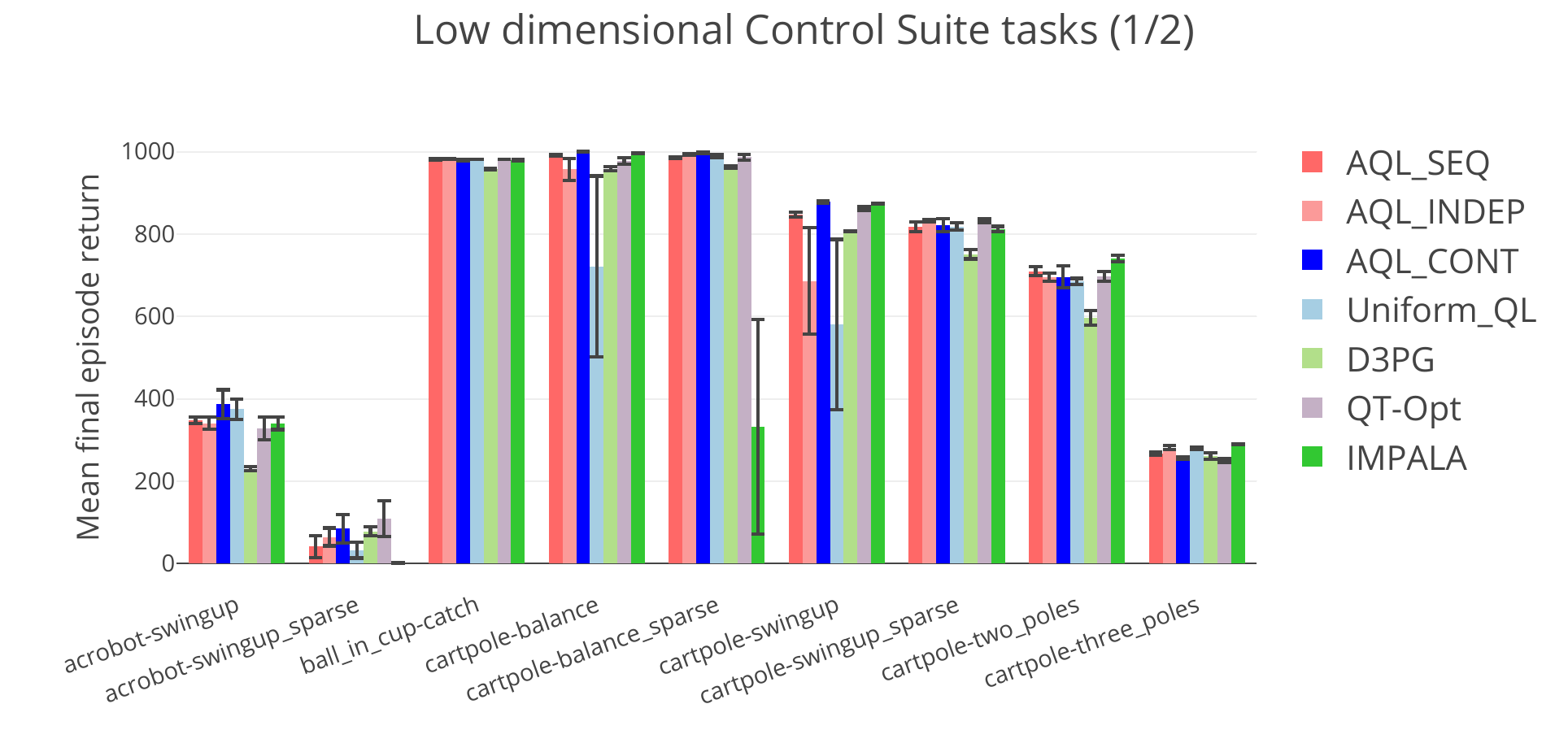}
\includegraphics[width=5.5in]{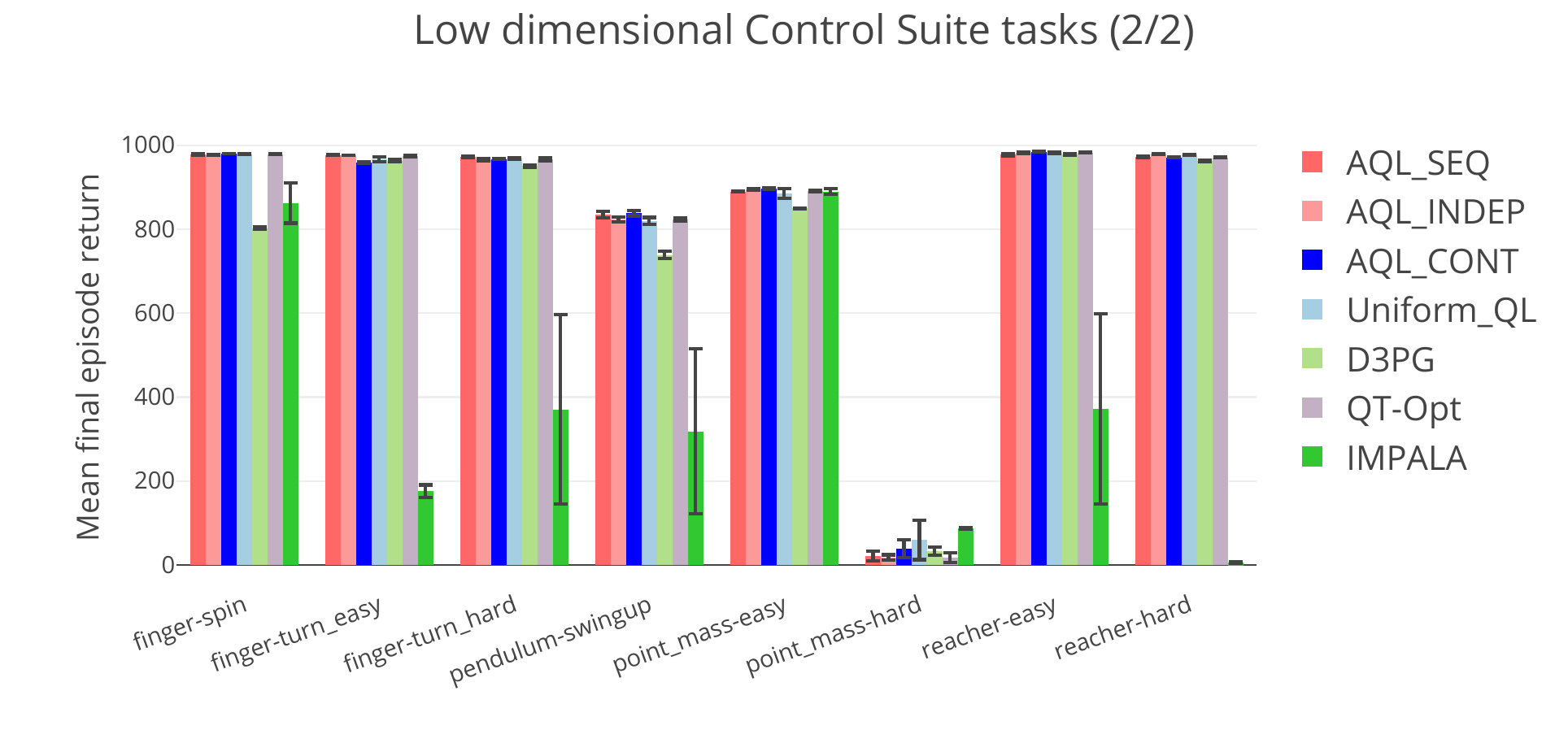}
\caption{Mean final performance, averaged over 3 seeds, for the low-dimensional Control Suite tasks.
The error bars represent the standard error of the mean episode return.}
\label{fig:control_suite_lowdim}
\end{figure}

\begin{figure}[h]
\centering
\includegraphics[width=5.5in]{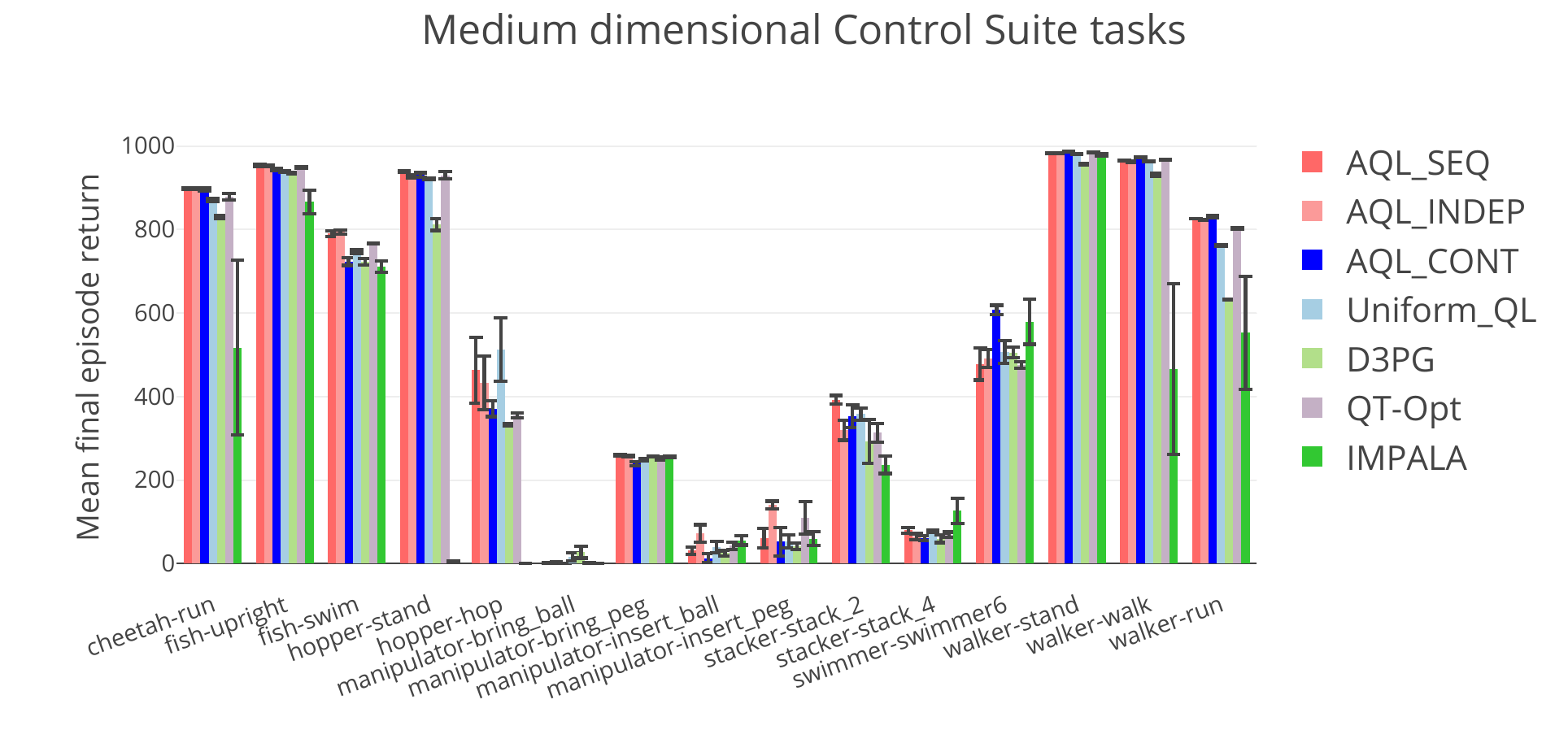}
\caption{Mean final performance, averaged over 3 seeds, for the medium-dimensional Control Suite tasks.
The error bars represent the standard error of the mean episode return.}
\label{fig:control_suite_mediumdim}
\end{figure}

\begin{figure}[h]
\centering
\includegraphics[width=5.5in]{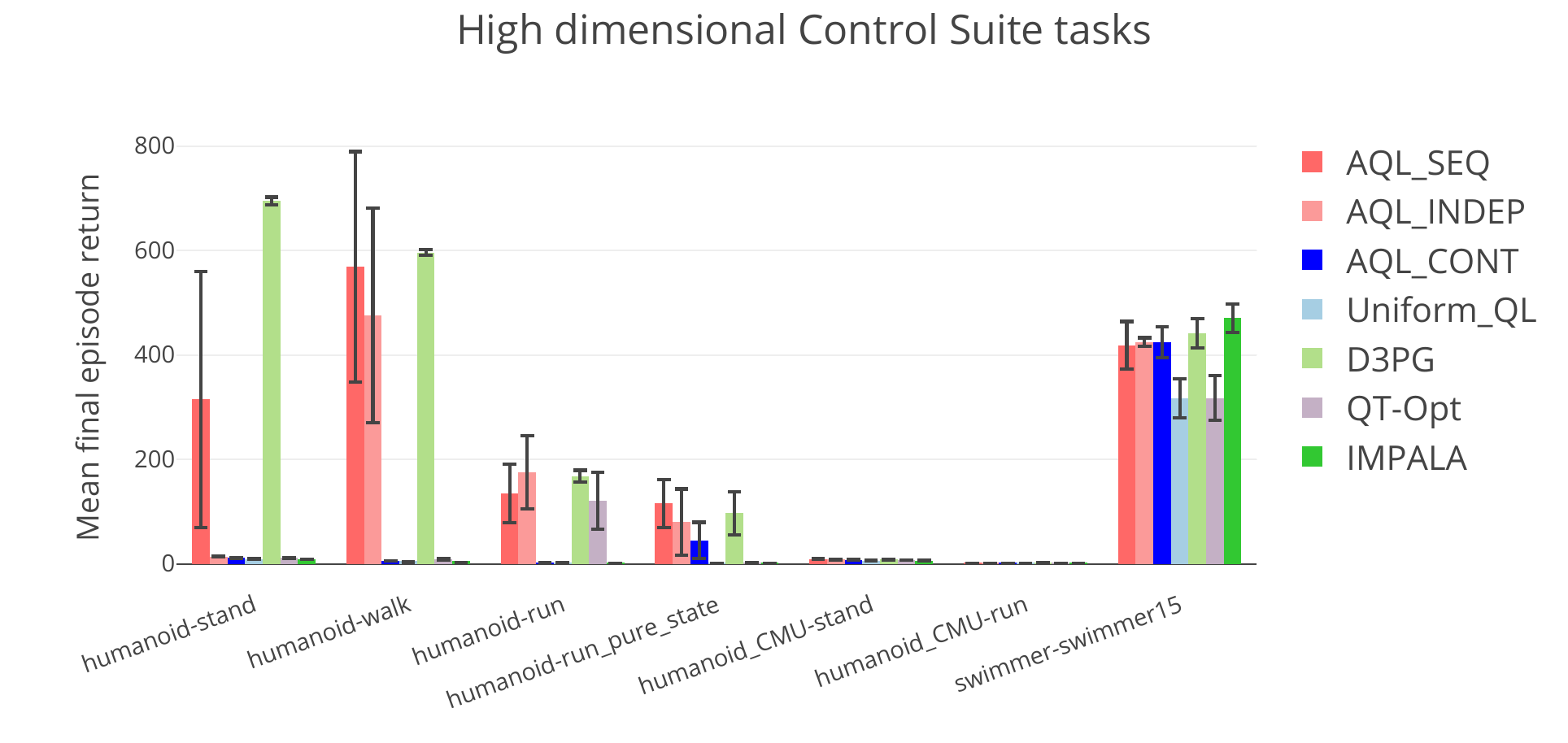}
\caption{Mean final performance, averaged over 3 seeds, for the high-dimensional Control Suite tasks.
The error bars represent the standard error of the mean episode return.}
\label{fig:control_suite_highdim}
\end{figure}

Figure~\ref{fig:control_suite_lowdim} shows the mean final performance results for the low-dimensional Control Suite tasks, with 1 or 2 sub-actions.
IMPALA and D3PG represent the weakest baselines on these tasks.
The mean final performance results for the medium-dimensional Control Suite tasks, with 4 to 6 sub-actions, is shown in Figure~\ref{fig:control_suite_mediumdim}. For medium-dimensional tasks, D3PG and IMPALA again stand out as the methods that perform worse than the other baselines on average. Results for high-dimensional tasks are shown in Figure~\ref{fig:control_suite_highdim}. Here, D3PG and the AQL methods outperform QT-Opt, IMPALA and Uniform Q-Learning.

\begin{figure*}[h]
\centering
\hspace*{0cm}\includegraphics[width=1.45in]{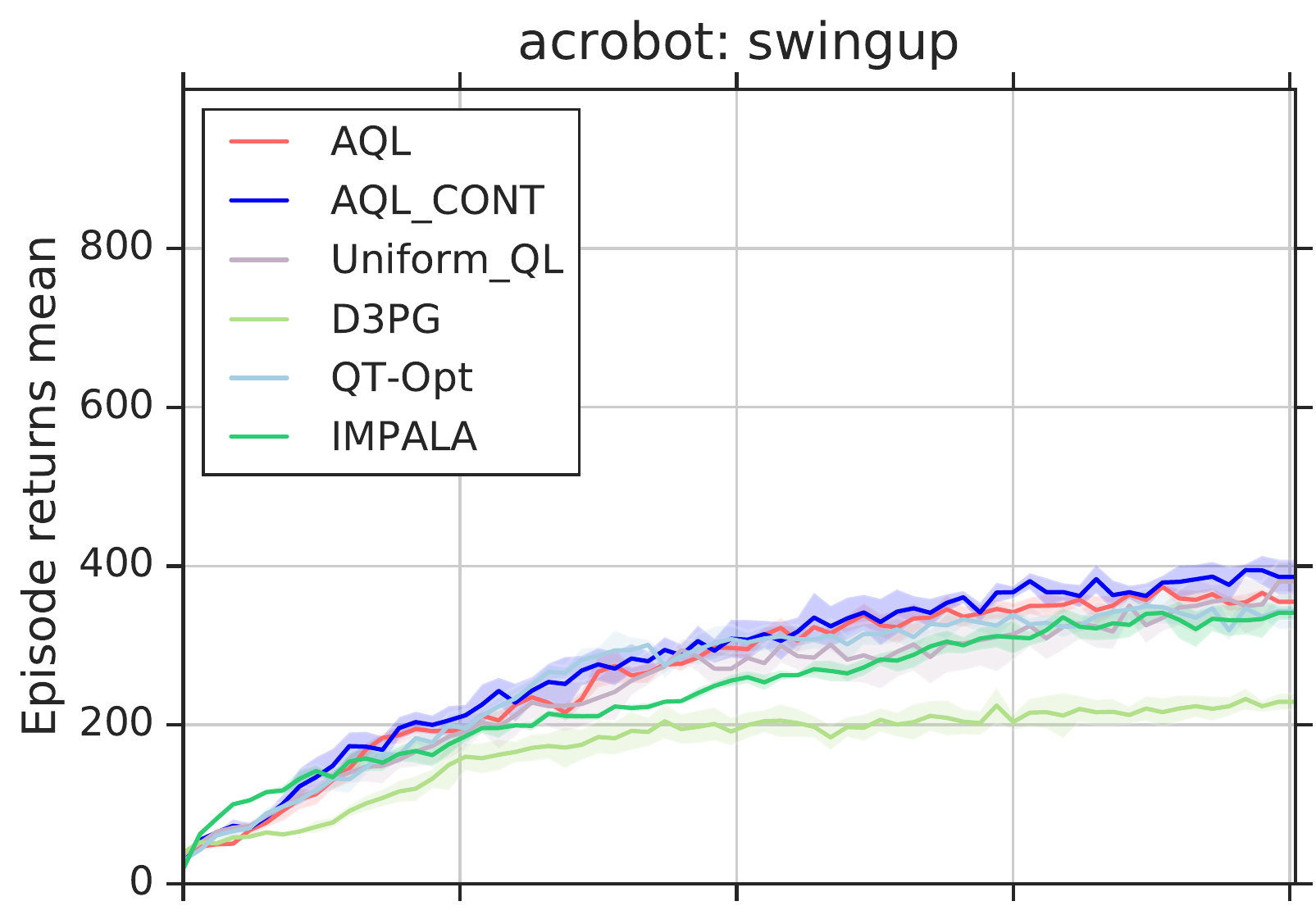}
\hspace*{0cm}\includegraphics[width=1.3in]{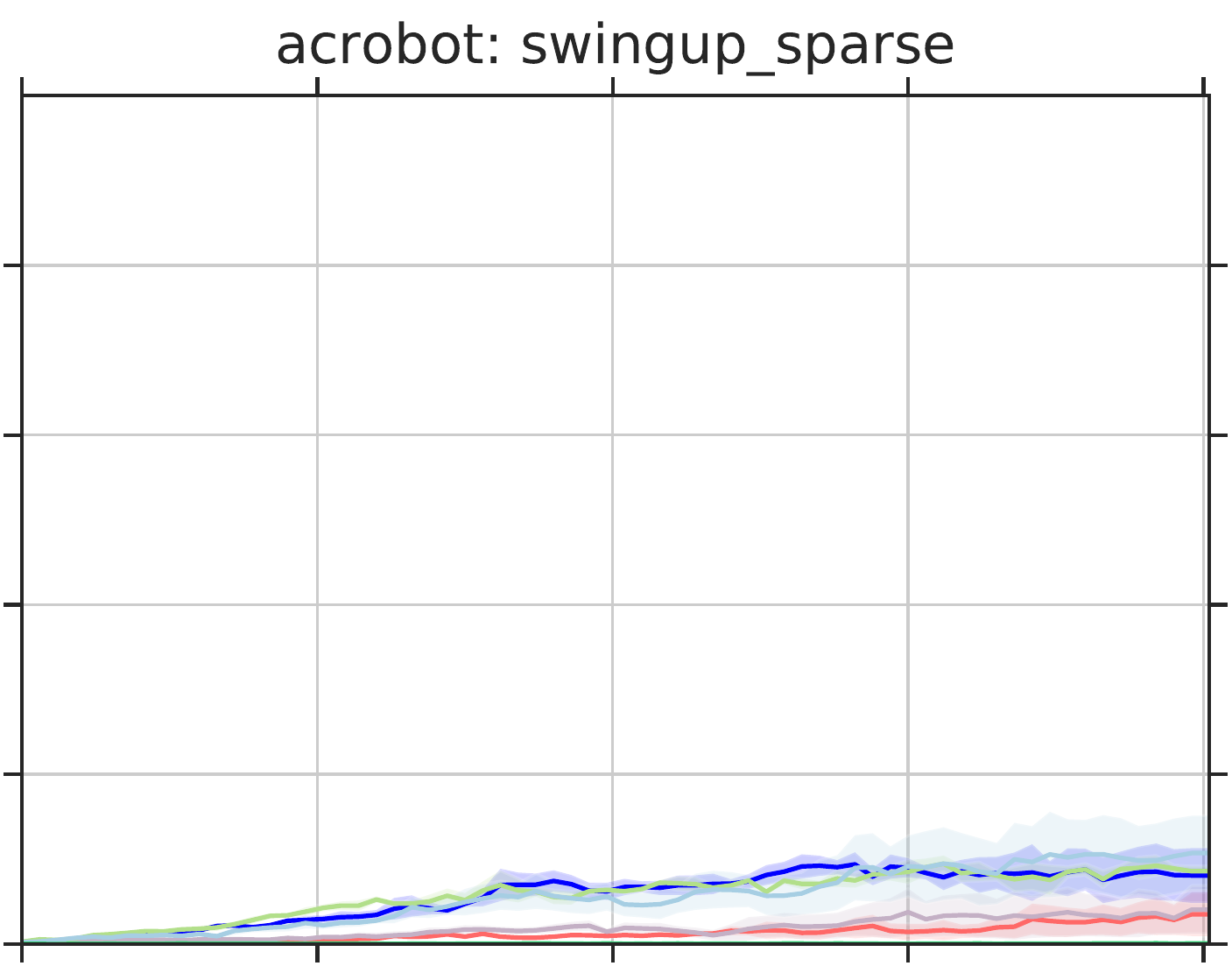}
\hspace*{0cm}\includegraphics[width=1.3in]{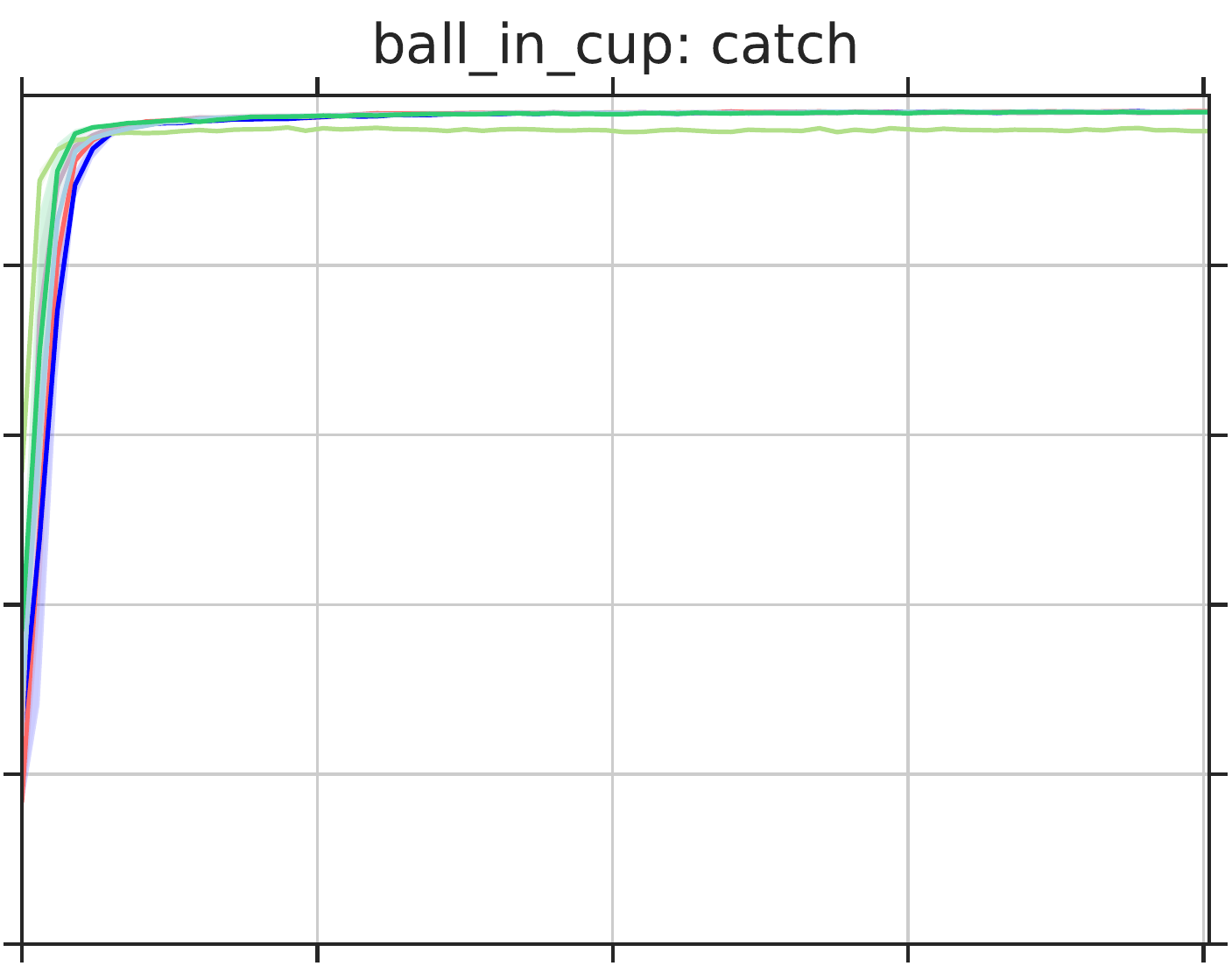}\\
\hspace*{0cm}\includegraphics[width=1.45in]{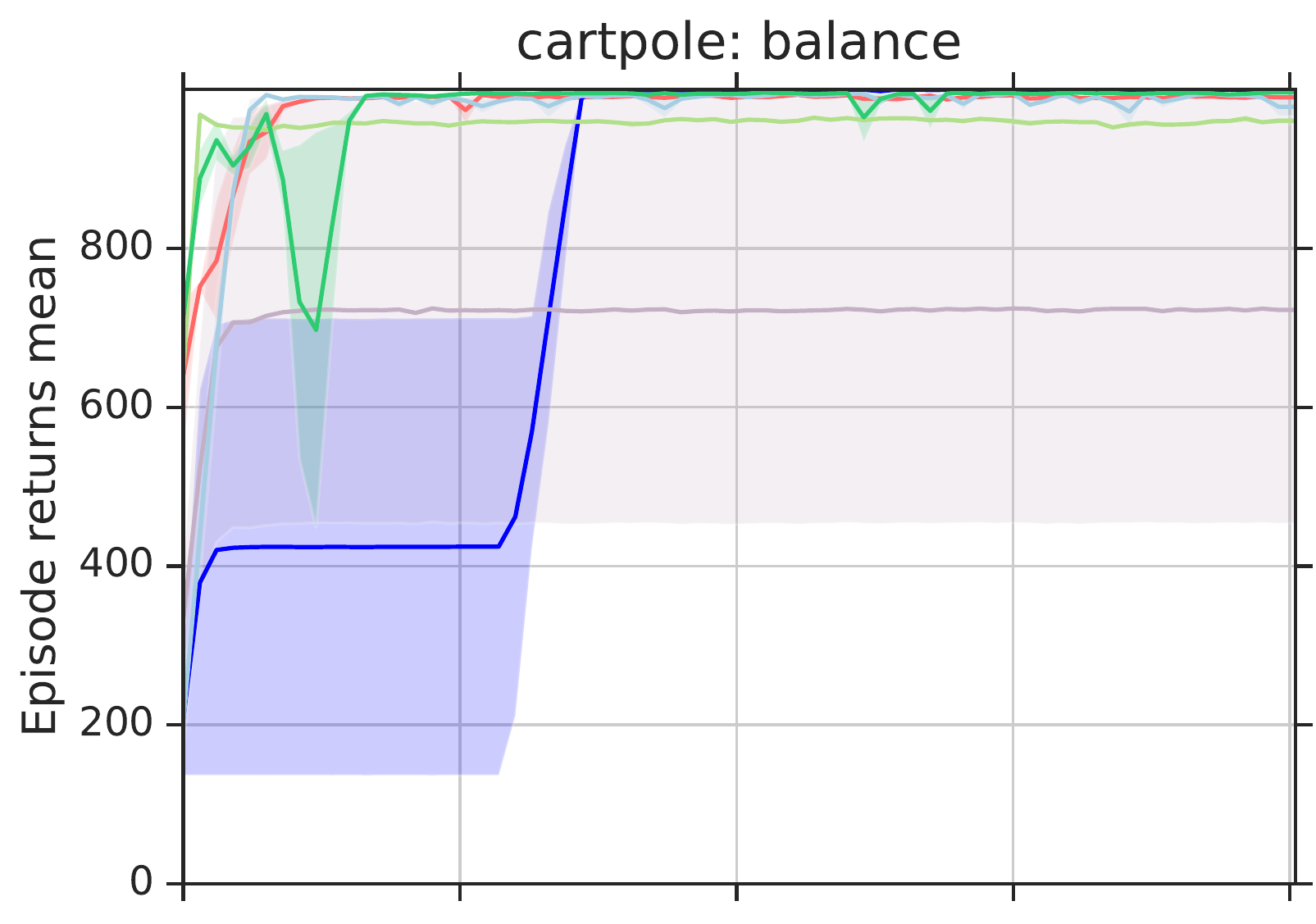}
\hspace*{0cm}\includegraphics[width=1.3in]{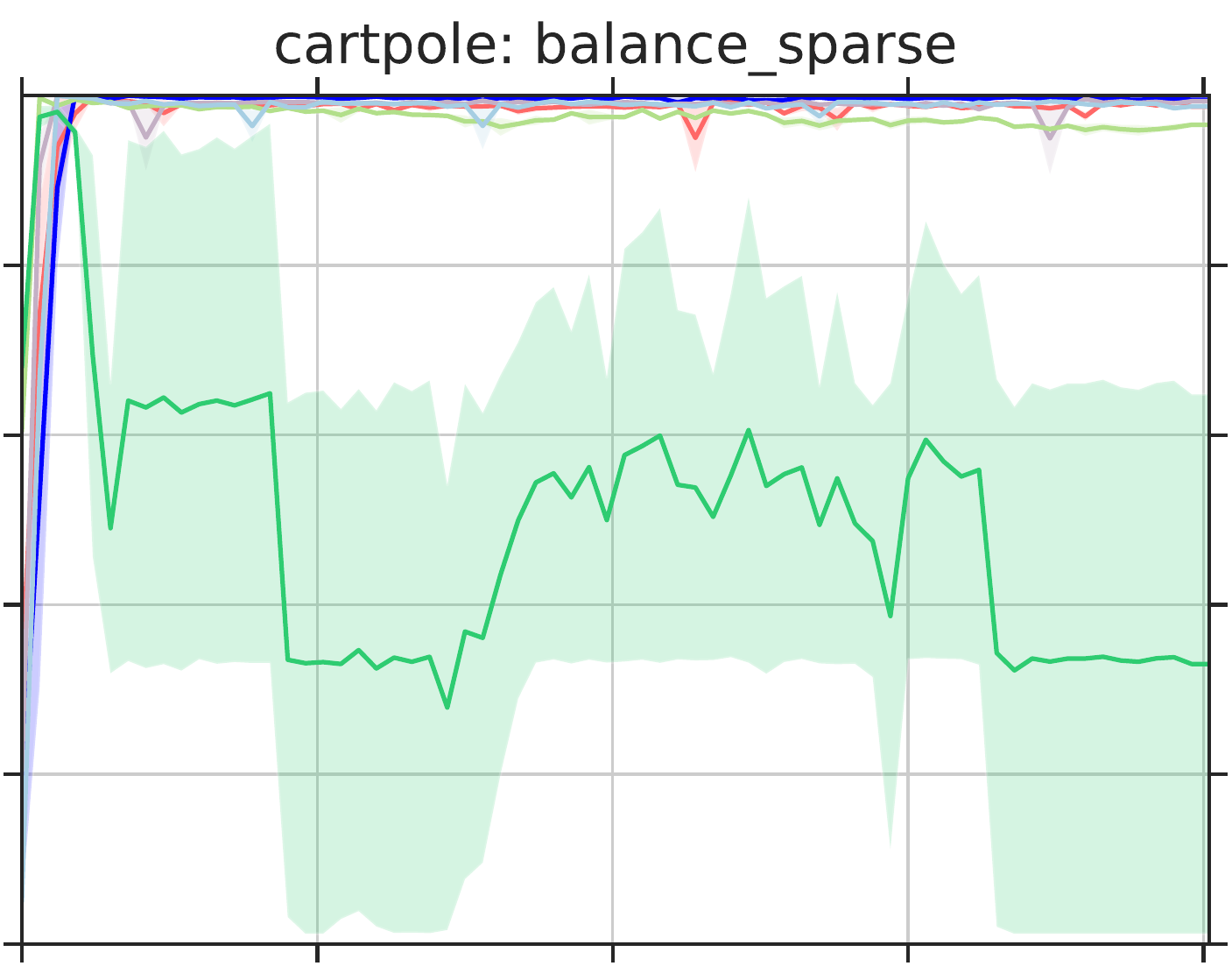}
\hspace*{0cm}\includegraphics[width=1.3in]{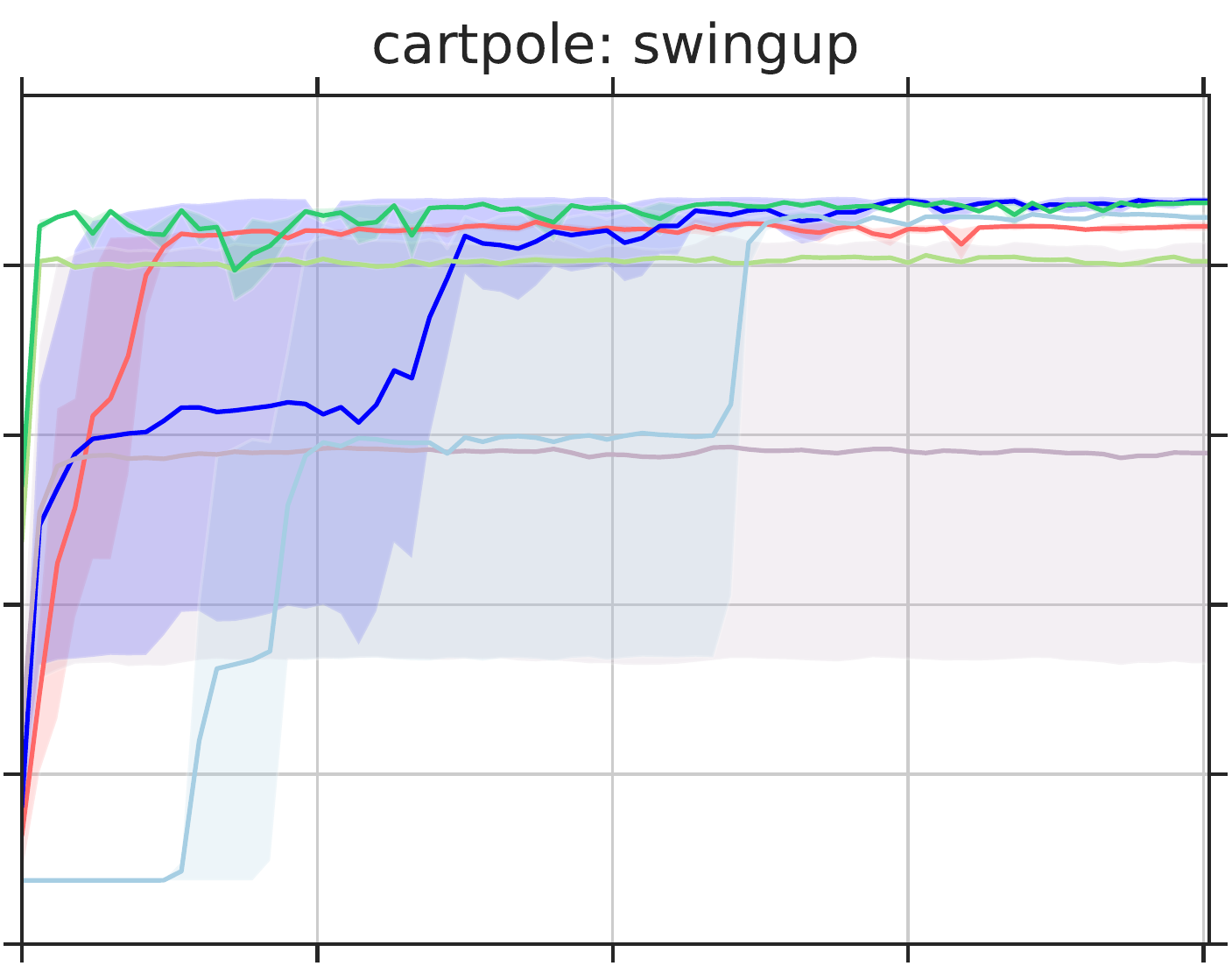}\\
\hspace*{0cm}\includegraphics[width=1.45in]{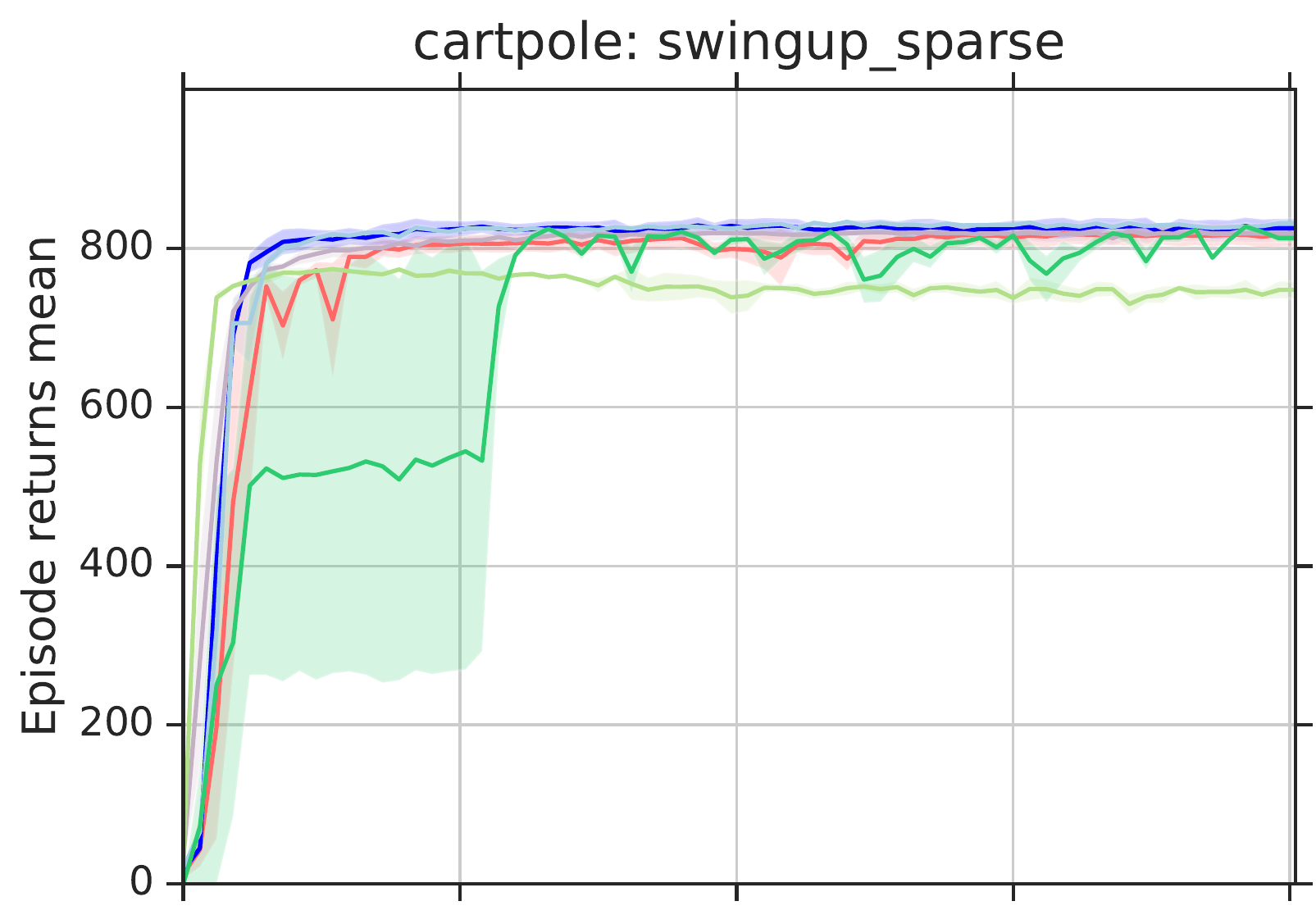}
\hspace*{0cm}\includegraphics[width=1.3in]{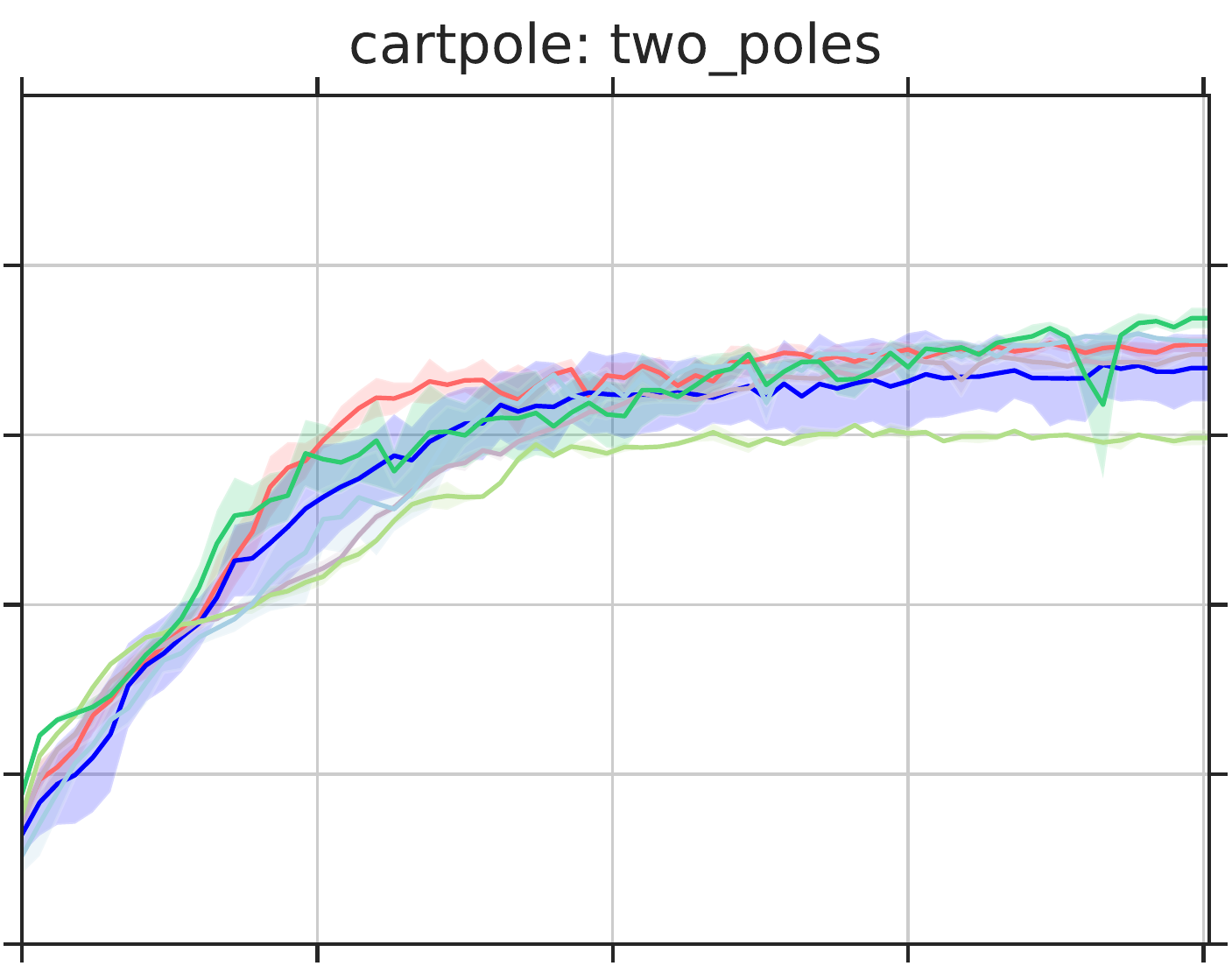}
\hspace*{0cm}\includegraphics[width=1.3in]{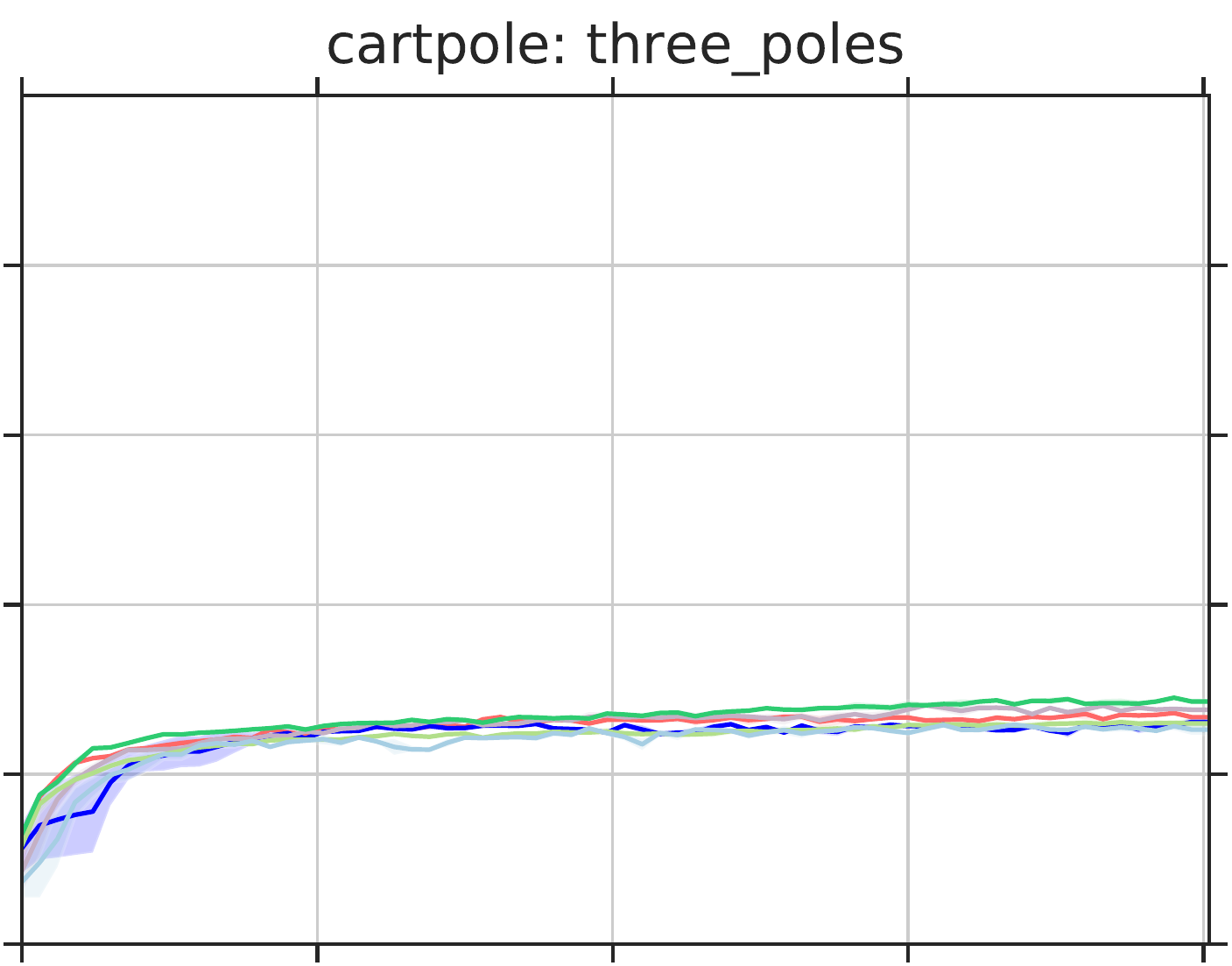}\\
\hspace*{0cm}\includegraphics[width=1.45in]{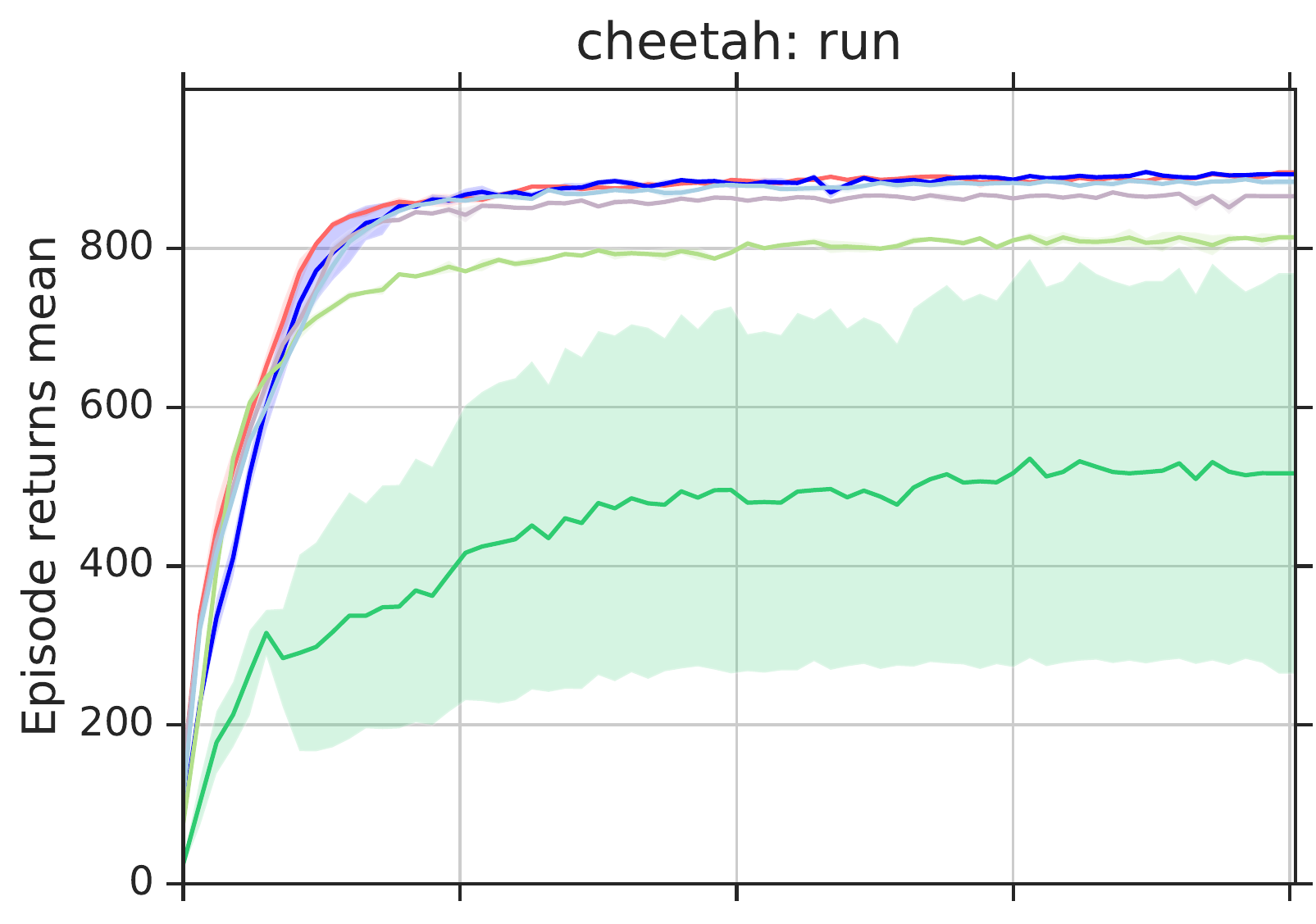}
\hspace*{0cm}\includegraphics[width=1.3in]{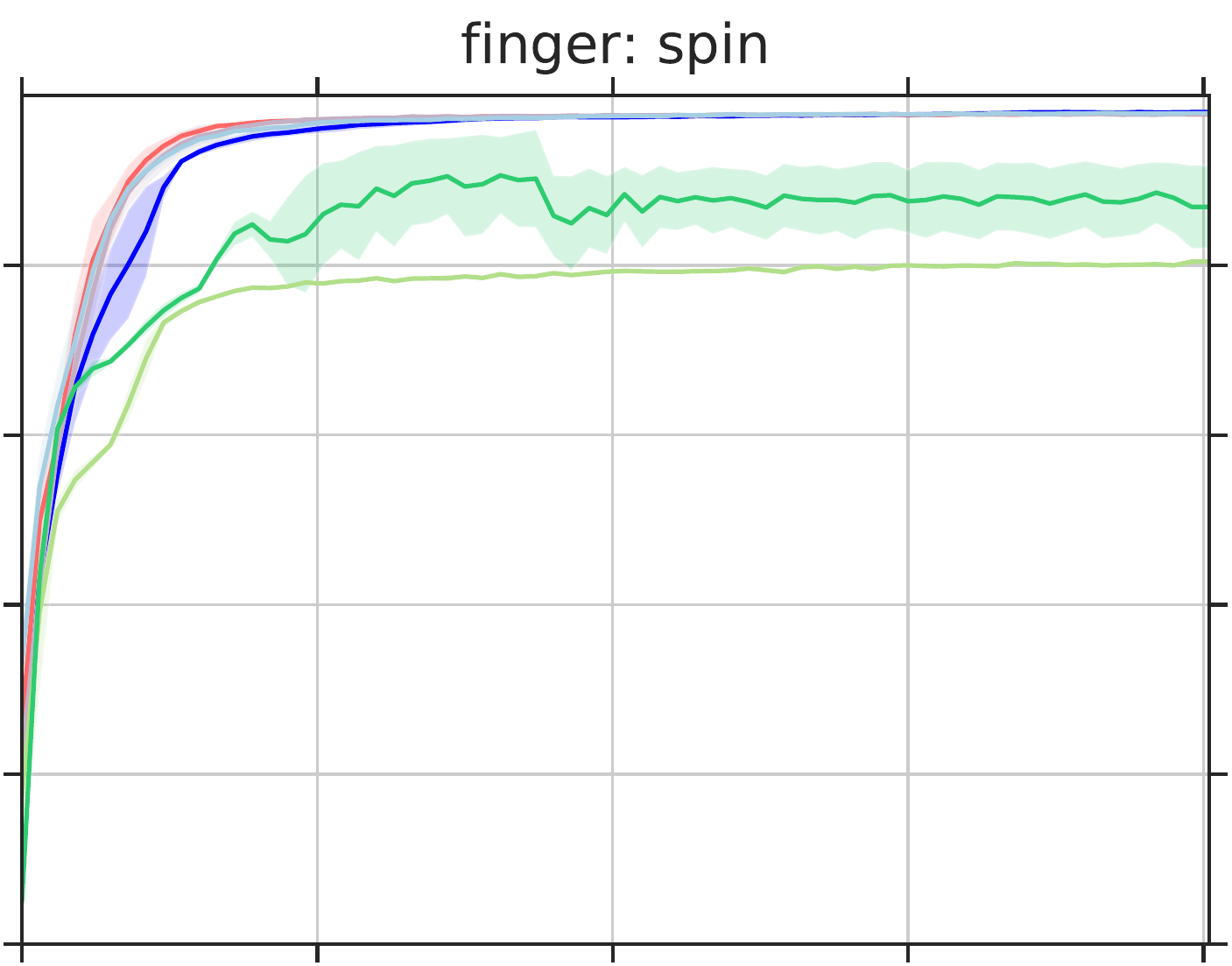}
\hspace*{0cm}\includegraphics[width=1.3in]{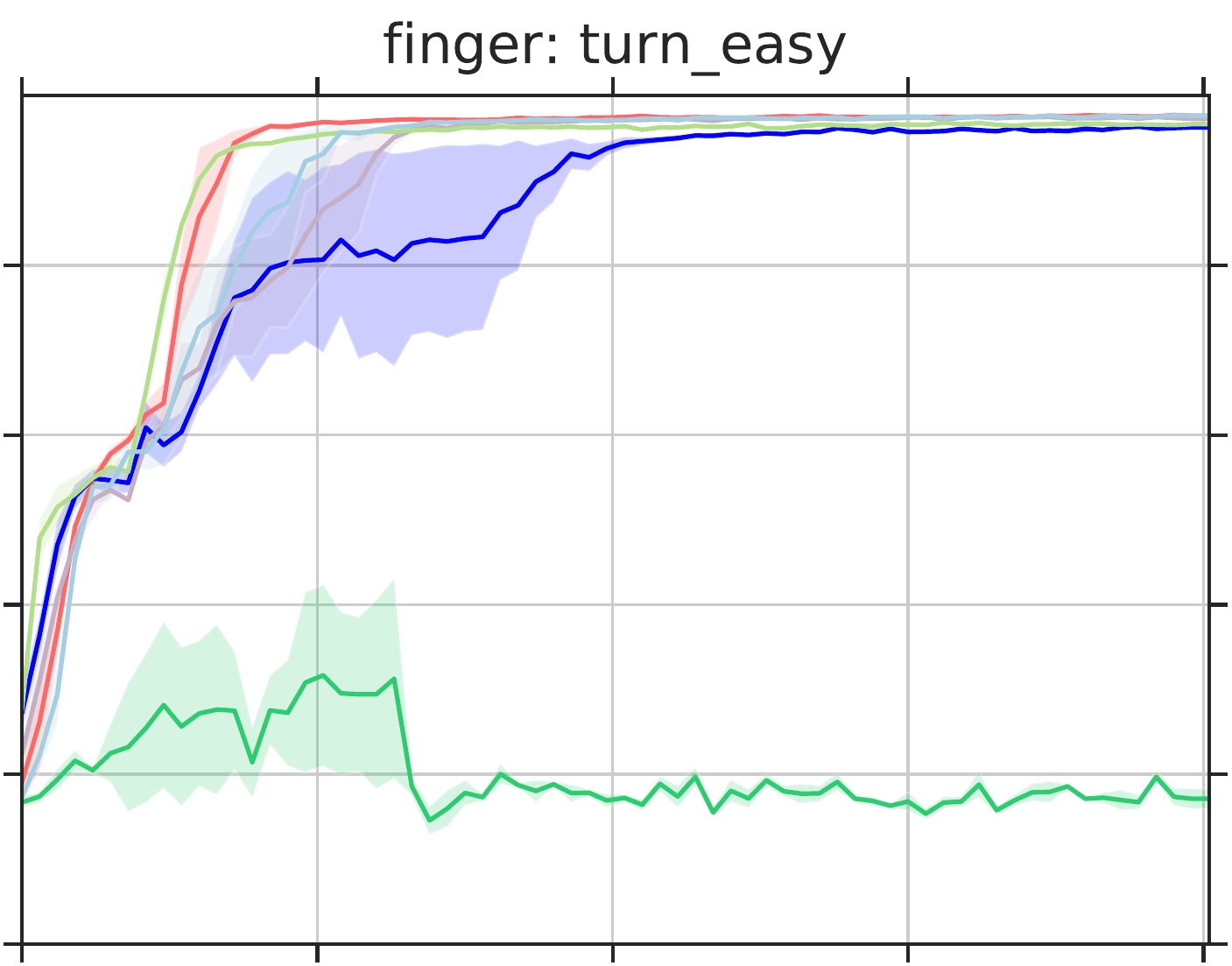}\\
\hspace*{0cm}\includegraphics[width=1.45in]{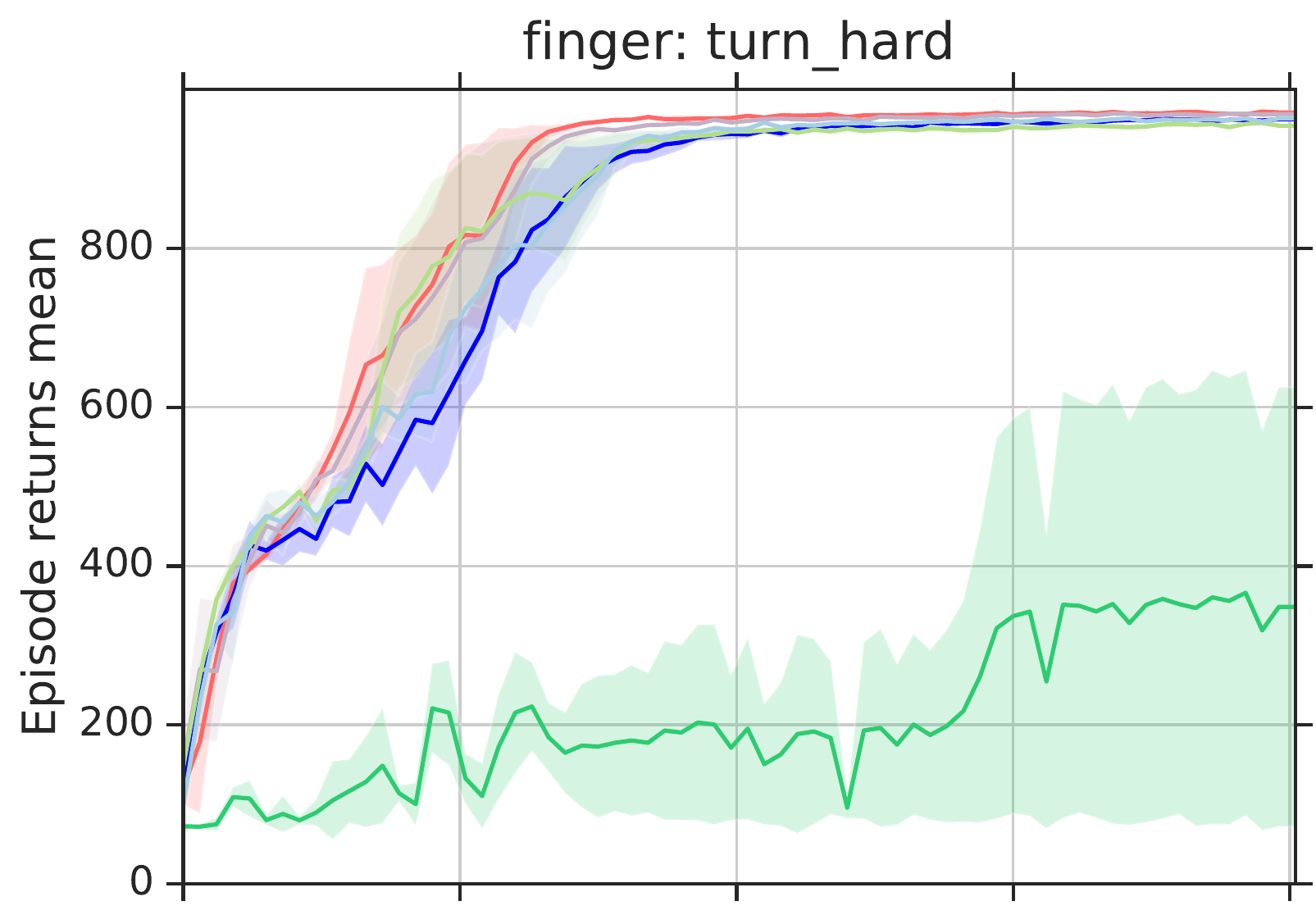}
\hspace*{0cm}\includegraphics[width=1.3in]{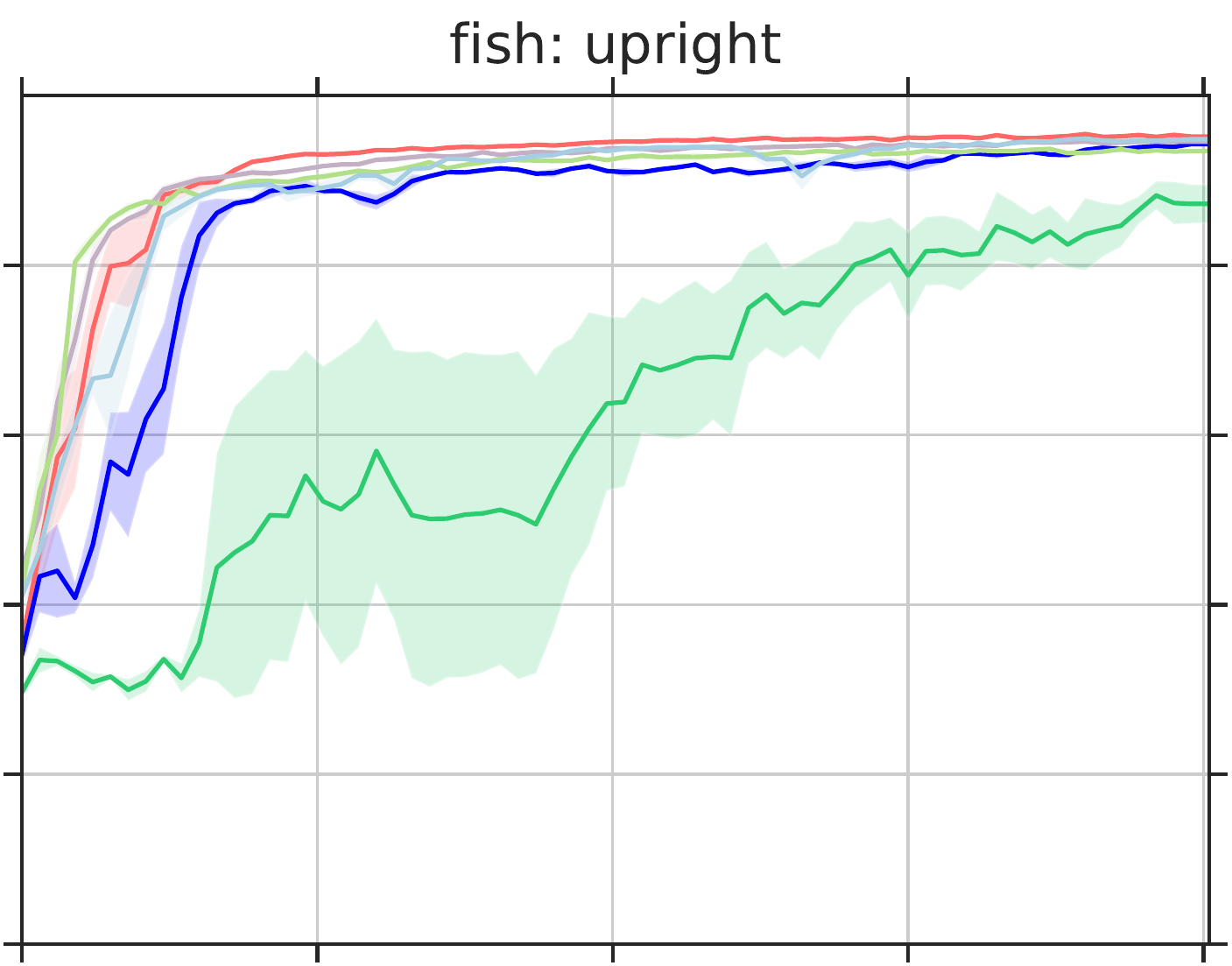}
\hspace*{0cm}\includegraphics[width=1.3in]{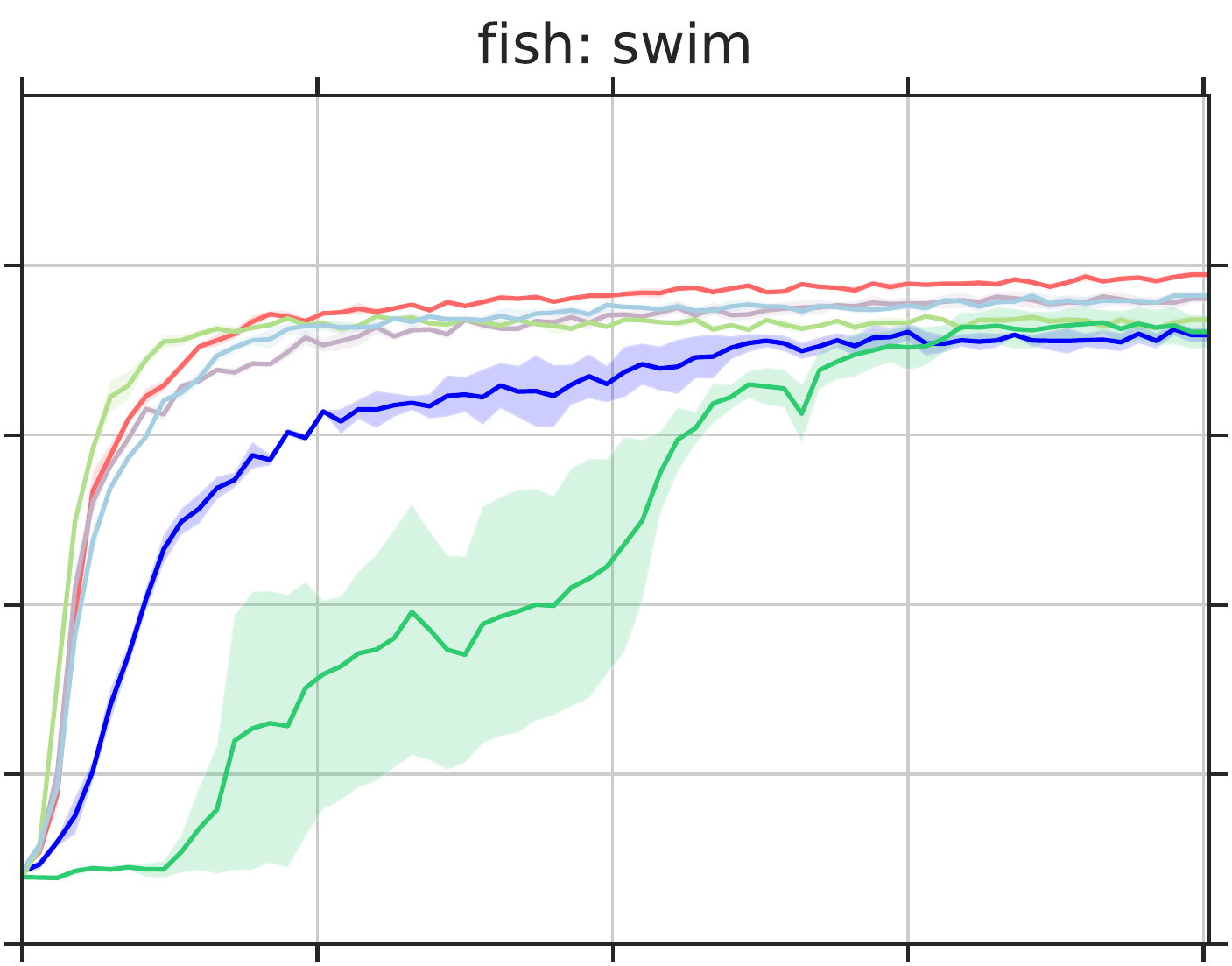}\\
\hspace*{0cm}\includegraphics[width=1.45in]{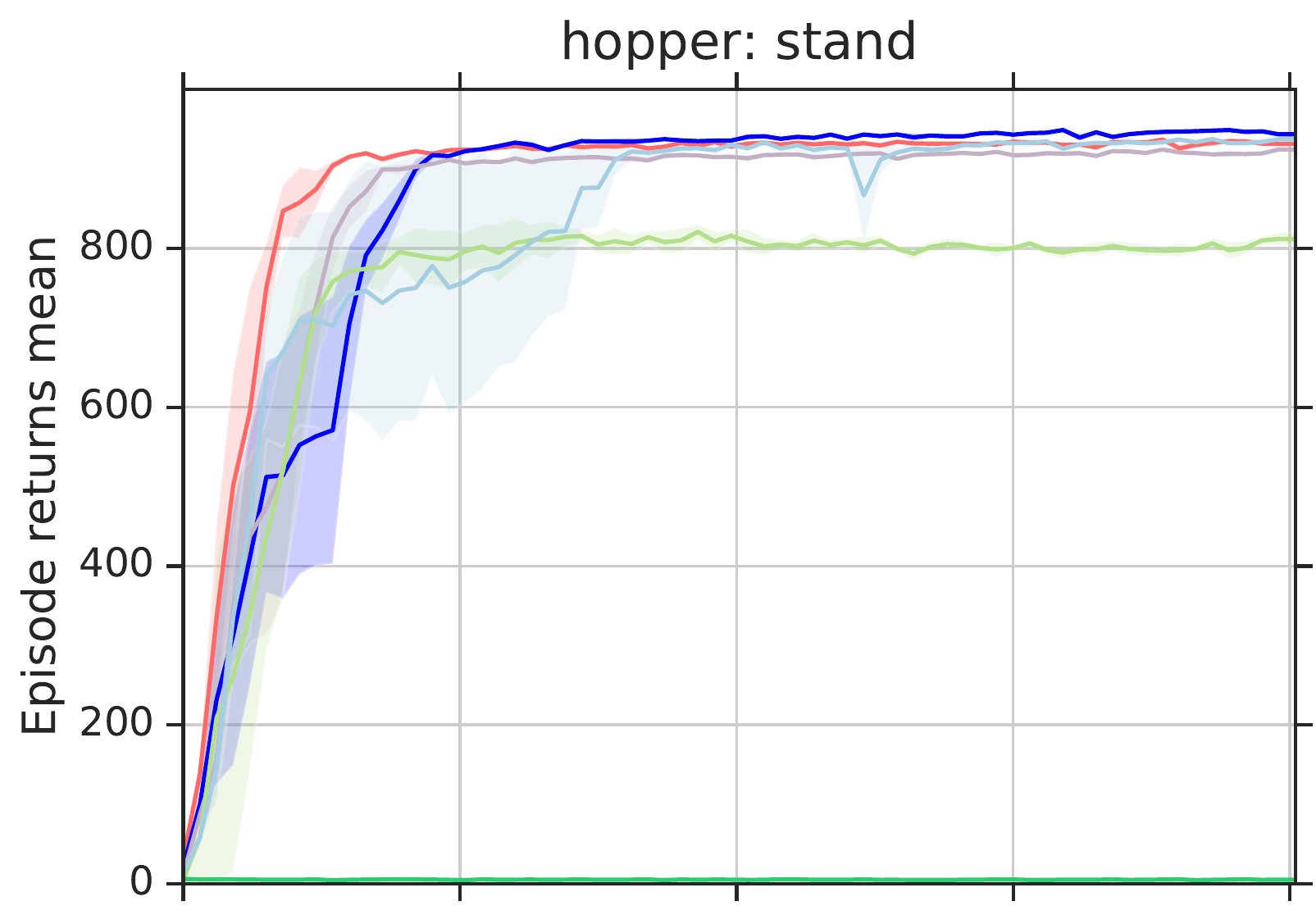}
\hspace*{0cm}\includegraphics[width=1.3in]{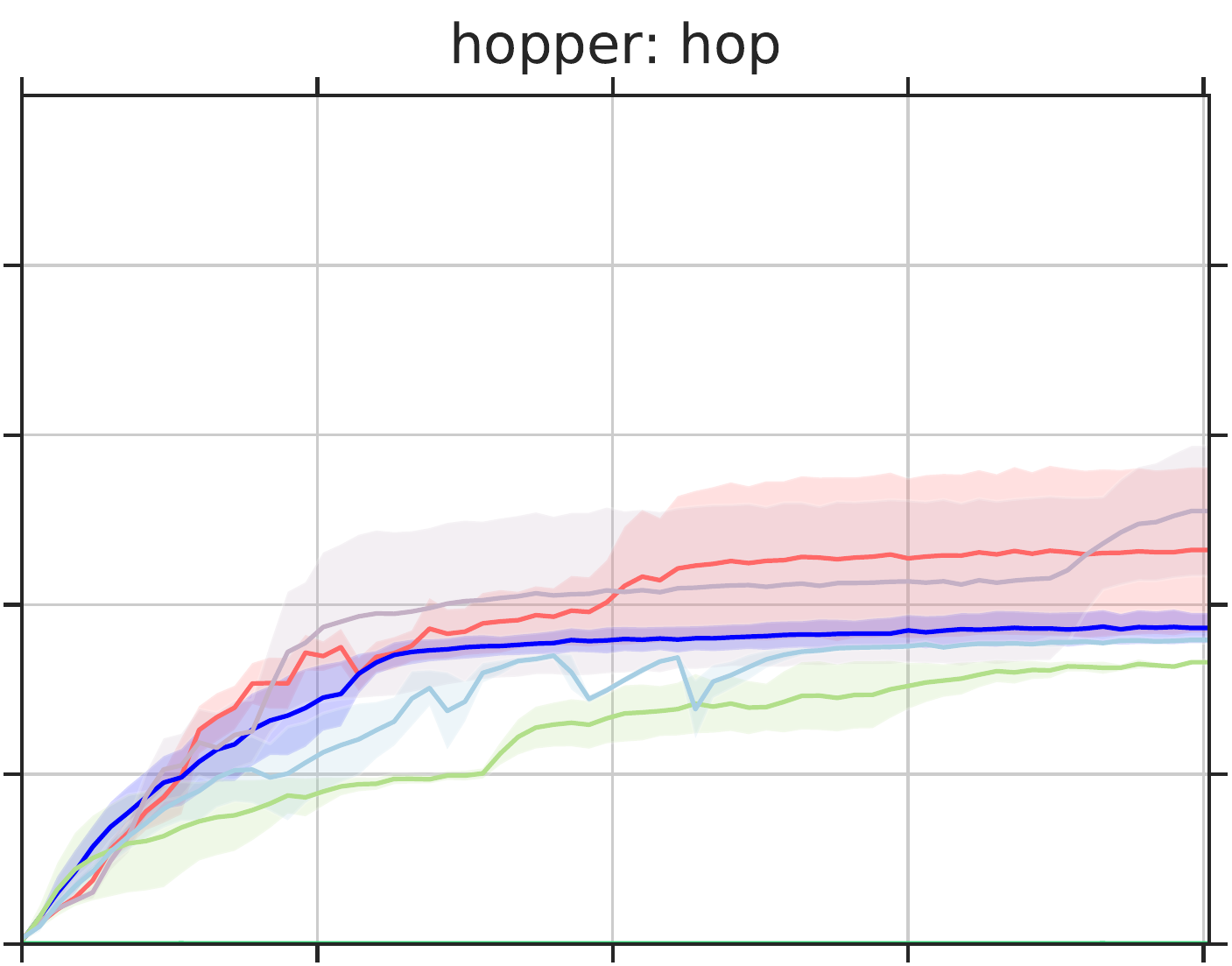}
\hspace*{0cm}\includegraphics[width=1.3in]{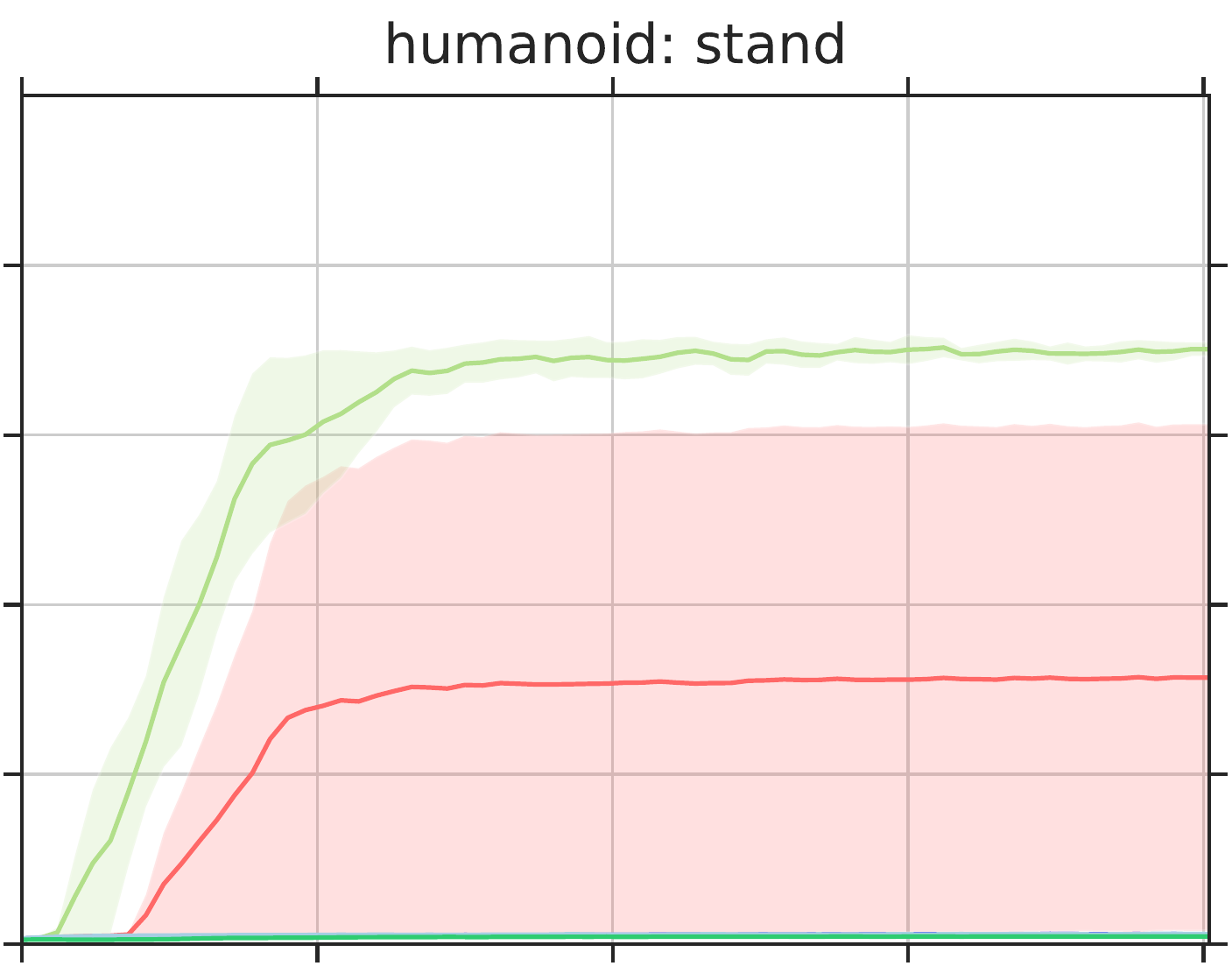}\\
\hspace*{0cm}\includegraphics[width=1.45in]{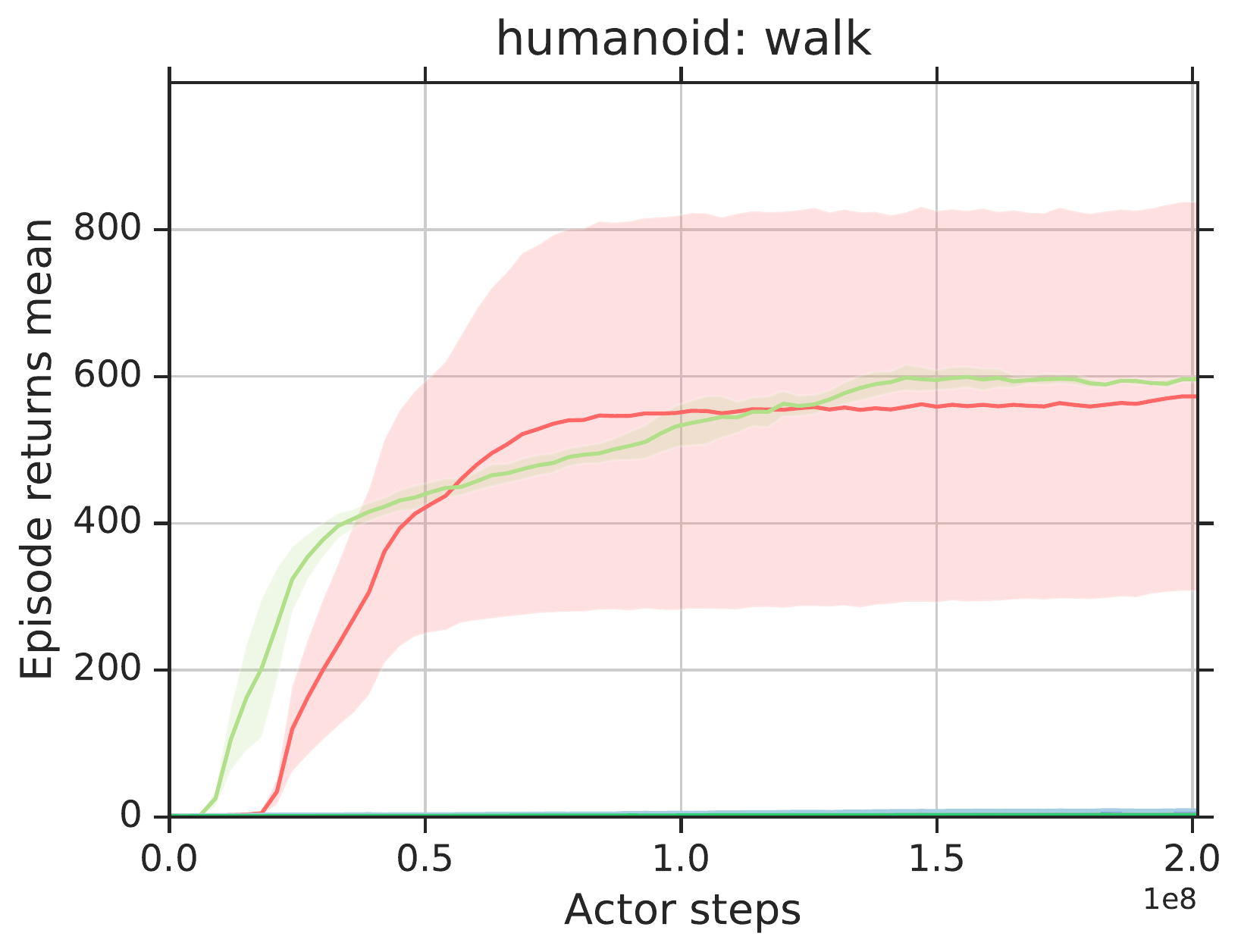}
\hspace*{0cm}\includegraphics[width=1.3in]{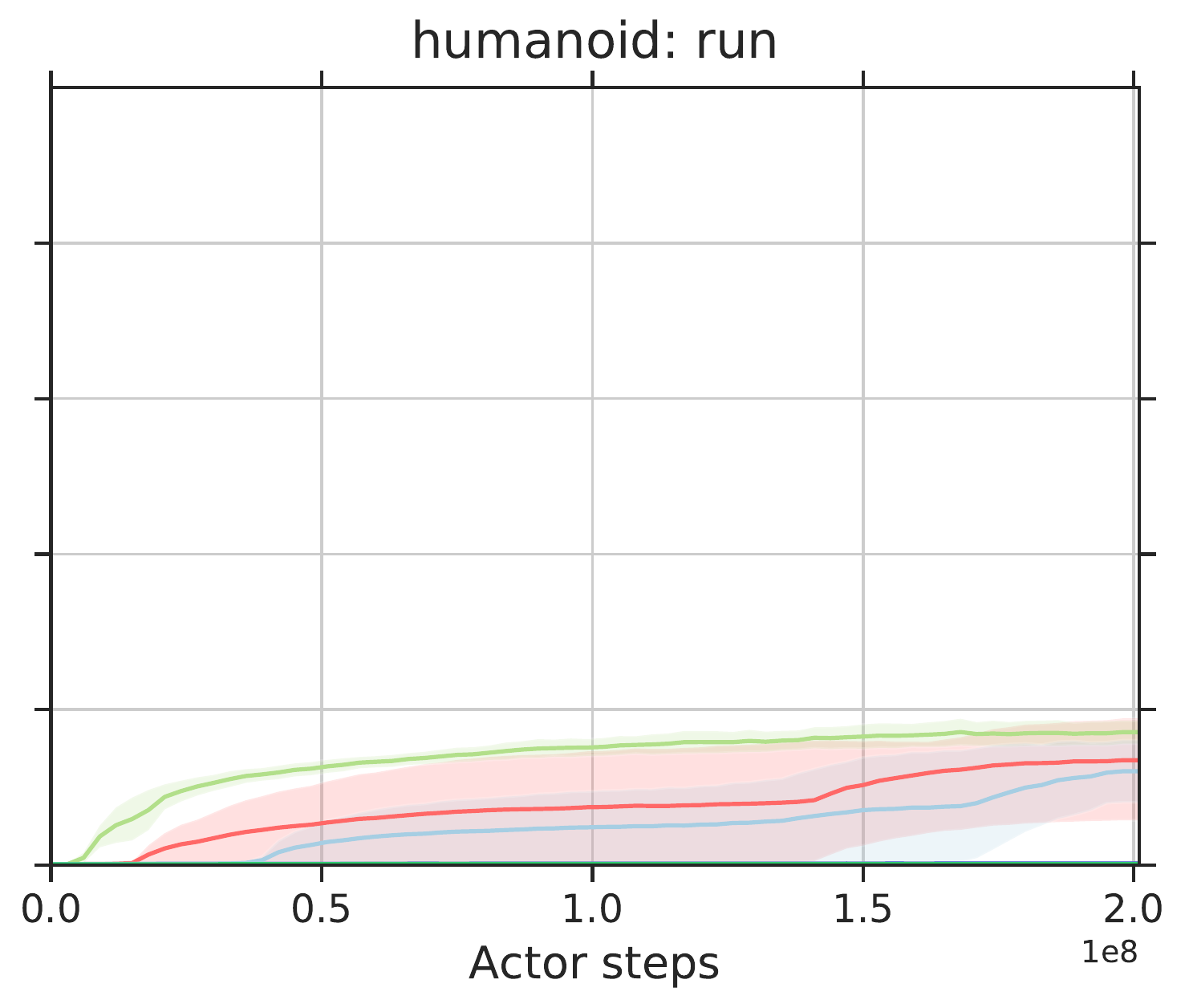}
\hspace*{0cm}\includegraphics[width=1.3in]{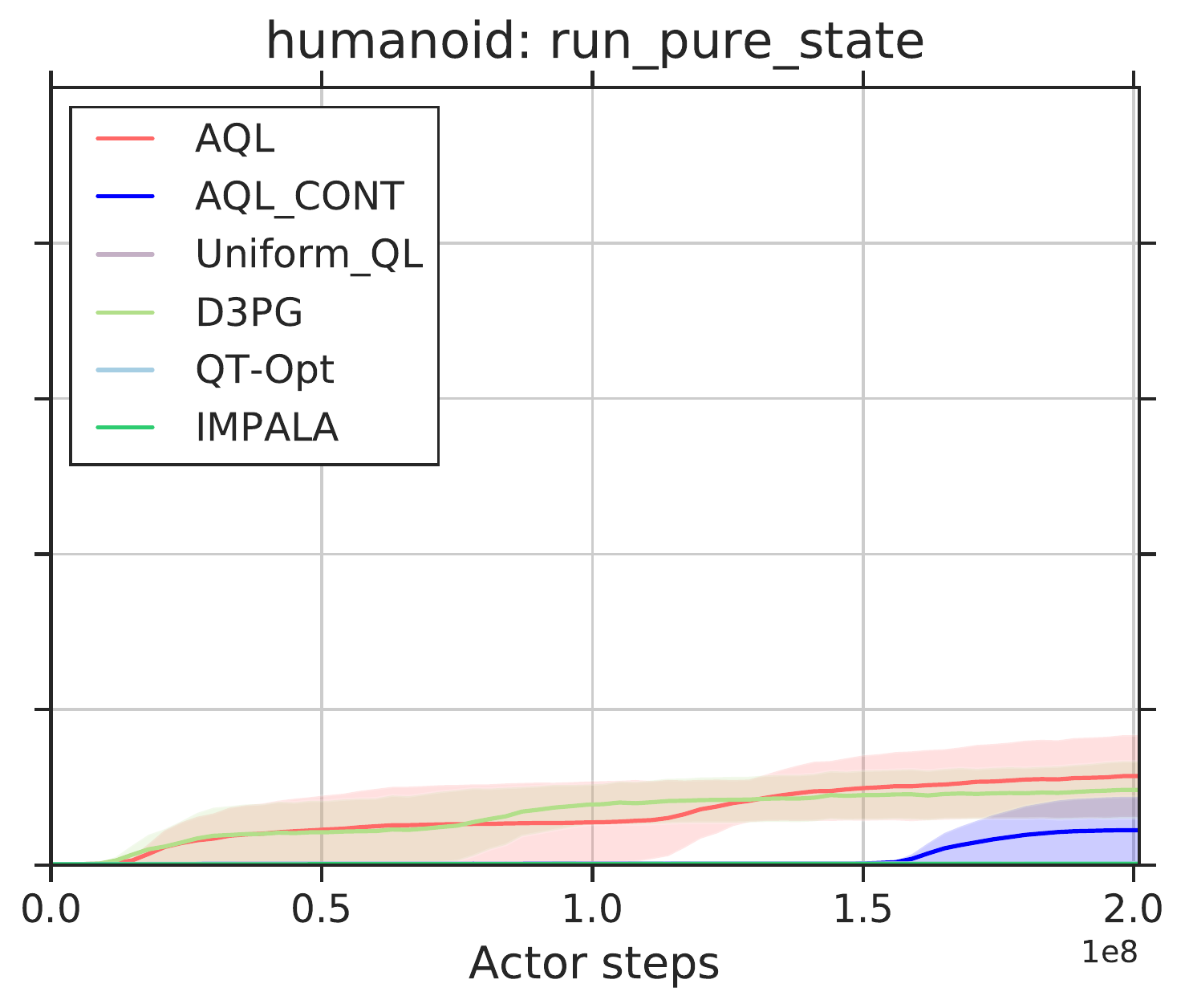}\\
\caption{Task specific learning curves for the Control Suite (1/2).}
\label{fig:control_suite_av_learning_curves_1}
\end{figure*}

\begin{figure*}[h]
\centering
\hspace*{0cm}\includegraphics[width=1.45in]{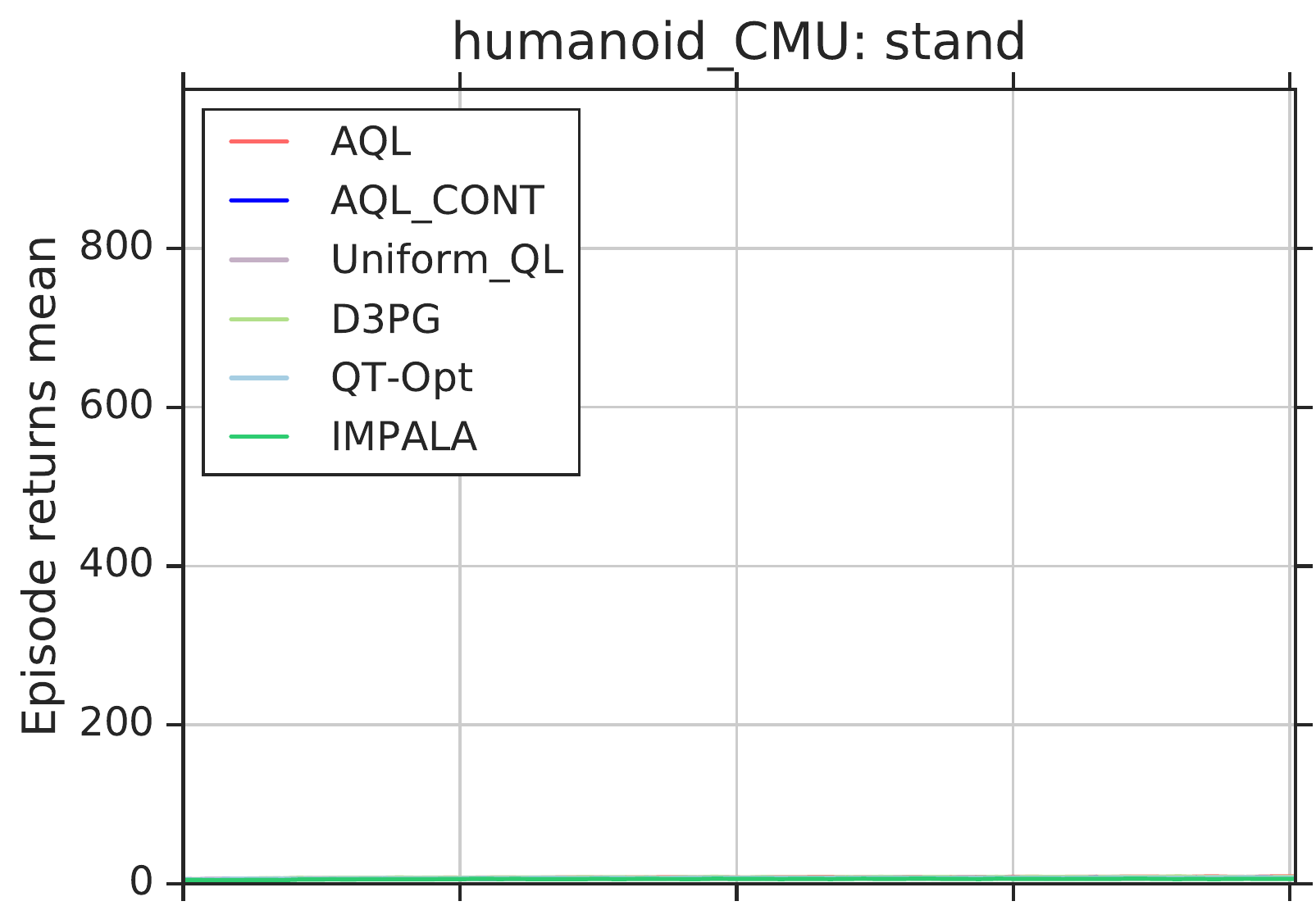}
\hspace*{0cm}\includegraphics[width=1.3in]{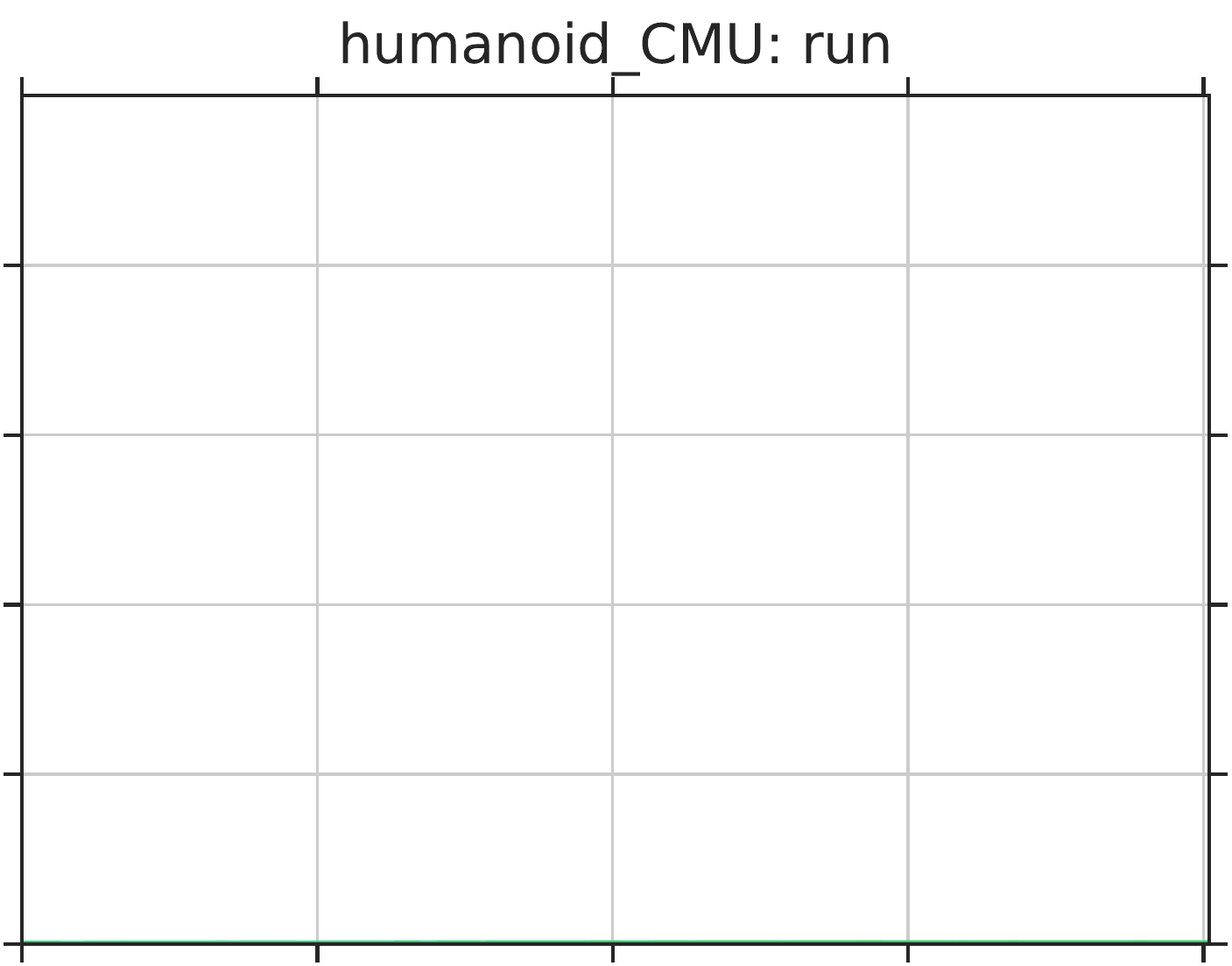}
\hspace*{0cm}\includegraphics[width=1.3in]{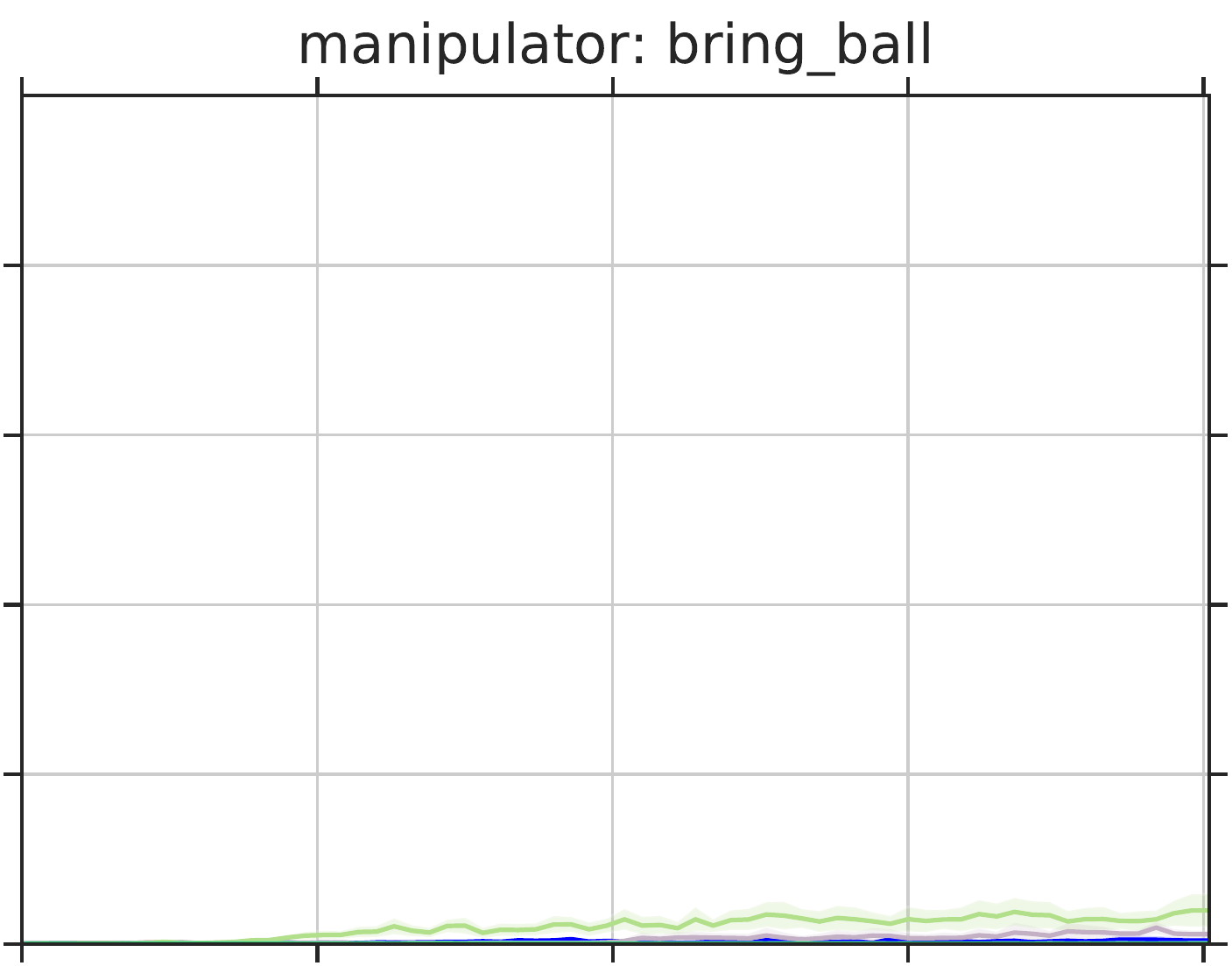}\\
\hspace*{0cm}\includegraphics[width=1.45in]{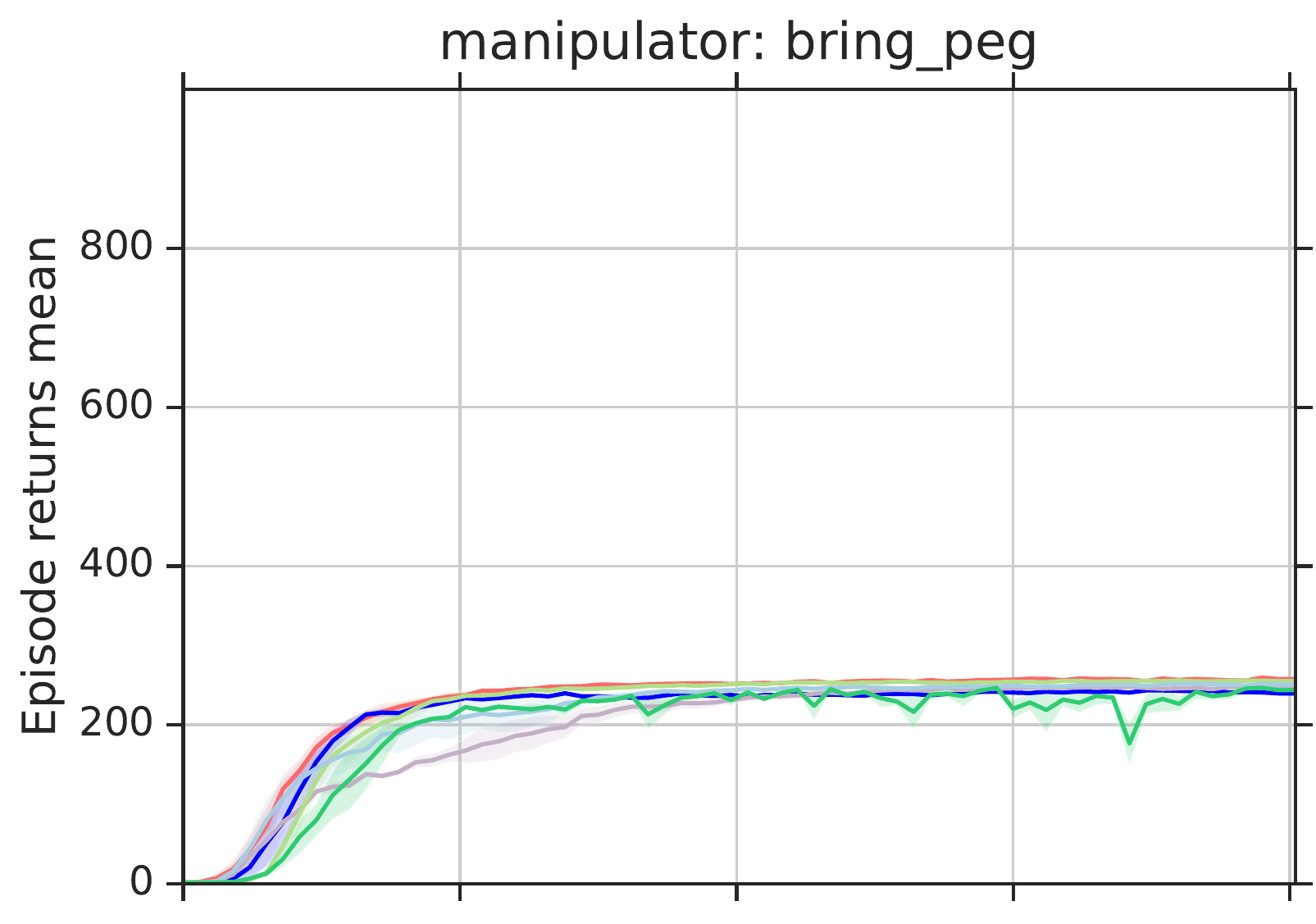}
\hspace*{0cm}\includegraphics[width=1.3in]{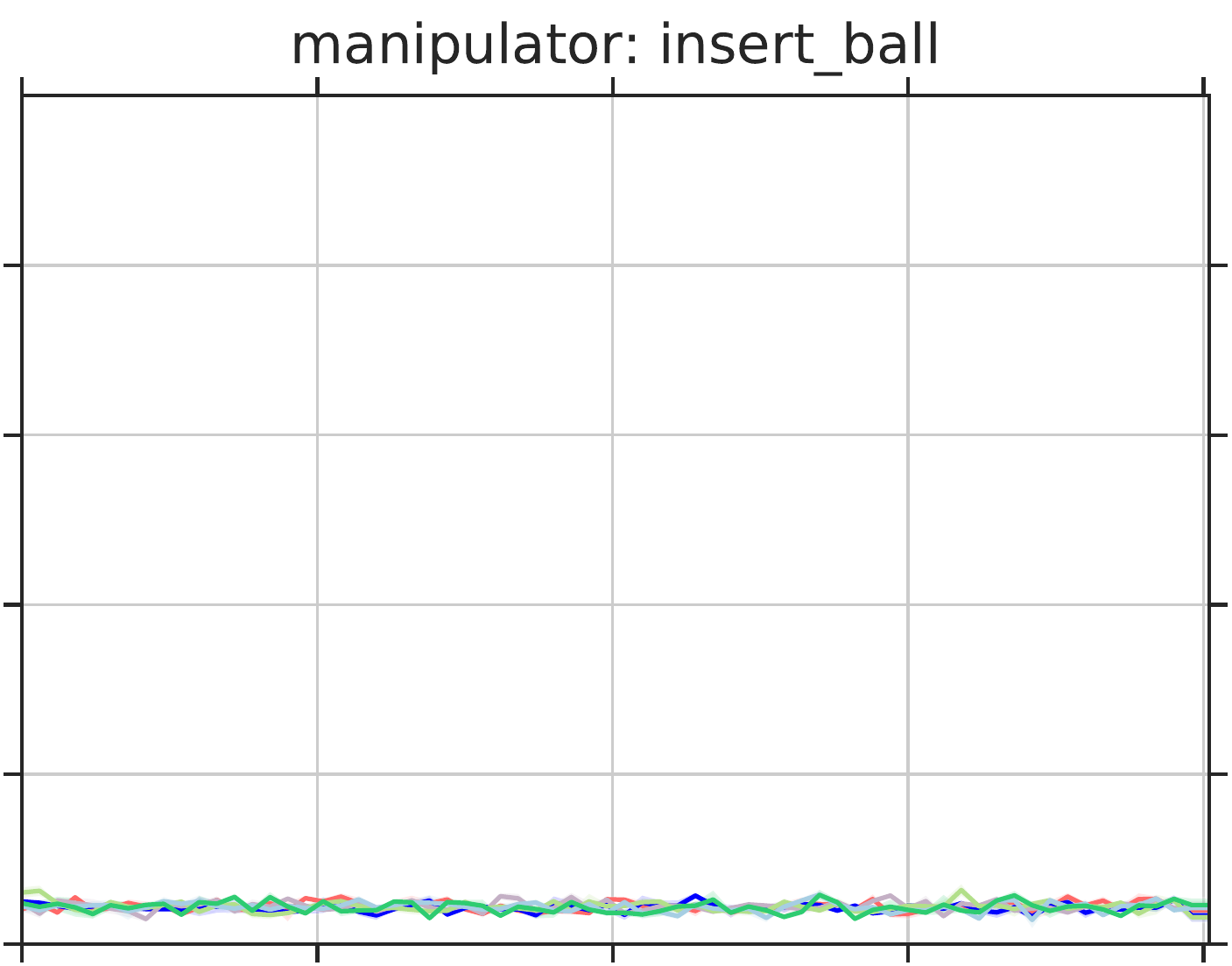}
\hspace*{0cm}\includegraphics[width=1.3in]{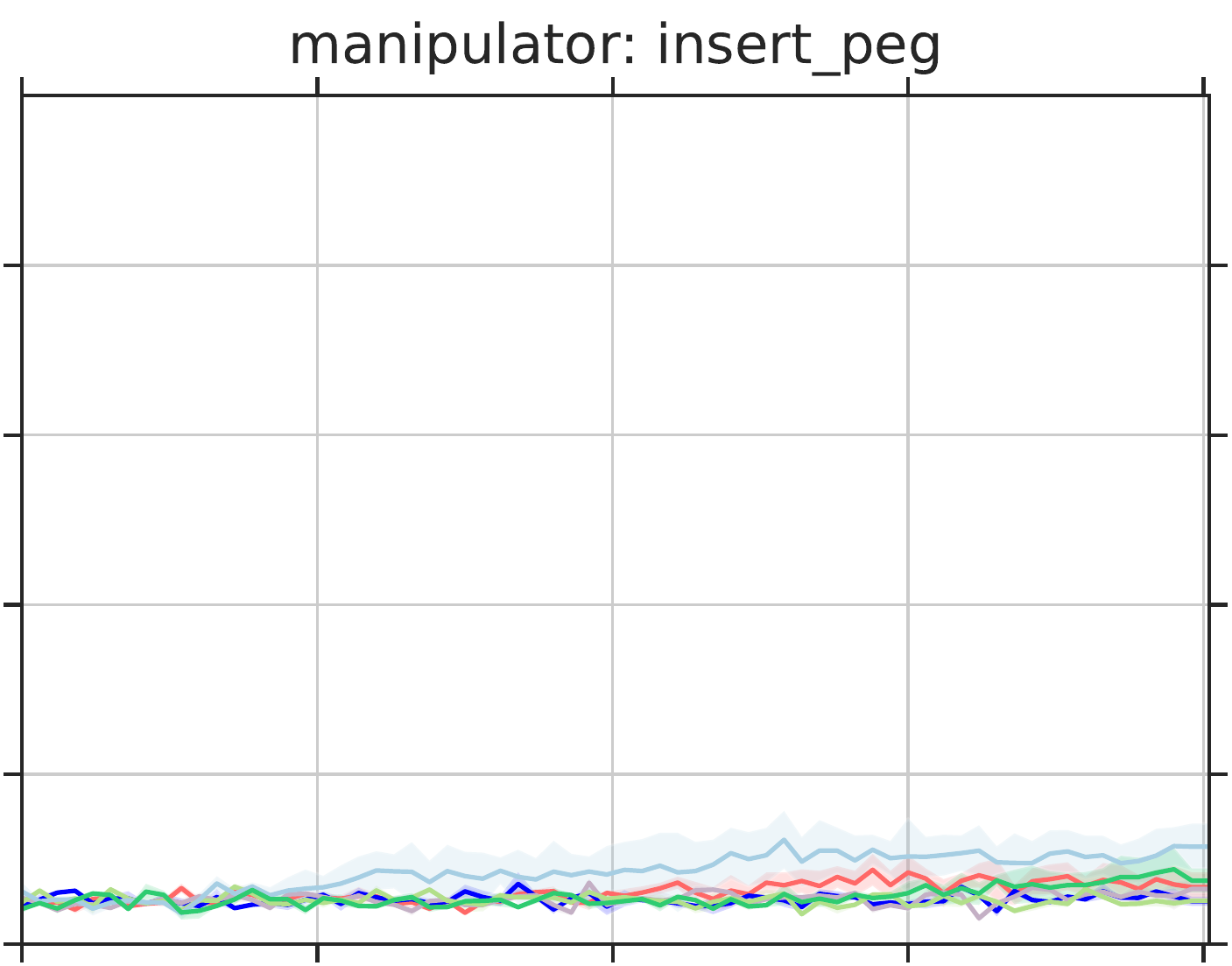}\\
\hspace*{0cm}\includegraphics[width=1.45in]{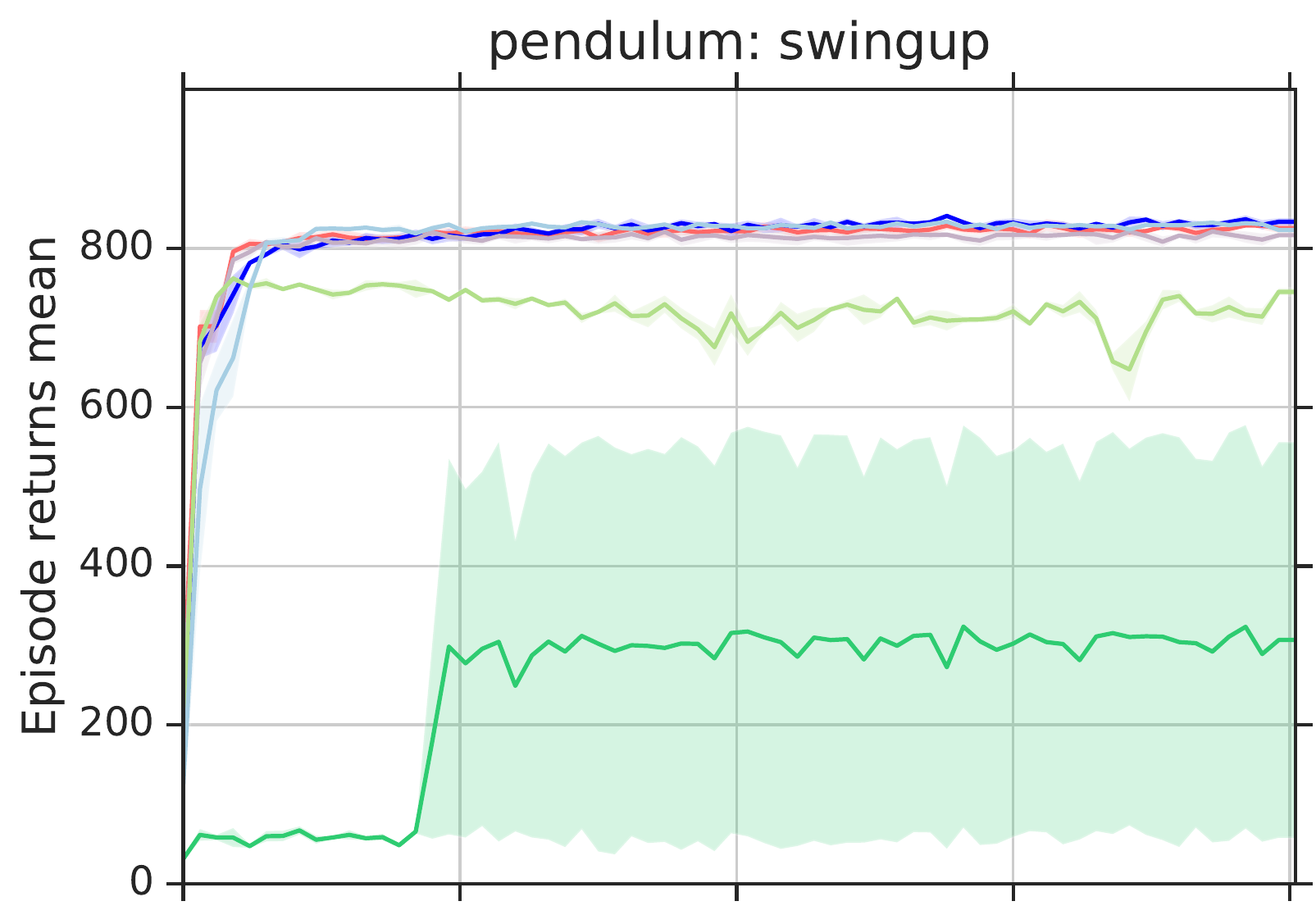}
\hspace*{0cm}\includegraphics[width=1.3in]{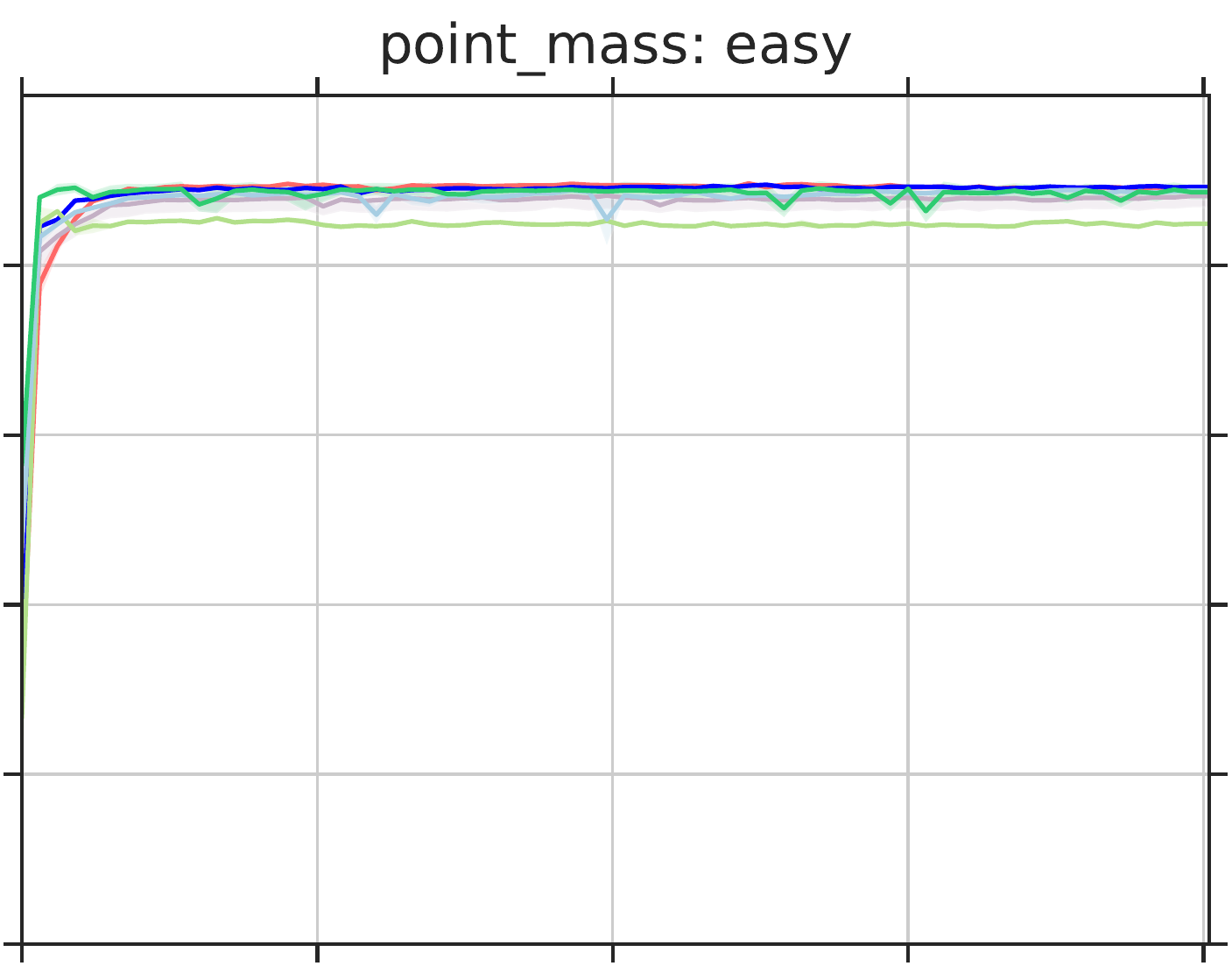}
\hspace*{0cm}\includegraphics[width=1.3in]{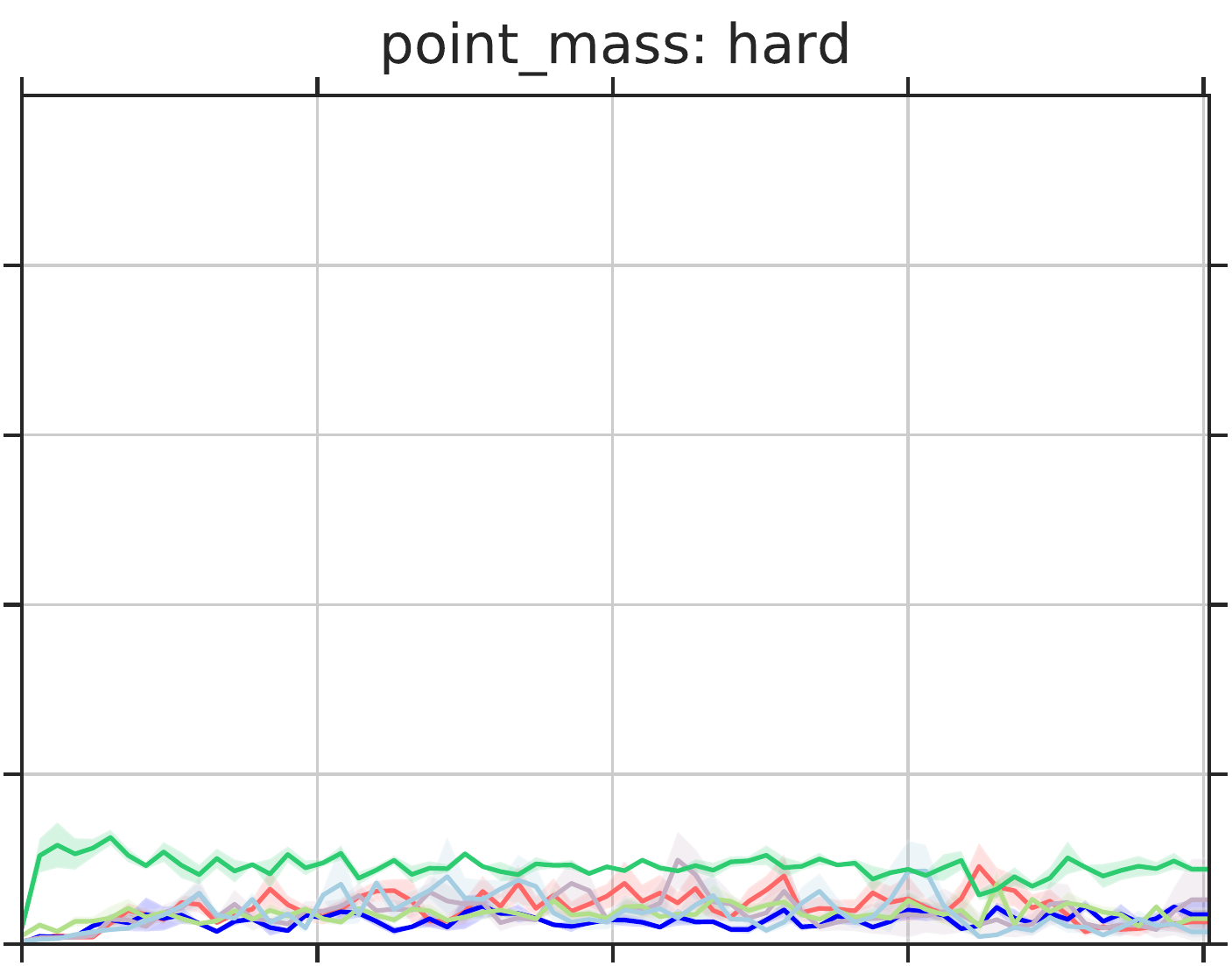}\\
\hspace*{0cm}\includegraphics[width=1.45in]{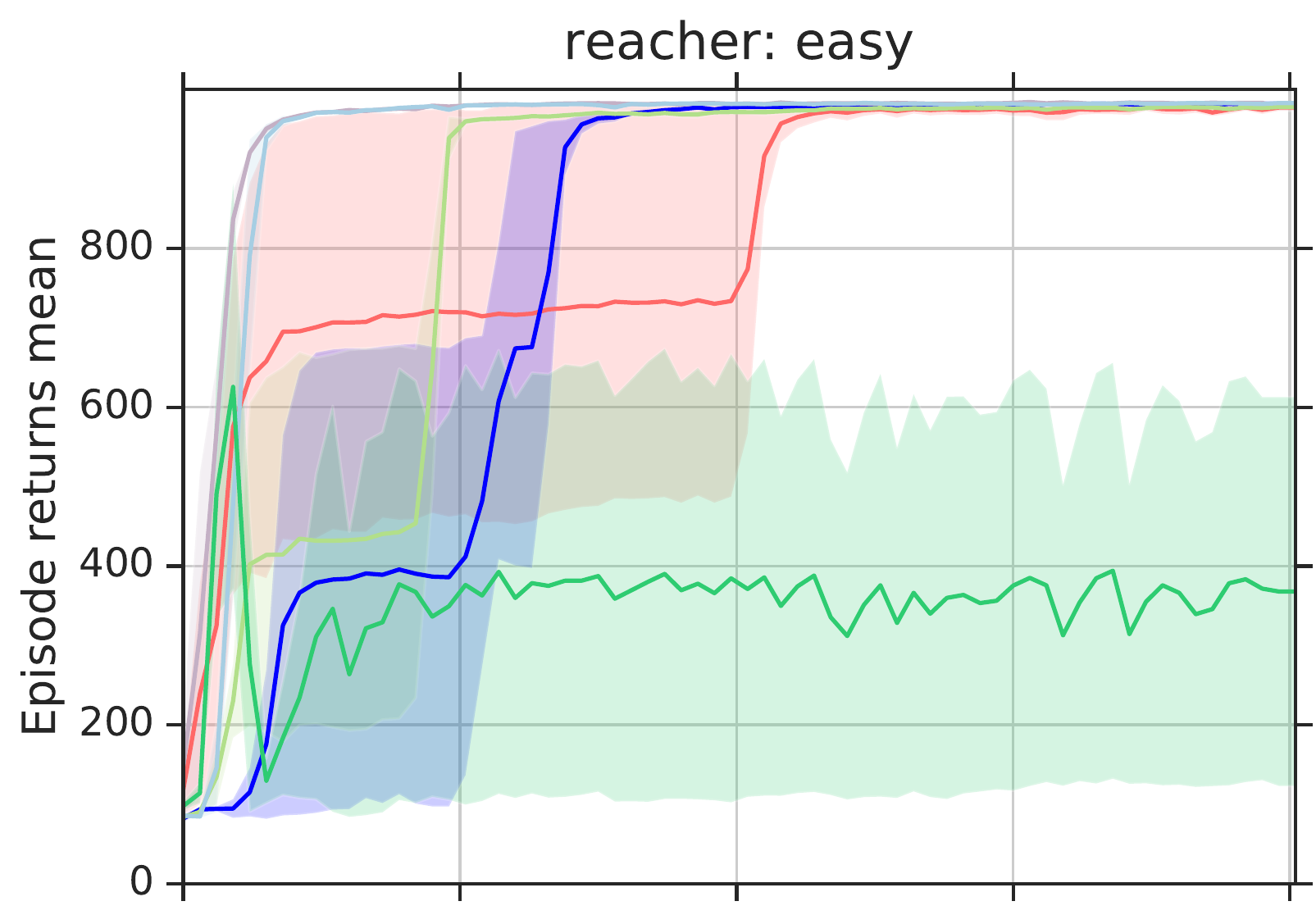}
\hspace*{0cm}\includegraphics[width=1.3in]{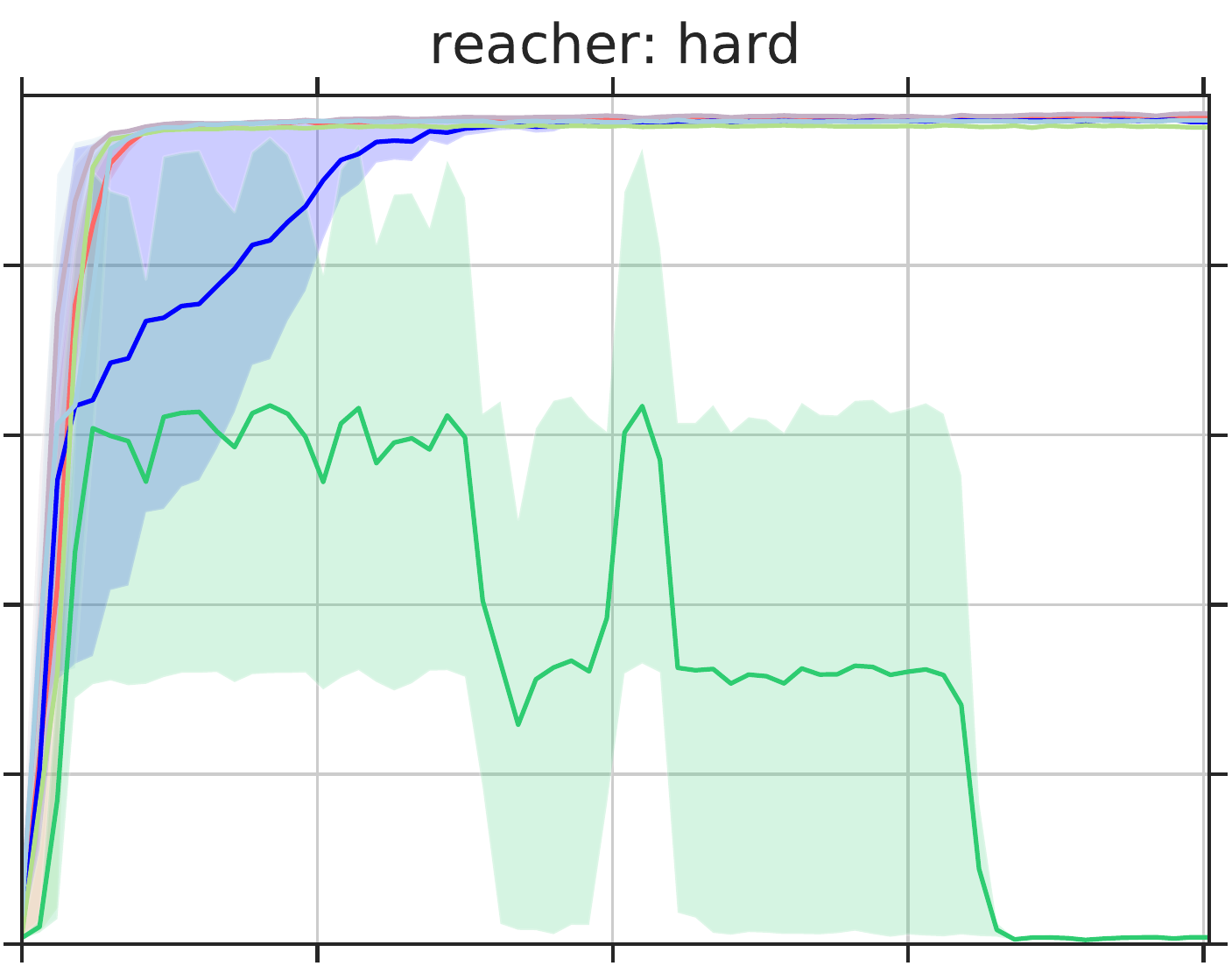}
\hspace*{0cm}\includegraphics[width=1.3in]{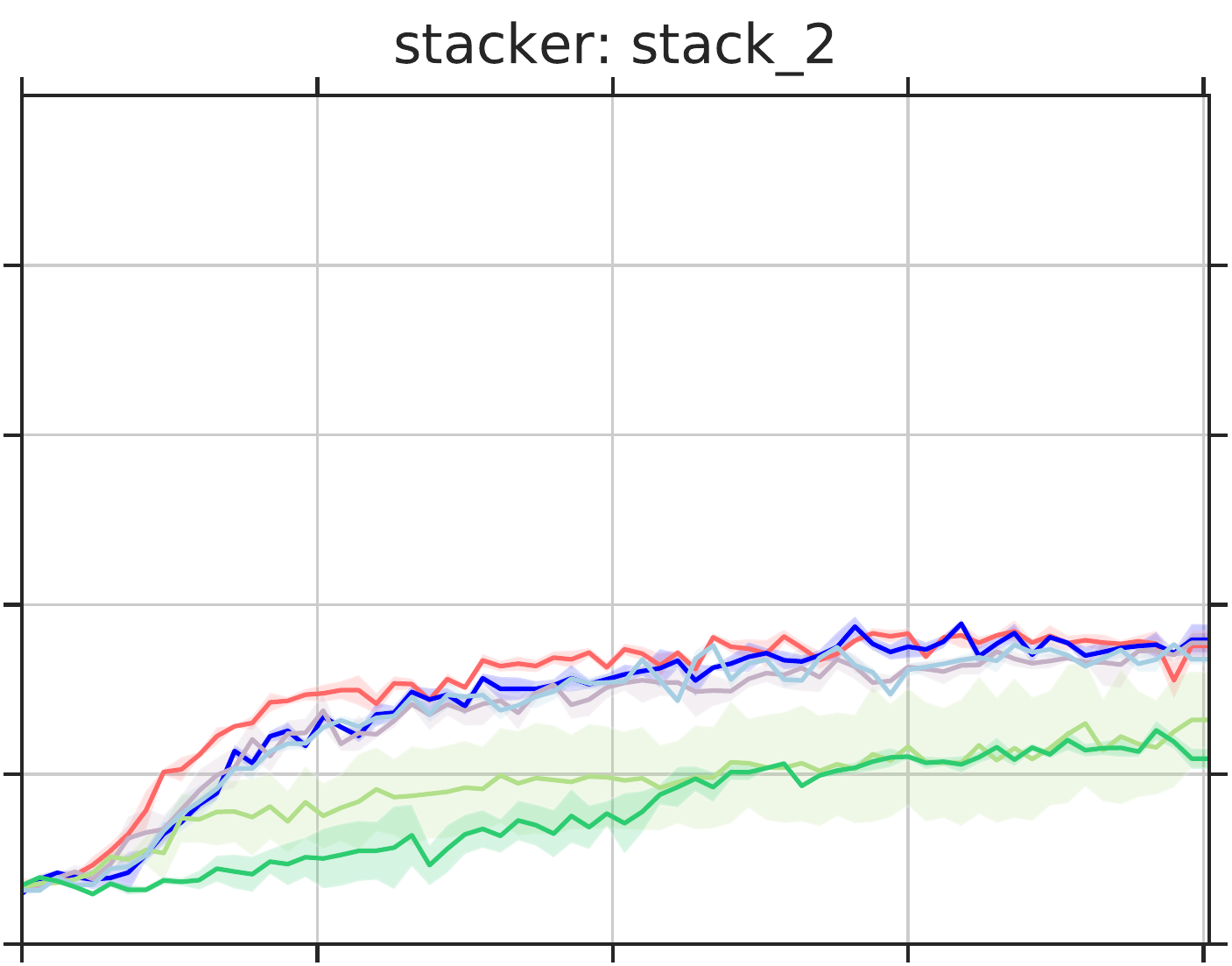}\\
\hspace*{0cm}\includegraphics[width=1.45in]{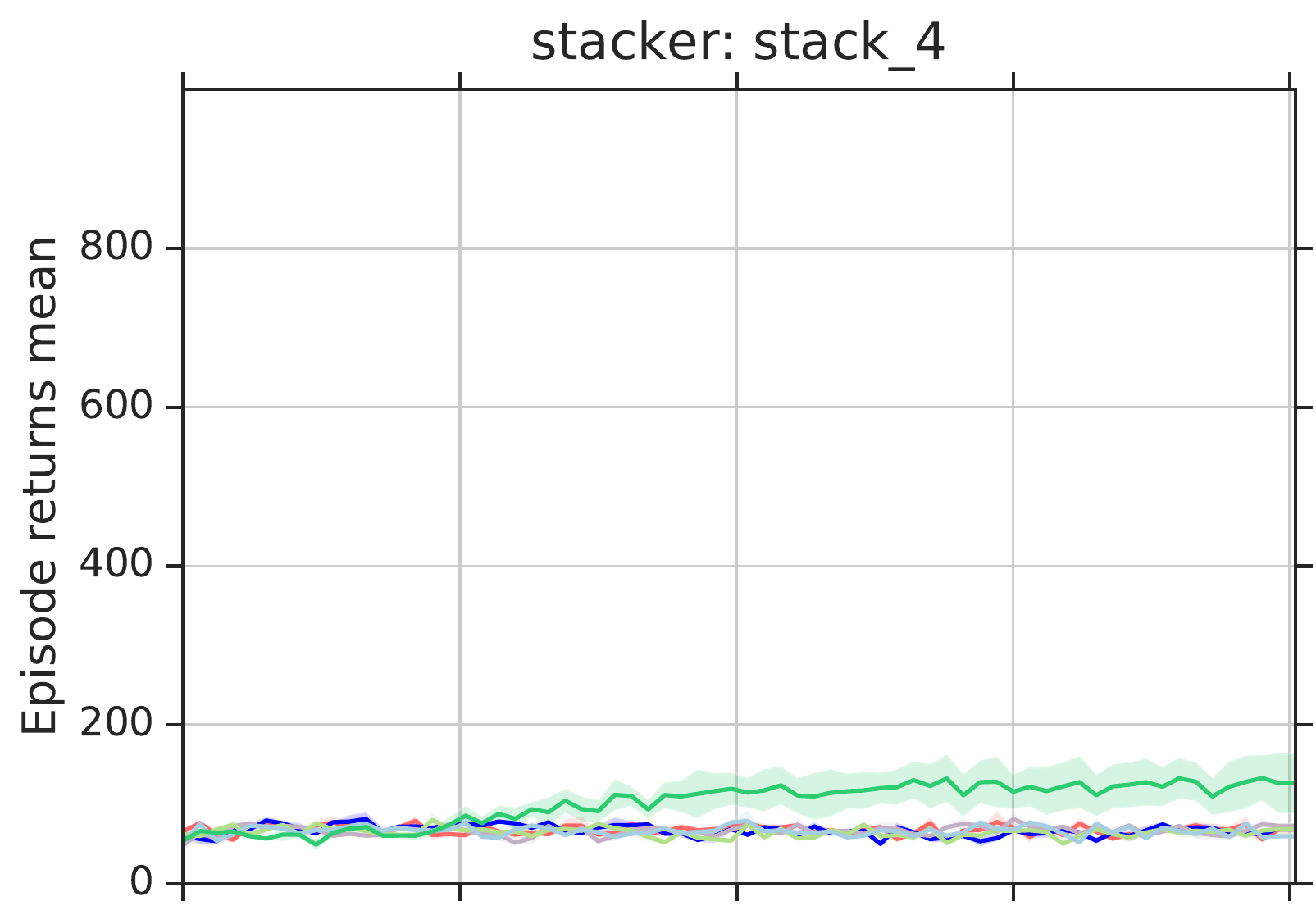}
\hspace*{0cm}\includegraphics[width=1.3in]{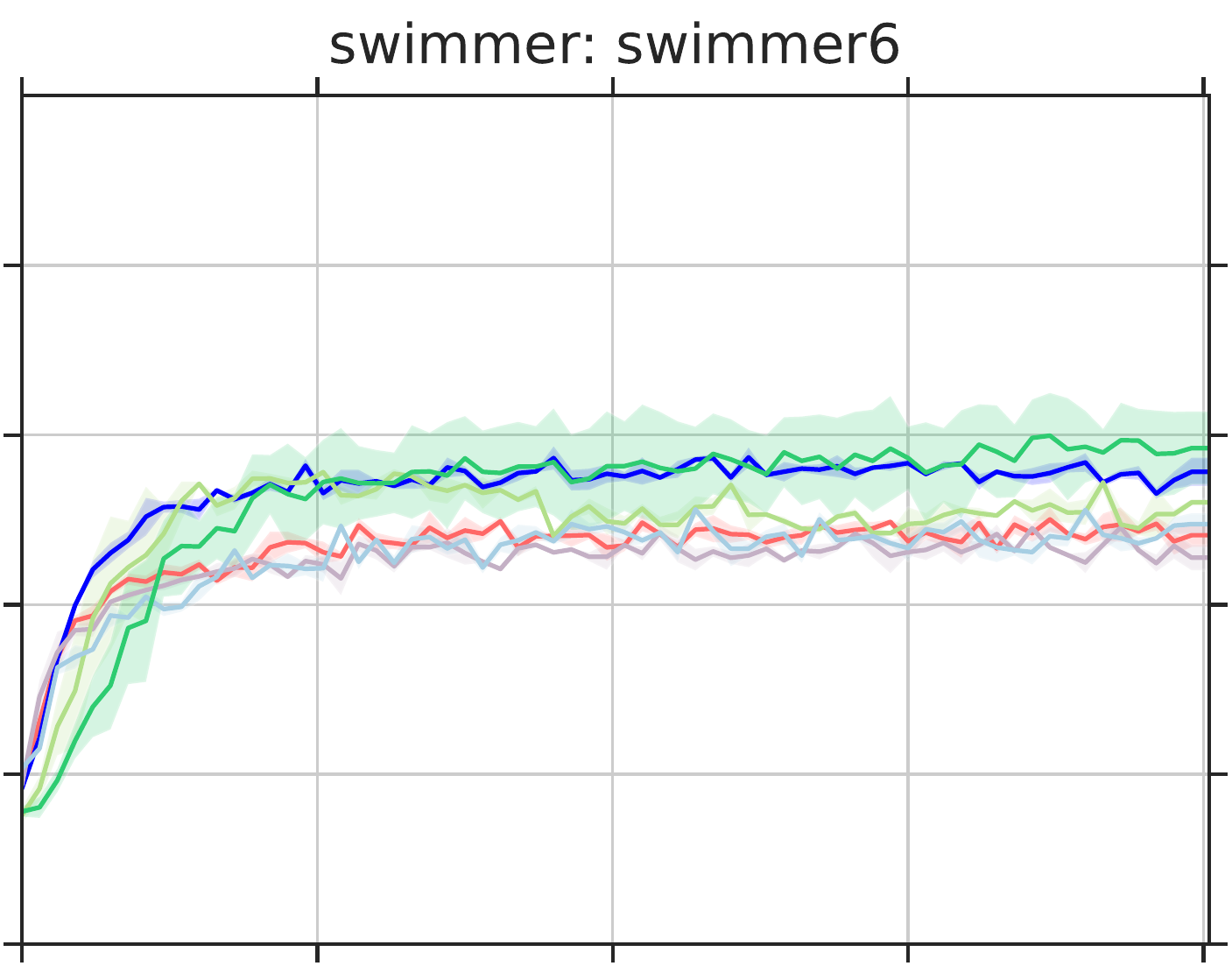}
\hspace*{0cm}\includegraphics[width=1.3in]{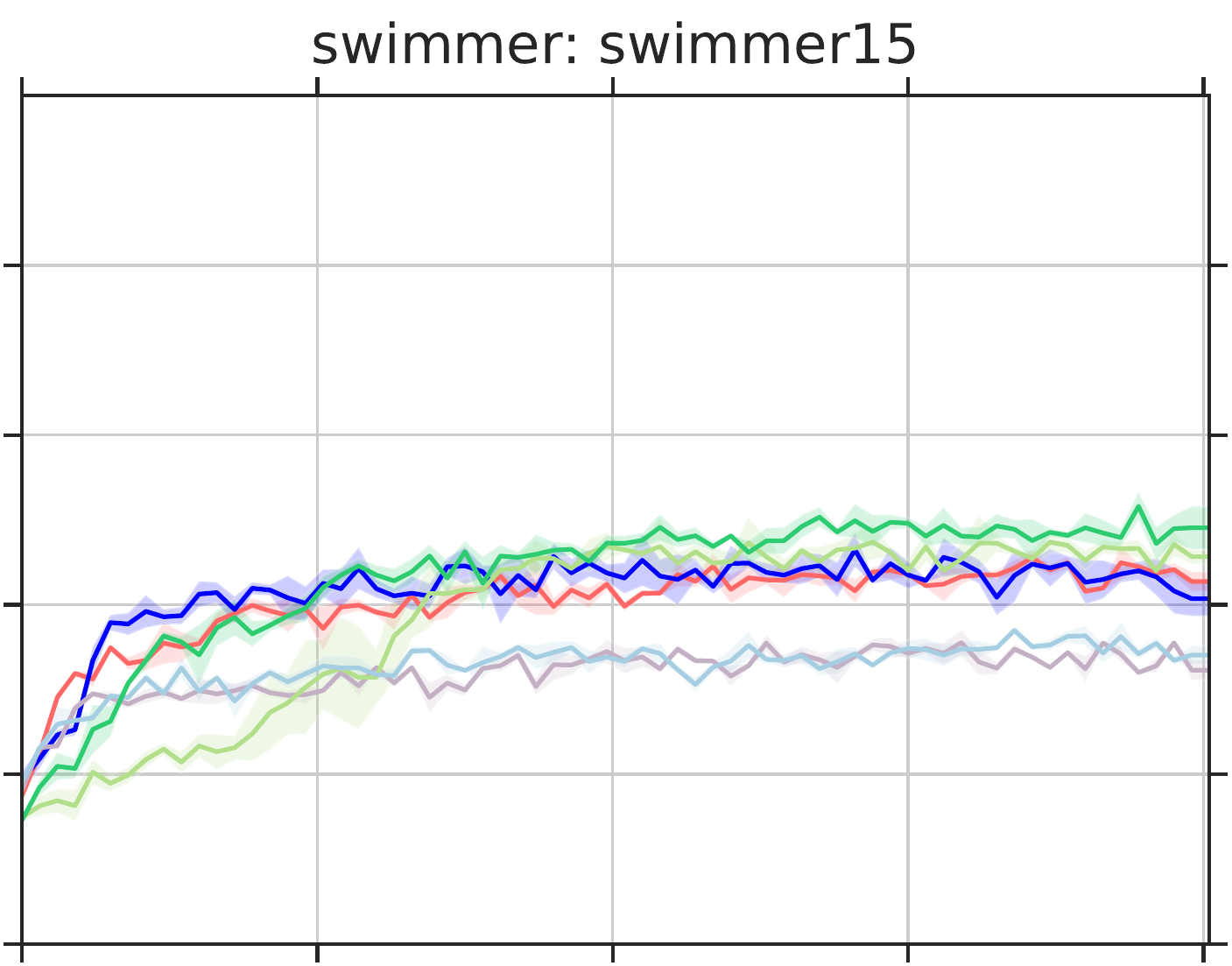}\\
\hspace*{0cm}\includegraphics[width=1.45in]{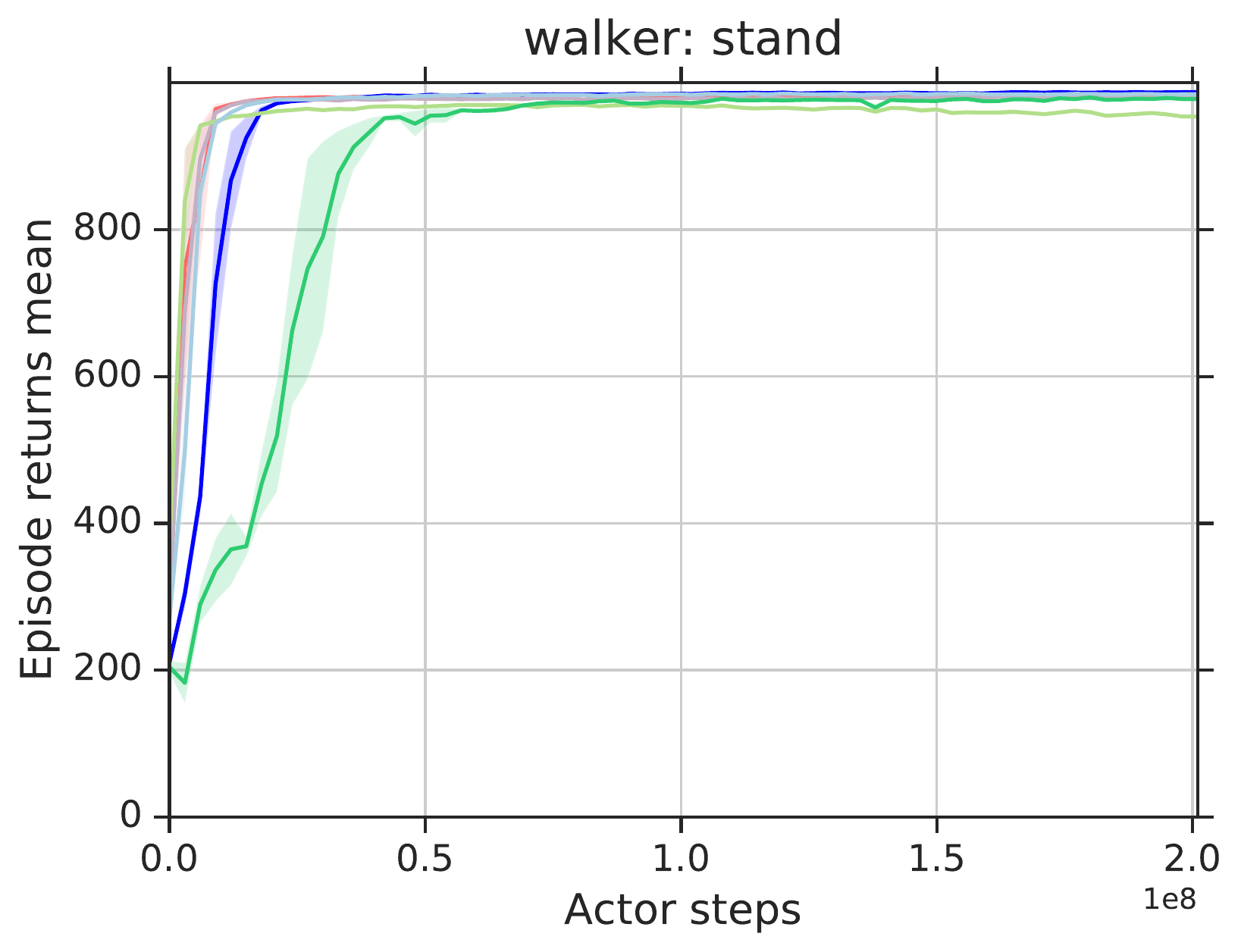}
\hspace*{0cm}\includegraphics[width=1.3in]{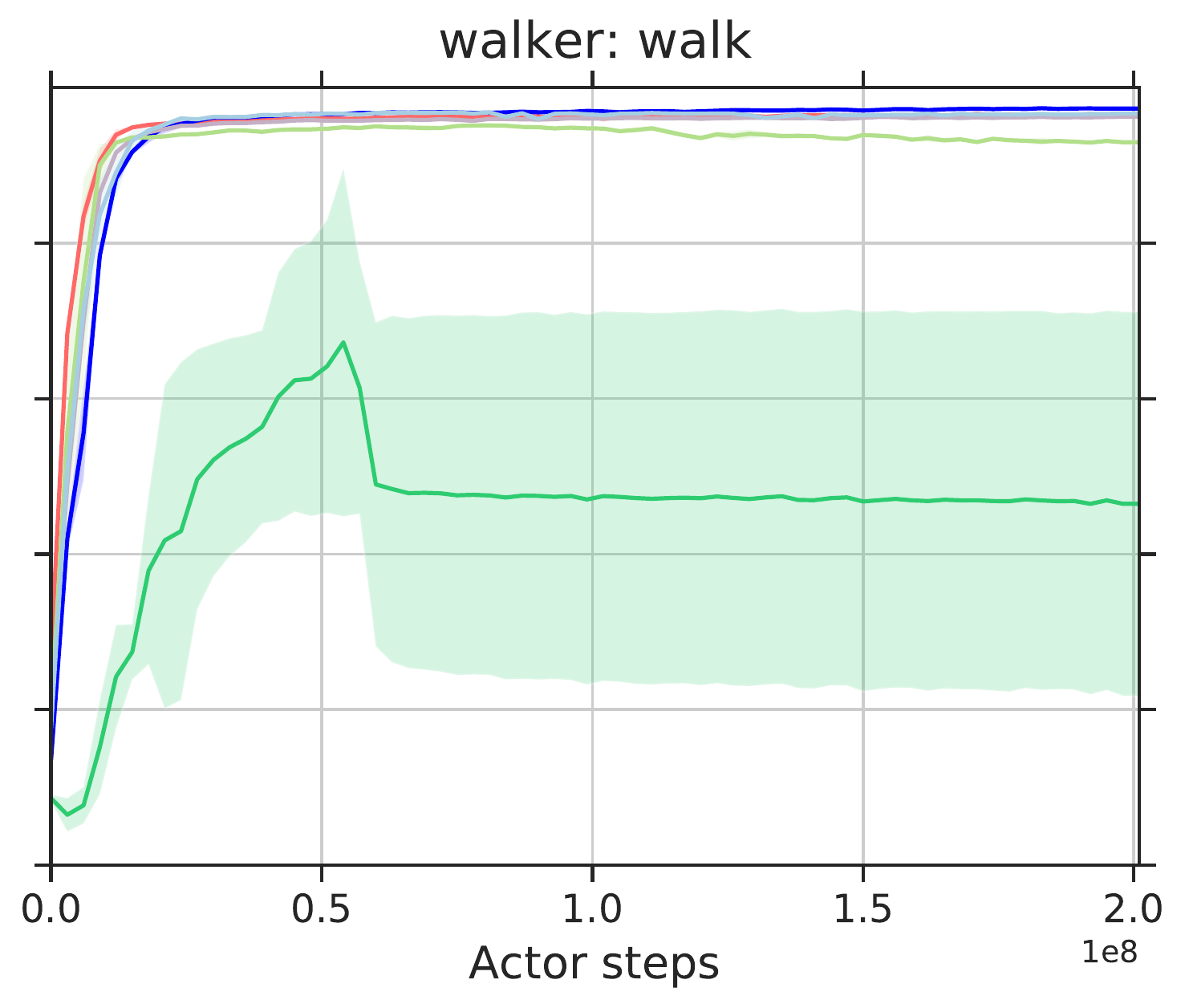}
\hspace*{0cm}\includegraphics[width=1.3in]{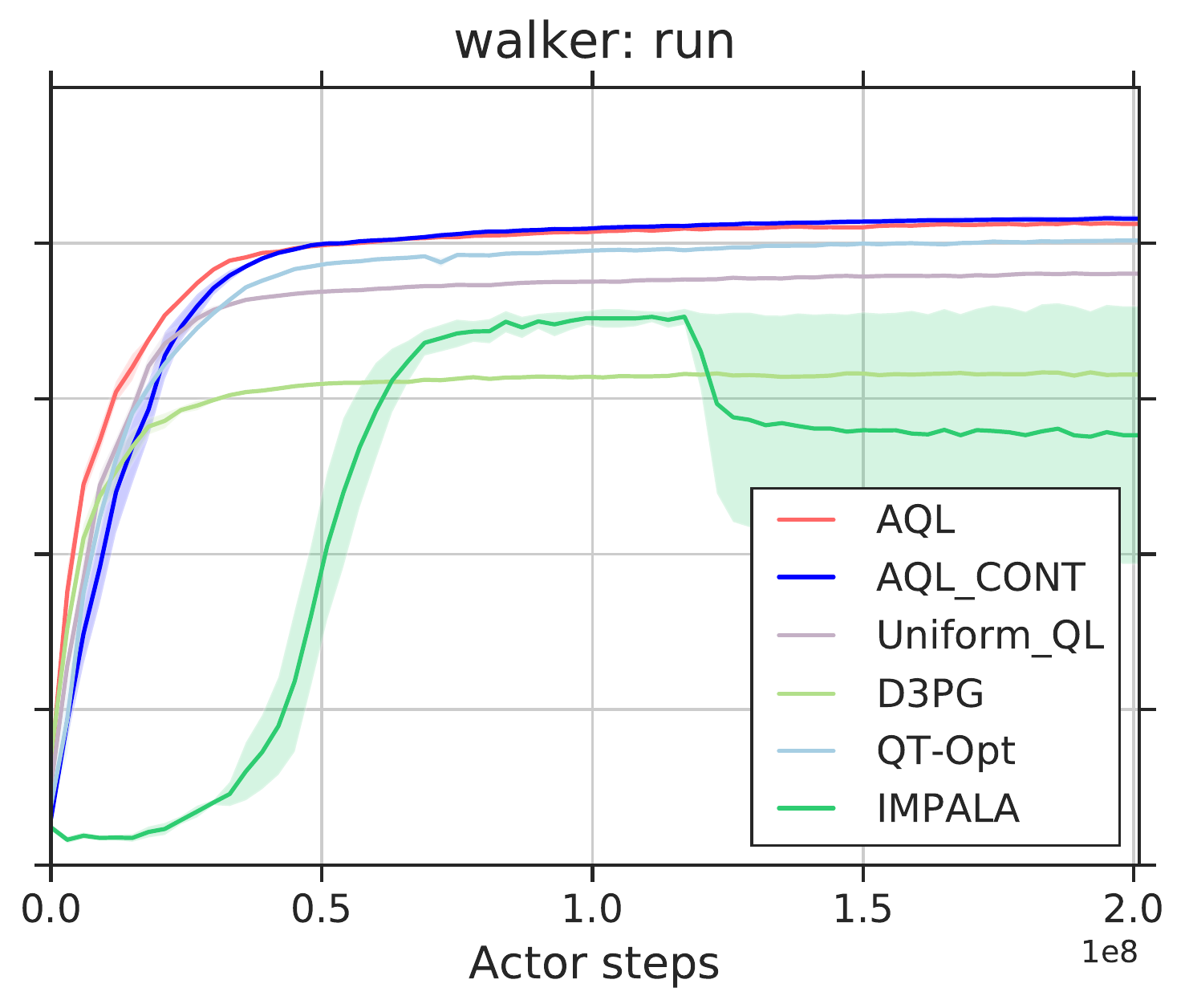}\\
\caption{Task specific learning curves for the Control Suite (2/2).}
\label{fig:control_suite_av_learning_curves_2}
\end{figure*}

Figures~\ref{fig:control_suite_av_learning_curves_1} and~\ref{fig:control_suite_av_learning_curves_2} show the individual learning curves of the Control Suite for all tasks. 

\begin{figure*}[h]
\centering
\includegraphics[width=2.5in]{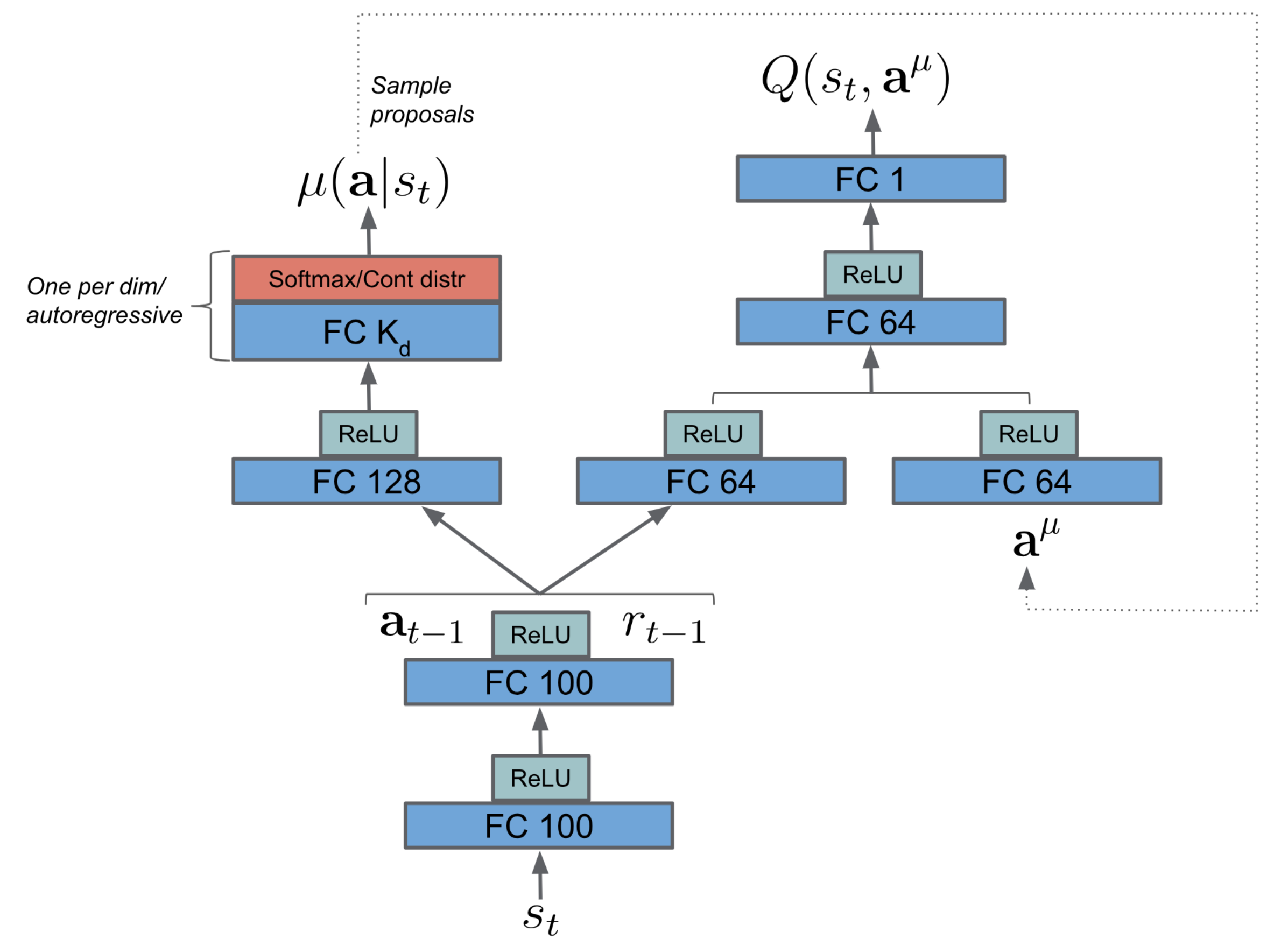}
\includegraphics[width=2.8in]{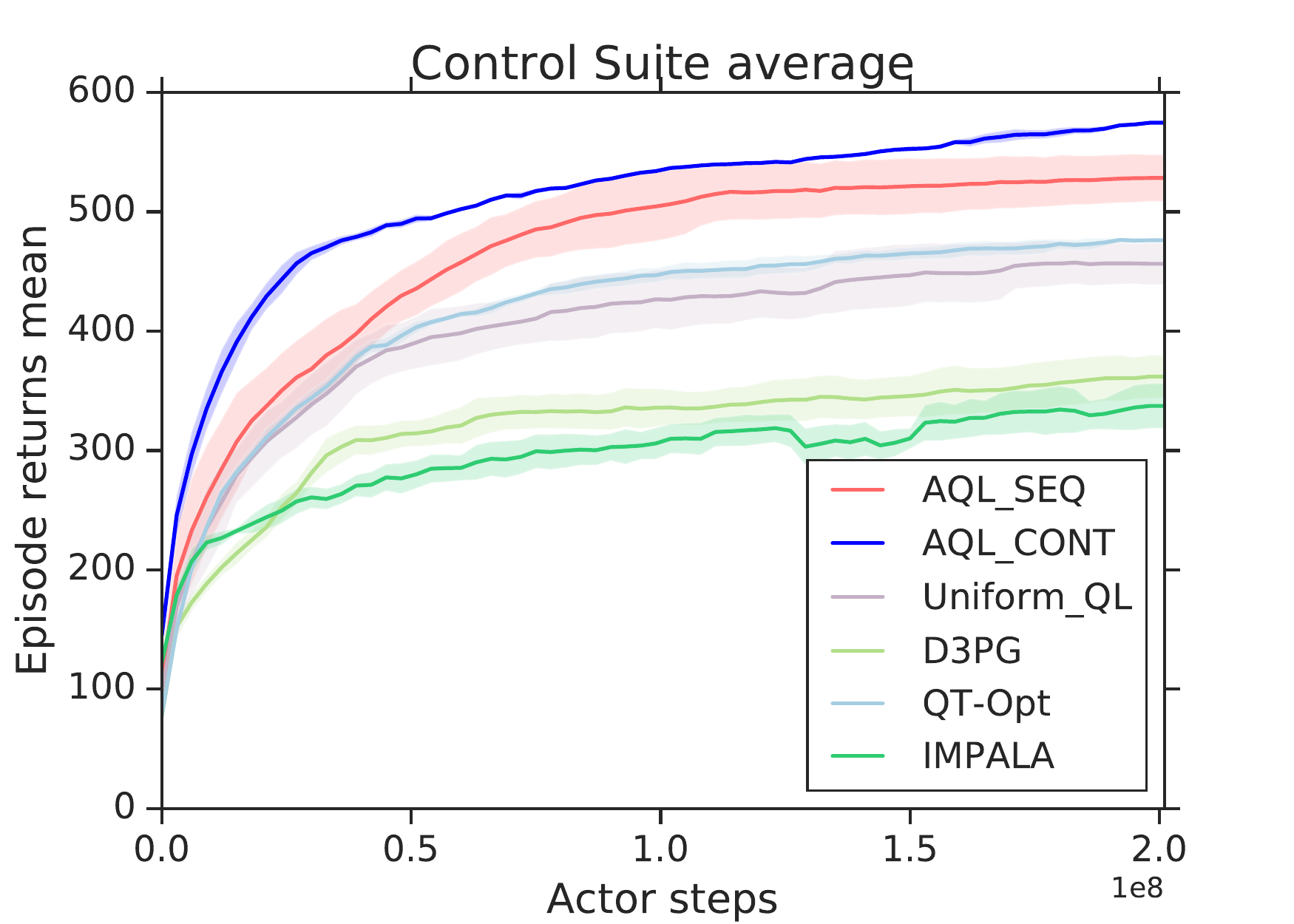}
\caption{Left: alternative architecture with RMSProp as the optimizer. Right: the corresponding learning curves of the mean return across all tasks in the Control Suite. The error bars represent the standard error of the mean episode return. The results are significantly worse for all baselines compared with the results in the main text.}
\label{fig:control_suite_old}
\end{figure*}

Somewhat surprisingly, we found that the choice of the network architecture and optimizer along with the optimizer hyperparameters can have a dramatic effect on the final performance.
We chose the hyperparameters for the preceding experiments by first running the baselines on some of the high-dimensional tasks for the considered methods and then committing to those settings that were found to work best on average for the all tasks.
Figure~\ref{fig:control_suite_old} shows the results of an earlier sweep with an alternative architecture and the RMSProp optimizer~\citep{tieleman2012rmsprop} instead of Adam.
The architecture in the earlier sweep shared weights in the first two layers and was deeper but had fewer units in each layer.
The results of the earlier sweep are significantly worse on average for all baselines except for IMPALA and continuous AQL. The results for continuous AQL are competitive with the best baseline of AQL (discretized, autoregressive proposal distribution), while the IMPALA results using this sharing architecture are comparable to the results from the main text.
D3PG was the most affected by the choice of architecture and optimizer type, being the second worst performing method when trained with RMSProp and the second best performing method when trained with Adam.

\FloatBarrier
 \end{document}